# Generalizing Boolean Satisfiability III: Implementation

**Heidi E. Dixon**                                     DIXON@OTSYS.COM
**Matthew L. Ginsberg**                          GINSBERG@OTSYS.COM
*On Time Systems, Inc.*
*1850 Millrace, Suite 1*
*Eugene, OR 97403 USA*

**David Hofer**                                 HOFER@CS.UOREGON.EDU
**Eugene M. Luks**                              LUKS@CS.UOREGON.EDU
*Computer and Information Science*
*University of Oregon*
*Eugene, OR 97403 USA*

**Andrew J. Parkes**                      PARKES@CIRL.UOREGON.EDU
*CIRL*
*1269 University of Oregon*
*Eugene, OR 97403 USA*

## Abstract

This is the third of three papers describing ZAP, a satisfiability engine that substantially generalizes existing tools while retaining the performance characteristics of modern high-performance solvers. The fundamental idea underlying ZAP is that many problems passed to such engines contain rich internal structure that is obscured by the Boolean representation used; our goal has been to define a representation in which this structure is apparent and can be exploited to improve computational performance. The first paper surveyed existing work that (knowingly or not) exploited problem structure to improve the performance of satisfiability engines, and the second paper showed that this structure could be understood in terms of groups of permutations acting on individual clauses in any particular Boolean theory. We conclude the series by discussing the techniques needed to implement our ideas, and by reporting on their performance on a variety of problem instances.

## 1. Introduction

This is the third of a series of three papers describing ZAP, a satisfiability engine that substantially generalizes existing tools while retaining the performance characteristics of modern high-performance solvers such as ZCHAFF (Moskewicz, Madigan, Zhao, Zhang, & Malik, 2001). In the first two papers in this series, we made arguments to the effect that:

- Many Boolean satisfiability problems incorporate a rich structure that reflects properties of the domain from which the problems arise, and recent improvements in the performance of satisfiability engines can be understood in terms of their ability to exploit this structure (Dixon, Ginsberg, & Parkes, 2004b, to which we will refer as ZAP1).

- The structure itself can be understood in terms of groups (in the algebraic sense) of permutations acting on individual clauses (Dixon, Ginsberg, Luks, & Parkes, 2004a, to which we will refer as ZAP2).

We showed that an implementation based on these ideas could be expected to combine the attractive computational properties of a variety of recent ideas, including efficient implementations of unit propagation (Zhang & Stickel, 2000) and extensions of the Boolean language to include cardinality or pseudo-Boolean constraints (Barth, 1995; Dixon & Ginsberg, 2000; Hooker, 1988), parity problems (Tseitin, 1970), or a limited form of quantification known as QPROP (Ginsberg & Parkes, 2000). In this paper, we discuss the implementation of a prover based on these ideas, and describe its performance on pigeonhole, parity and clique coloring problems. These classes of problems are known to be exponentially difficult for conventional Boolean satisfiability engines, and their formalization also highlights the group-based nature of the reasoning involved.

From a technical point of view, this is the most difficult of the three ZAP papers; we need to draw on the algorithms and theoretical constructions from ZAP2 and on results from computational group theory (GAP Group, 2004; Seress, 2003) regarding their implementation. Our overall plan for describing the implementation is as follows:

1. Section 2 is a review of material from ZAP2. We begin in Section 2.1 by presenting both the Boolean satisfiability algorithms that we hope to generalize and the basic algebraic ideas underlying ZAP. Section 2.2 describes the group-theoretic computations required by the ZAP implementation.

2. Section 3 gives a brief – and necessarily incomplete – introduction to some of the ideas in computational group theory that we use.

3. Sections 4 and 5 describe the implementations of the computations discussed in Section 2. For each basic construction, we describe the algorithm used and give an example of the computation in action. If there is an existing implementation of something in the public domain system GAP (2004), we only provide a pointer to that implementation; for concepts that we needed to implement from scratch, additional detail is provided.

4. Section 6 extends the basic algorithms of Section 5 to deal with unit propagation, where we want to compute not a single unit clause instance, but a list of all of the unit consequences of an augmented clause.

5. Section 7 discusses the implementation of Zhang and Stickel's (2000) watched literal idea in our setting.

6. Section 8 describes a technique that can be used to select among the possible resolvents of two augmented clauses. This is functionality with no analog in a conventional prover, where there is only a single ground reason for the truth or falsity of any given variable. If the reasons are augmented clauses, there may be a variety of ways in which ground instances of those clauses can be combined.

7. After describing the algorithms, we present experimental results regarding performance in Sections 9 and 10. Section 9 reports on the performance of ZAP's individual algorithmic components, while Section 10 contrasts ZAP's overall performance to that of its CNF-based predecessors.[1] Since our focus in this paper is on the algorithms needed by ZAP, we report performance only for relatively theoretical examples that clearly involve group-based reasoning. Performance on a wider range of problem classes will be reported elsewhere.

---

1. A description of ZAP's input language is contained in Appendix B.

8. Concluding remarks appear in Section 11.

Except for Section 3, proofs are generally deferred to Appendix A in the interests of maintaining the continuity of our exposition. Given the importance of computational group theory to the ideas that we will be presenting, we strongly suggest that the reader work through the proofs in Section 3 of the paper.

This is a long and complex paper; we make no apologies. ZAP is an attempt to synthesize two very different fields, each complex in its own right: computational group theory and implementations of Boolean satisfiability engines. Computational group theory, in addition to its inherent complexity, is likely to be foreign to an AI audience. Work on complete algorithms for Boolean satisfiability has also become increasingly sophisticated over the past decade or so, with the introduction of substantial and nonintuitive modifications to the original DPLL algorithm such as relevance-bounded learning (Bayardo & Miranker, 1996; Bayardo & Schrag, 1997; Ginsberg, 1993) and watched literals (Zhang & Stickel, 2000). As we bring these two fields together, we will see that a wide range of techniques from computational group theory is relevant to the problems of interest to us; our goal is also not simply to translate DPLL to the new setting, but to show that *all* of the recent work on Boolean satisfiability can be moved across. In at least one case (Lemma 5.26), we also need to extend existing computational group theory results. And finally, there are new satisfiability techniques and possibilities that arise only *because* of the synthesis that we are proposing (Section 8), and we will describe some of those as well.

This paper is not intended to be self-contained. We assume throughout that the reader is familiar with the material that we presented in ZAP2; some of the results from that paper are repeated here for convenience, but the accompanying text is not intended to stand alone.

Finally – and in spite of the disclaimers of the previous two paragraphs – this paper is not intended to be complete. Our goal is to present a practical minimum of what is required to implement an effective group-based reasoning system. The results that we have obtained, both theoretical as described in ZAP2 and practical as described here, excite us. But we are just as excited by the number of issues that we have not yet explored. Our primary goal is to present the foundation needed if other interested researchers are to explore these ideas with us.

## 2. ZAP Fundamentals and Basic Structure

Our overview of ZAP involves summarizing work from two distinct areas: existing Boolean satisfiability engines, and the group-theoretic elements underlying ZAP.

### 2.1 Boolean Satisfiability

We begin with a description of the architecture of modern Boolean satisfiability engines. We start with the unit propagation procedure, which we describe as follows:

**Definition 2.1** *Given a Boolean satisfiability problem described in terms of a set $C$ of clauses, a* partial assignment *is an assignment of values (true or false) to some subset of the variables appearing in $C$. We represent a partial assignment $P$ as a sequence of consistent literals $P = \langle l_i \rangle$ where the appearance of $v_i$ in the sequence means that $v_i$ has been set to true, and the appearance of $\neg v_i$ means that $v_i$ has been set to false.*

*An* annotated partial assignment *is a sequence $P = \langle (l_i, c_i) \rangle$ where $c_i$ is the reason for the associated choice $l_i$. If $c_i = \texttt{true}$, it means that the variable was set as the result of a branching decision; otherwise, $c_i$ is a clause that entails $l_i$ by virtue of the choices of the previous $l_j$ for $j < i$. An annotated partial assignment will be called* sound *with respect to a set of constraints $C$ if $C \models c_i$ for each reason $c_i$. (See* ZAP2 *for additional details.)*

*Given a (possibly annotated) partial assignment $P$, we denote by $S(P)$ the literals that are satisfied by $P$, and by $U(P)$ the set of literals that are unvalued by $P$.*

**Procedure 2.2 (Unit propagation)** *To compute* UNIT-PROPAGATE$(C, P)$ *for a set $C$ of clauses and an annotated partial assignment $P = \langle (l_1, c_1), \ldots, (l_n, c_n) \rangle$:*

1  **while** there is a $c \in C$ with $c \cap S(P) = \varnothing$ and $|c \cap U(P)| \leq 1$
2      **do if** $c \cap U(P) = \varnothing$
3          **then** $l_i \leftarrow$ the literal in $c$ with the highest index in $P$
4              **return** $\langle \texttt{true}, \texttt{resolve}(c, c_i) \rangle$
5          **else** $l \leftarrow$ the literal in $c$ unassigned by $P$
6              $P \leftarrow \langle P, (l, c) \rangle$
7  **return** $\langle \texttt{false}, P \rangle$

The result returned depends on whether or not a contradiction was encountered during the propagation, with the first result returned being `true` if a contradiction was found and `false` if none was found. In the former case, where the clause $c$ has no unvalued literals (line 2), $l_i$ is the last literal set in $c$, and $c_i$ is the reason that $l_i$ was set in a way that caused $c$ to be unsatisfiable. We resolve $c$ with $c_i$ and return the result as a new nogood for the problem in question. Otherwise, we eventually return the partial assignment, augmented to include the variables that were set during the propagation process.

Given unit propagation, the overall inference procedure is the following:

**Procedure 2.3 (Relevance-bounded learning,** RBL**)** *Given a* SAT *problem $C$, a set of learned nogoods $D$ and an annotated partial assignment $P$, to compute* RBL$(C, D, P)$:

1   $\langle x, y \rangle \leftarrow$ UNIT-PROPAGATE$(C \cup D, P)$
2   **if** $x = \mathtt{true}$
3      **then** $c \leftarrow y$
4          **if** $c$ is empty
5            **then return** FAILURE
6            **else** remove successive elements from $P$ so that $c$ is unit
7                 $D \leftarrow \mathtt{learn}(D, P, c)$
8                 **return** RBL$(C, D, P)$
9      **else** $P \leftarrow y$
10         **if** $P$ is a solution to $C$
11         **then return** $P$
12         **else** $l \leftarrow$ a literal not assigned a value by $P$
13            **return** RBL$(C, D, \langle P, (l, \mathtt{true}) \rangle)$

As might be expected, the procedure is recursive. If at any point unit propagation produces a contradiction $c$, we use the (currently unspecified) $\mathtt{learn}$ procedure to incorporate $c$ into the solver's current state, and then recurse. If $c$ is empty, it means that we have derived a contradiction and the procedure fails. In the backtracking step (line 6), we backtrack not just until $c$ is satisfiable, but until it enables a unit propagation. This technique is used in zChaff (Moskewicz et al., 2001). It leads to increased flexibility in the choice of variable to be assigned after the backtrack is complete, and generally improves performance.

If unit propagation does not indicate the presence of a contradiction or produce a solution to the problem in question, we pick an unvalued literal, set it to true, and recurse again. Note that we don't need to set the literal $l$ to true or false; if we eventually need to backtrack and set $l$ to false, that will be handled by the modification to $P$ in line 6.

Finally, we need to present the procedure used to incorporate a new nogood into the clausal database $C$. In order to do that, we make the following definition:

**Definition 2.4** *Let $\vee_i l_i$ be a clause, which we will denote by $c$, and let $P$ be a partial assignment. We will say that the* possible value *of $c$ under $P$ is given by*

$$\mathtt{poss}(c, P) = |\{i | \neg l_i \notin P\}| - 1$$

*If no ambiguity is possible, we will write simply $\mathtt{poss}(c)$ instead of $\mathtt{poss}(c, P)$. In other words, $\mathtt{poss}(c)$ is the number of literals that are either already satisfied or not valued by $P$, reduced by one (since the clause requires at least one true literal).*

Note that $\mathtt{poss}(c, P) = |c \cap [U(P) \cup S(P)]| - 1$, since each expression is one less than the number of potentially satisfied literals in $c$.

The possible value of a clause is essentially a measure of what other authors have called its *irrelevance* (Bayardo & Miranker, 1996; Bayardo & Schrag, 1997; Ginsberg, 1993). An unsatisfied clause $c$ with $\mathtt{poss}(c, P) = 0$ can be used for unit propagation; we will say that such a clause is *unit*. If $\mathtt{poss}(c, P) = 1$, it means that a change to a single variable can lead to a unit propagation, and so on. The notion of learning used in relevance-bounded inference is now captured by:

**Procedure 2.5** *Given a set of clauses $C$ and an annotated partial assignment $P$, to compute* `learn`$(C, P, c)$*, the result of adding to $C$ a clause $c$ and removing irrelevant clauses:*

1   remove from $C$ any $d \in C$ with `poss`$(d, P) > k$
2   **return** $C \cup \{c\}$

We hope that all of this is familiar; if not, please refer to ZAP2 or to the other papers that we have cited for fuller explanations.

In ZAP, we continue to work with these procedures in approximately their current form, but replace the idea of a clause (a disjunction of literals) with that of an *augmented* clause:

**Definition 2.6** *An* augmented clause *in an n-variable Boolean satisfiability problem is a pair $(c, G)$ where $c$ is a Boolean clause and $G$ is a group such that $G \leq W_n$. A (nonaugmented) clause $c'$ is an* instance *of an augmented clause $(c, G)$ if there is some $g \in G$ such that $c' = c^g$.[2] The clause $c$ itself will be called the* base instance *of $(c, G)$.*

Roughly speaking, an augmented clause consists of a conventional clause and a group $G$ of permutations of the literals in the theory; the intent is that we can act on the clause with any element of the group and still get a clause that is "part" of the original theory. The group $G$ is required to be a subgroup of the group of "permutations and complementations" (Harrison, 1989) $W_n = S_2 \wr S_n$, where each permutation $g \in G$ can permute the variables in the problem and flip the signs of an arbitrary subset as well. We showed in ZAP2 that suitably chosen groups correspond to cardinality constraints, parity constraints (the group flips the signs of any even number of variables), and universal quantification over finite domains.

We must now lift the previous three procedures to an augmented setting. In unit propagation, for example, instead of checking to see if any clause $c \in C$ is unit given the assignments in $P$, we now check to see if any augmented clause $(c, G)$ has a unit instance. Other than that, the procedure is essentially unchanged from Procedure 2.2:

**Procedure 2.7 (Unit propagation)** *To compute* UNIT-PROPAGATE$(C, P)$ *for a set of clauses $C$ and an annotated partial assignment $P = \langle (l_1, c_1), \ldots, (l_n, c_n) \rangle$:*

1   **while** there is a $(c, G) \in C$ and $g \in G$ with $c^g \cap S(P) = \varnothing$ and $|c^g \cap U(P)| \leq 1$
2       **do if** $c^g \cap U(P) = \varnothing$
3           **then** $l_i \leftarrow$ the literal in $c^g$ with the highest index in $P$
4               **return** $\langle$`true`, `resolve`$((c^g, G), c_i) \rangle$
5           **else** $l \leftarrow$ the literal in $c^g$ unassigned by $P$
6               $P \leftarrow \langle P, (l, (c^g, G)) \rangle$
7   **return** $\langle$`false`, $P \rangle$

The basic inference procedure itself is also virtually unchanged:

---

2. As in ZAP2 and as used by the computational group theory community, we denote the image of a clause $c$ under a group element $g$ by $c^g$ instead of the possibly more familiar $g(c)$. As explained in ZAP2, this reflects the fact that the composition $fg$ of two permutations acts with $f$ first and with $g$ second.

**Procedure 2.8 (Relevance-bounded learning, RBL)** *Given a SAT problem $C$, a set of learned clauses $D$, and an annotated partial assignment $P$, to compute* RBL$(C, D, P)$:

```
1   ⟨x, y⟩ ← UNIT-PROPAGATE(C ∪ D, P)
2   if x = true
3      then (c, G) ← y
4              if c is empty
5                 then return FAILURE
6                 else remove successive elements from P so that c is unit
7                          D ← learn(D, P, (c, G))
8                          return RBL(C, D, P)
9      else  P ← y
10              if P is a solution to C
11                 then return P
12                 else  l ← a literal not assigned a value by P
13                          return RBL(C, D, ⟨P, (l, true)⟩)
```

In line 3, although unit propagation returns an augmented clause $(c, G)$, the base instance $c$ is still the reason for the backtrack by virtue of line 6 of Procedure 2.7. It follows that line 6 of Procedure 2.8 is unchanged from the Boolean version.

To lift Procedure 2.5 to our setting, we need an augmented version of Definition 2.4:

**Definition 2.9** *Let $(c, G)$ be an augmented clause, and $P$ a partial assignment. Then by* poss$((c, G), P)$ *we will mean the minimum possible value of an instance of $(c, G)$, so that*

$$\text{poss}((c, G), P) = \min_{g \in G} \text{poss}(c^g, P)$$

Procedure 2.5 can now be used unchanged, with $d$ being an augmented clause instead of a simple one. The effect of Definition 2.9 is to cause us to remove only augmented clauses for which every instance is irrelevant. Presumably, it will be useful to retain the clause as long as it has some relevant instance.

In ZAP2, we showed that a proof engine built around the above three procedures would have the following properties:

- Since the number of generators of a group can be made logarithmic in the group size, it would achieve exponential improvements in basic representational efficiency.

- Since only $k$-relevant nogoods are retained as the search proceeds, the memory requirements remain polynomial in the size of the problem being solved.

- It can produce polynomially sized proofs of the pigeonhole and clique coloring problems, and any parity problem.

- It generalizes first-order inference provided that all quantifiers are universal and all domains of quantification are finite.

We stated without proof (and will show in this paper) that the unit propagation procedure 2.7 can be implemented in a way that generalizes both subsearch (Ginsberg & Parkes, 2000) and Zhang and Stickel's (2000) watched literal idea.

## 2.2 Group-Theoretic Elements

Examining the above three procedures, the elements that are new relative to Boolean engines are the following:

1. In line 1 of the unit propagation procedure 2.7, we need to find unit instances of an augmented clause $(c, G)$.

2. In line 4 of the same procedure 2.7, we need to compute the resolvent of two augmented clauses.

3. In line 1 of the learning procedure 2.5, we need to determine if an augmented clause has any relevant instances.

The first and third of these needs are different from the second. For resolution, we need the following definitions:

**Definition 2.10** *For a permutation $p$ and set $S$ with $S^p = S$, by $p|_S$ we will mean the restriction of $p$ to the given set, and we will say that $p$ is a lifting of $p|_S$ back to the original set on which $p$ acts.*

**Definition 2.11** *For a set $\Omega$, we will denote by $\mathrm{Sym}(\Omega)$ the group of permutations of $\Omega$. If $G \leq \mathrm{Sym}(\Omega)$ is a subgroup of this group and $S \leq \Omega$, we will say that $G$ acts on $S$.*[3]

**Definition 2.12** *Suppose that $G$ acts on a set $S$. Then for any $x \in S$, the orbit of $x$ in $G$, to be denoted by $x^G$, is given by $x^G = \{x^g | g \in G\}$. If $T \subseteq S$, then the $G$-closure of $T$, to be denoted $T^G$, is the set*

$$T^G = \{t^g | t \in T \text{ and } g \in G\}$$

**Definition 2.13** *For $K_1, \ldots, K_n \subseteq \Omega$ and $G_1, \ldots, G_n \leq \mathrm{Sym}(\Omega)$, we will say that a permutation $\omega \in \mathrm{Sym}(\Omega)$ is a stable extension of $G_1, \ldots, G_n$ for $K_1, \ldots, K_n$ if there are $g_i \in G_i$ such that for all $i$, $\omega|_{K_i^{G_i}} = g_i|_{K_i^{G_i}}$. We will denote the set of stable extensions of $G_1, \ldots, G_n$ for $K_1, \ldots, K_n$ by $\mathtt{stab}(K_i, G_i)$.*

The set of stable extensions $\mathtt{stab}(K_i, G_i)$ is closed under composition, and is therefore a subgroup of $\mathrm{Sym}(\Omega)$.

**Definition 2.14** *Suppose that $(c_1, G_1)$ and $(c_2, G_2)$ are augmented clauses. Then the result of resolving $(c_1, G_1)$ and $(c_2, G_2)$, to be denoted by $\mathtt{resolve}((c_1, G_1), (c_2, G_2))$, is the augmented clause $(\mathtt{resolve}(c_1, c_2), \mathtt{stab}(c_i, G_i) \cap W_n)$.*

It follows from the above definitions that computing the resolvent of two augmented clauses as required by Procedure 2.7 is essentially a matter of computing the set of stable extensions of the groups in question. We will return to this problem in Section 4.

The other two problems can both be viewed as instances of the following:

---

3. For convenience, we depart from standard usage and permit $G$ to map points in $S$ to images outside of $S$.

**Definition 2.15** *Let $c$ be a clause, viewed as a set of literals, and $G$ a group of permutations acting on $c$. Now fix sets of literals $S$ and $U$, and an integer $k$. We will say that the $k$-transporter problem is that of finding a $g \in G$ such that $c^g \cap S = \emptyset$ and $|c^g \cap U| \leq k$, or reporting that no such $g$ exists.*

To find a unit instance of $(c, G)$, we set $S$ to be the set of satisfied literals and $U$ the set of unvalued literals. Taking $k = 1$ implies that we are searching for an instance with no satisfied and at most one unvalued literal.

To find a relevant instance, we set $S = \emptyset$ and $U$ to be the set of all satisfied *or* unvalued literals. Taking $k$ to be the relevance bound corresponds to a search for a relevant instance.

The remainder of the theoretical material in this paper is therefore focused on these two problems: computing the stable extensions of a pair of groups, and solving the $k$-transporter problem. Before we discuss the techniques used to solve these two problems, we present a brief overview of computational group theory generally.

## 3. Computational Group Theory

Both group theory at large and computational group theory specifically (the study of effective computational algorithms that solve group-theoretic problems) are far too broad to allow detailed presentations in a single journal paper. We ourselves generally refer to Rotman's *An Introduction to the Theory of Groups* (1994) for general information, and to Seress' *Permutation Group Algorithms* (2003) for computational group theory specifically, although there are many excellent texts in both areas. There is also an abbreviated introduction to group theory in ZAP2.

If we cannot substitute for these other references, our goal here is to provide enough general understanding of computational group theory that it will be possible to work through some examples in what follows. With that in mind, there are three basic ideas that we hope to convey:

1. Stabilizer chains. These underlie the fundamental technique whereby large groups are represented efficiently. They also underlie many of the subsequent computations done using those groups.

2. Group decompositions. Given a group $G$ and a subgroup $H < G$, $H$ can be used in a natural way to partition $G$. Each of the partitions can itself be partitioned using a subgroup of $H$, and so on; this gradual refinement underpins many of the search-based group algorithms that have been developed.

3. Lex-leader search. In general, it is possible to establish a lexicographic ordering on the elements of a permutation group; if we are searching for an element of the group having a particular property (as in the $k$-transporter problem), we can assume without loss of generality that we are looking for an element that is minimal under this ordering. This often allows the search to be pruned, since any portion of the search that can be shown not to contain such a minimal element can be eliminated.

### 3.1 Stabilizer Chains

While the fact that a group $G$ can be described in terms of an exponentially smaller number of generators is attractive from a representational point of view, there are many issues that arise if a large set of clauses is represented in this way. Perhaps the most fundamental is that of simple membership: How can we tell if a fixed clause $c'$ is an instance of the augmented clause $(c, G)$?

In general, this is an instance of the 0-transporter problem; we need some $g \in G$ for which $c^g$, the image of $c$ under $g$, does not intersect the complement of $c'$. A simpler but clearly related problem assumes that we have a fixed permutation $g$ such that $c^g = c'$; is $g \in G$ or not? Given a representation of $G$ in terms simply of its generators, it is not obvious how this can be determined quickly.

Of course, if $G$ is represented via a list of all of its elements, we could sort the elements lexicographically and use a binary search to determine if $g$ were included. Virtually any problem of interest to us can be solved in time polynomial in the size of the groups involved, but we would like to do better, solving the problems in time polynomial in the total size of the generators, and therefore generally polynomial in the logarithm of the size of the groups (and so polylog in the size of the original clausal database). We will call a procedure polynomial only if it is indeed polytime in the number of generators of $G$ and in the size of the set of literals on which $G$ acts. It is only for such polynomial procedures that we can be assured that ZAP's representational efficiencies will mature into computational gains.[4]

For the membership problem, that of determining if $g \in G$ given a representation of $G$ in terms of its generators, we need to have a coherent way of understanding the structure of the group $G$ itself. We suppose that $G$ is a subgroup of the group $\mathrm{Sym}(\Omega)$ of symmetries of some set $\Omega$, and we enumerate the elements of $\Omega$ as $\Omega = \{l_1, \ldots, l_n\}$.

There will now be some subset $G^{[2]} \subseteq G$ that fixes $l_1$ in that for any $h \in G^{[2]}$, we have $l_1^h = l_1$. It is easy to see that $G^{[2]}$ is closed under composition, since if any two elements fix $l_1$, then so does their composition. It follows that $G^{[2]}$ is a subgroup of $G$. In fact, we have:

**Definition 3.1** *Given a group $G$ acting on a set $\Omega$ and a subset $L \subseteq \Omega$, the point stabilizer of $L$ is the subgroup $G_L \le G$ of all $g \in G$ such that $l^g = l$ for every $l \in L$. The set stabilizer of $L$ is that subgroup $G_{\{L\}} \le G$ of all $g \in G$ such that $L^g = L$.*

Having defined $G^{[2]}$ as the point stabilizer of $l_1$, we can go on to define $G^{[3]}$ as the point stabilizer of $l_2$ within $G^{[2]}$, so that $G^{[3]}$ is in fact the point stabilizer of $\{l_1, l_2\}$ in $G$. Similarly, we define $G^{[i+1]}$ to be the point stabilizer of $l_i$ in $G^{[i]}$ and thereby construct a chain of stabilizers

$$G = G^{[1]} \ge G^{[2]} \ge \cdots \ge G^{[n]} = \mathbf{1}$$

where the last group is necessarily trivial because once $n-1$ points of $\Omega$ are stabilized, the last point must be also.

If we want to describe $G$ in terms of its generators, we will now assume that we describe all of the $G^{[i]}$ in terms of generators, and furthermore, that the generators for $G^{[i]}$ are a superset of the generators for $G^{[i+1]}$. We can do this because $G^{[i+1]}$ is a subgroup of $G^{[i]}$.

---

4. The development of computationally efficient procedures for solving permutation group problems appears to have begun with Sims' (1970) pioneering work on stabilizer chains.

**Definition 3.2** *A strong generating set $S$ for a group $G \leq \mathrm{Sym}(l_1, \ldots, l_n)$ is a set of generators for $G$ with the property that*

$$\langle S \cap G^{[i]} \rangle = G^{[i]}$$

*for $i = 1, \ldots, n$.*

As usual, $\langle g_i \rangle$ denotes the group generated by the $g_i$.

It is easy to see that a generating set is strong just in case it has the property discussed above, in that each $G^{[i]}$ can be generated incrementally from $G^{[i+1]}$ and the generators that are in fact elements of $G^{[i]} - G^{[i+1]}$.

As an example, suppose that $G = S_4$, the symmetric group on 4 elements (which we denote $1, 2, 3, 4$). Now it is not hard to see that $S_4$ is generated by the 4-cycle $(1, 2, 3, 4)$ and the transposition $(3, 4)$, but this is not a strong generating set. $G^{[2]}$ is the subgroup of $S_4$ that stabilizes 1 (and is therefore isomorphic to $S_3$, since it can randomly permute the remaining three points) but

$$\langle S \cap G^{[2]} \rangle = \langle (3, 4) \rangle = G^{[3]} \neq G^{[2]} \tag{1}$$

If we want a strong generating set, we need to add $(2, 3, 4)$ or a similar permutation to the generating set, so that (1) becomes

$$\langle S \cap G^{[2]} \rangle = \langle (2, 3, 4), (3, 4) \rangle = G^{[2]}$$

Here is a slightly more interesting example. Given a permutation, it is always possible to write that permutation as a composition of transpositions. One possible construction maps 1 where it is supposed to go, then ignores it for the rest of the construction, and so on. Thus we have for example

$$(1, 2, 3, 4) = (1, 2)(1, 3)(1, 4) \tag{2}$$

where the order of composition is from left to right, so that 1 maps to 2 by virtue of the first transposition and is then left unaffected by the other two, and so on.

While the representation of a permutation in terms of transpositions is not unique, the parity of the number of transpositions is; a permutation can always be represented as a product of an even or an odd number of transpositions, but not both. Furthermore, the product of two transposition products of lengths $l_1$ and $l_2$ can obviously be represented as a product of length $l_1 + l_2$, and it follows that the product of two "even" permutations is itself even, and we have:

**Definition 3.3** *The* alternating group of order $n$*, to be denoted by $A_n$, is the subgroup of even permutations of $S_n$.*

What about a strong generating set for $A_n$? If we fix the first $n - 2$ points, then the transposition $(n - 1, n)$ is obviously odd, so we must have $A_n^{[n-1]} = \mathbf{1}$, the trivial group. For any smaller $i$, we can get a subset of $A_n$ by taking the generators for $S_n^{[i]}$ and operating on each as necessary with the transposition $(n - 1, n)$ to make it even. It is not hard to

451

see that an $n$-cycle is odd if and only if $n$ is even (consider (2) above), so given the strong generating set

$$\{(n-1, n), (n-2, n-1, n), \ldots, (2, 3, \ldots, n), (1, 2, \ldots, n)\}$$

for $S_n$, a strong generating set for $A_n$ if $n$ is odd is

$$\{(n-2, n-1, n), (n-1, n)(n-3, n-2, n-1, n), \ldots, (n-1, n)(2, 3, \ldots, n), (1, 2, \ldots, n)\}$$

and if $n$ is even is

$$\{(n-2, n-1, n), (n-1, n)(n-3, n-2, n-1, n), \ldots, (2, 3, \ldots, n), (n-1, n)(1, 2, \ldots, n)\}$$

We can simplify these expressions slightly to get

$$\{(n-2, n-1, n), (n-3, n-2, n-1), \ldots, (2, 3, \ldots, n-1), (1, 2, \ldots, n)\}$$

if $n$ is odd and

$$\{(n-2, n-1, n), (n-3, n-2, n-1), \ldots, (2, 3, \ldots, n), (1, 2, \ldots, n-1)\}$$

if $n$ is even.

Given a strong generating set, it is easy to compute the size of the original group $G$. To do this, we need the following well known definition and result:

**Definition 3.4** *Given groups $H \leq G$ and $g \in G$, we define $Hg$ to be the set of all $hg$ for $h \in H$. For any such $g$, we will say that $Hg$ is a (right) coset of $H$ in $G$.*

**Proposition 3.5** *Let $Hg_1$ and $Hg_2$ be two cosets of $H$ in $G$. Then $|Hg_1| = |Hg_2|$ and the cosets are either identical or disjoint.* ☐

In other words, given a subgroup $H$ of a group $G$, the cosets of $H$ partition $G$. This leads to:

**Definition 3.6** *For groups $H \leq G$, the index of $H$ in $G$, denoted $[G : H]$, is the number of distinct cosets of $H$ in $G$.*

**Corollary 3.7** *For a finite group $G$, $[G : H] = \frac{|G|}{|H|}$.* ☐

Given that the cosets partition the original group $G$, it is natural to think of them as defining an equivalence relation on $G$, where $x \approx y$ if and only if $x$ and $y$ belong to the same coset of $H$. We have:

**Proposition 3.8** *$x \approx y$ if and only if $xy^{-1} \in H$.*

**Proof.** If $xy^{-1} = h \in H$ and $x$ is in a coset $Hg$ so that $x = h'g$ for some $h' \in H$, then $y = h^{-1}x = h^{-1}h'g$ is in the same coset. Conversely, if $x = hg$ and $y = h'g$ are in the same coset, then $xy^{-1} = hgg^{-1}h'^{-1} = hh'^{-1} \in H$. ☐

Many equivalence relations on groups are of this form. Indeed, if $\approx$ is *any* right invariant equivalence relation on the elements of a group $G$ (so that if $x \approx y$, then $xz \approx yz$ for any $z \in G$), then there is some $H \leq G$ such that the cosets of $H$ define the equivalence relation.

Returning to stabilizer chains, recall that we denote by $l_i^{G^{[i]}}$ the orbit of $l_i$ under $G^{[i]}$ (i.e. the set of all points to which $G^{[i]}$ maps $l_i$). We now have:

**Proposition 3.9** *Given a group $G$ acting on a set $\{l_1, \ldots, l_n\}$ and associated stabilizer chain $G^{[1]} \geq \cdots \geq G^{[n]}$,*

$$|G| = \prod_i |l_i^{G^{[i]}}| \tag{3}$$

**Proof.** We know that

$$|G| = \frac{|G|}{|G^{[2]}|}|G^{[2]}| = [G : G^{[2]}]|G^{[2]}|$$

or inductively that

$$|G| = \prod_i [G^{[i]} : G^{[i+1]}]$$

But it is easy to see that the distinct cosets of $G^{[i+1]}$ in $G^{[i]}$ correspond exactly to the points to which $G^{[i]}$ maps $l_i$, so that

$$[G^{[i]} : G^{[i+1]}] = |l_i^{G^{[i]}}|$$

and the result follows. □

Note that the expression in (3) is easy to compute given a strong generating set. As an example, given the strong generating set $\{(1,2,3,4), (2,3,4), (3,4)\}$ for $S_4$, it is clear that $S_4^{[3]} = \langle(3,4)\rangle$ and the orbit of 3 is of size 2. The orbit of 2 in $S_4^{[2]} = \langle(2,3,4),(3,4)\rangle$ is of size 3, and the orbit of 1 in $S_4^{[1]}$ is of size 4. So the total size of the group is $4! = 24$, hardly a surprise.

For $A_4$, a strong generating set is $\{(3,4)(1,2,3,4),(2,3,4)\} = \{(1,2,3),(2,3,4)\}$. The orbit of 2 in $A_4^{[2]} = \langle(2,3,4)\rangle$ is clearly of size 3, and the orbit of 1 in $A_4^{[1]} = A_4$ is of size 4. So $|A_4| = 12$. In general, of course, there are exactly two cosets of the alternating group because all of the odd permutations can be constructed by multiplying the even permutations in $A_n$ by a fixed transposition $t$. Thus $|A_n| = n!/2$.

We can evaluate the size of $A_n$ using strong generators by realizing that the orbit of 1 is of size $n$, that of 2 is of size $n-1$, and so on, until the orbit of $n-2$ is of size 3. The orbit of $n-1$ is of size 1, however, since the transposition $(n-1, n)$ is not in $A_n$. Thus $|A_n| = n!/2$ as before.

We can also use the strong generating set to test membership in the following way. Suppose that we have a group $G$ described in terms of its strong generating set (and therefore its stabilizer chain $G^{[1]} \geq \cdots \geq G^{[n]}$), and a specific permutation $\omega$. Now if $\omega(1) = k$, there are two possibilities:

1. If $k$ is not in the orbit of 1 in $G = G^{[1]}$, then clearly $\omega \notin G$.

2. If $k$ is in the orbit of 1 in $G^{[1]}$, select $g_1 \in G^{[1]}$ with $1^{g_1} = g_1(1) = k$. Now we construct $\omega_1 = \omega g_1^{-1}$, which fixes 1, and we determine recursively if $\omega_1 \in G^{[2]}$.

At the end of the process, we will have stabilized all of the elements moved by $G$, and should have $\omega_{n+1} = 1$. If so, the original $\omega \in G$; if not, $\omega \notin G$. This procedure is known as *sifting*.

Continuing with our example, let us see if the 4-cycle $\omega = (1,2,3,4)$ is in $S_4$ and in $A_4$. For the former, we see that $\omega(1) = 2$ and $(1,2,3,4) \in S_4^{[1]}$. This produces $\omega_1 = 1$, and we can stop and conclude that $\omega \in S_4$ (once again, hardly a surprise).

For the second, we know that $(1, 2, 3) \in A_4^{[1]}$ and we get $\omega_1 = (1, 2, 3, 4)(1, 2, 3)^{-1} = (3, 4)$. Now we could actually stop, since $(3, 4)$ is obviously odd, but let us continue with the procedure. Since 2 is fixed by $\omega_1$, we have $\omega_2 = \omega_1$. Now 3 is moved to 4 by $\omega_2$, but $A_4^{[3]}$ is the trivial group, so we conclude correctly that $(1, 2, 3, 4) \notin A_4$.

## 3.2 Coset Decomposition

Some of the group problems that we will be considering (e.g., the $k$-transporter problem) subsume what was described in ZAP1 as *subsearch* (Dixon et al., 2004b; Ginsberg & Parkes, 2000). Subsearch is known to be NP-hard, so it follows that $k$-transporter must be as well. That suggests that the group-theoretic methods for solving it will involve search in some way.

The search involves a potential examination of all of the instances of some augmented clause $(c, G)$, or, in group theoretic terms, a potential examination of each member of the group $G$. The computational group theory community often approaches such a search problem by gradually decomposing $G$ into smaller and smaller cosets. What we will call a *coset decomposition tree* is produced, where the root of the tree is the entire group $G$ and the leaf nodes are individual elements of $G$:

**Definition 3.10** *Let $G$ be a group, and $G^{[1]} \geq \cdots \geq G^{[n]}$ a stabilizer chain for it. A* coset decomposition tree *for $G$ is a tree whose vertices at the $i$th level are the cosets of $G^{[i]}$ and for which the parent of a particular $G^{[i]}g$ is that coset of $G^{[i-1]}$ that contains it.*

At any particular level $i$, the cosets correspond to the points to which the sequence $\langle l_1, \ldots, l_i \rangle$ can be mapped, with the points in the image of $l_i$ identifying the children of any particular node at level $i - 1$.

As an example, suppose that we consider the augmented clause

$$(a \vee b, \mathrm{Sym}(a, b, c, d)) \tag{4}$$

corresponding to the collection of ground clauses

$$a \vee b$$
$$a \vee c$$
$$a \vee d$$
$$b \vee c$$
$$b \vee d$$
$$c \vee d$$

Suppose also that we are working with an assignment for which $a$ and $b$ are true and $c$ and $d$ are false, and are trying to determine if any instance of (4) is unsatisfied. Assuming that we take $l_1$ to be $a$ through $l_4 = d$, the coset decomposition tree associated with $S_4$ is the following:

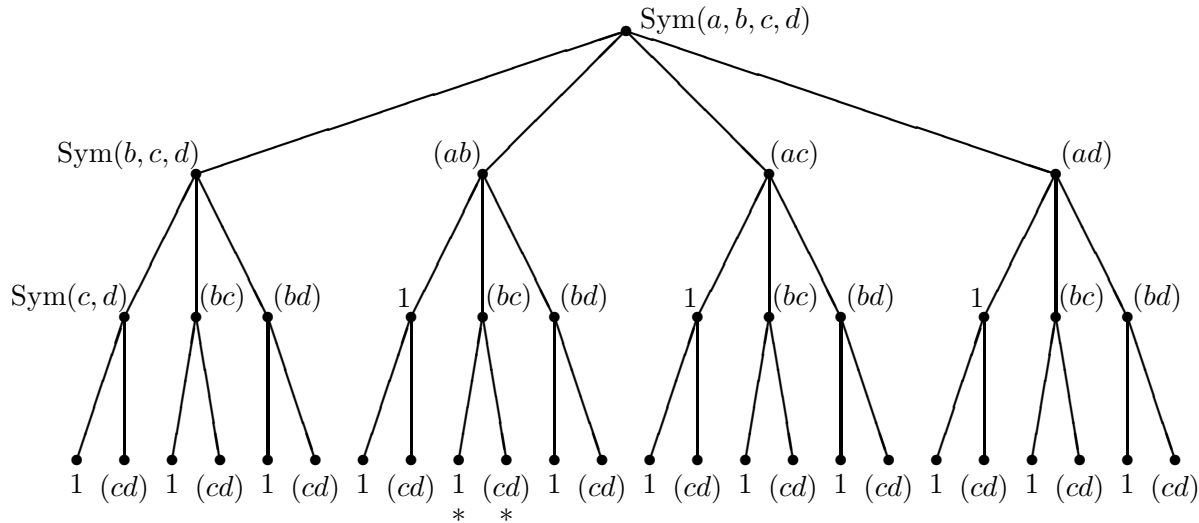

An explanation of the notation here is surely in order. The nodes on the lefthand edge are labeled by the associated groups; for example, the node at level 2 is labeled with $\mathrm{Sym}(b, c, d)$ because this is the point at which we have fixed $a$ but $b$, $c$ and $d$ are still allowed to vary.

As we move across the row, we find representatives of the cosets that are being considered. So moving across the second row, the first entry $(ab)$ means that we are taking the coset of the basic group $\mathrm{Sym}(b, c, d)$ that is obtained by multiplying each element by $(ab)$ on the right. This is the coset that maps $a$ uniformly to $b$.

On the lower rows, we multiply the coset representatives associated with the nodes leading to the root. So the third node in the third row, labeled with $(bd)$, corresponds to the coset $\mathrm{Sym}(c, d) \cdot (bd)$.[5] The two elements of this coset are $(bd)$ and $(cd)(bd) = (bdc)$. The point $b$ is uniformly mapped to $d$, $a$ is fixed, and $c$ can either be fixed or mapped to $b$.

The fourth point on this row corresponds to the coset

$$\mathrm{Sym}(c, d) \cdot (ab) = \{(ab), (cd)(ab)\}$$

The point $a$ is uniformly mapped to $b$, and $b$ is uniformly mapped to $a$. $c$ and $d$ can be swapped or not.

The fifth point is the coset

$$\mathrm{Sym}(c, d) \cdot (bc)(ab) = \mathrm{Sym}(c, d) \cdot (abc) = \{(abc), (abcd)\} \tag{5}$$

$a$ is still uniformly mapped to $b$, and $b$ is now uniformly mapped to $c$. $c$ can be mapped either to $a$ or to $d$.

For the fourth line, the basic group is trivial and the single member of the coset can be obtained by multiplying the coset representatives on the path to the root. Thus the ninth and tenth nodes (marked with asterisks in the tree) correspond to the permutations $(abc)$ and $(abcd)$ respectively, and do indeed partition the coset of (5).

---

5. As here, we will occasionally denote the group multiplication operator explicitly by $\cdot$ to improve the clarity of the typesetting.

Understanding how this structure is used in search is straightforward. At the root, the original augmented clause (4) may indeed have unsatisfiable instances. But when we move to the first child, we know that the image of $a$ is $a$, so that the instance of the clause in question is $a \lor x$ for some $x$. Since $a$ is true for the assignment in question, it follows that the clause must be satisfied. In a similar way, mapping $a$ to $b$ also must produce a satisfied clause. The search space is already reduced to:

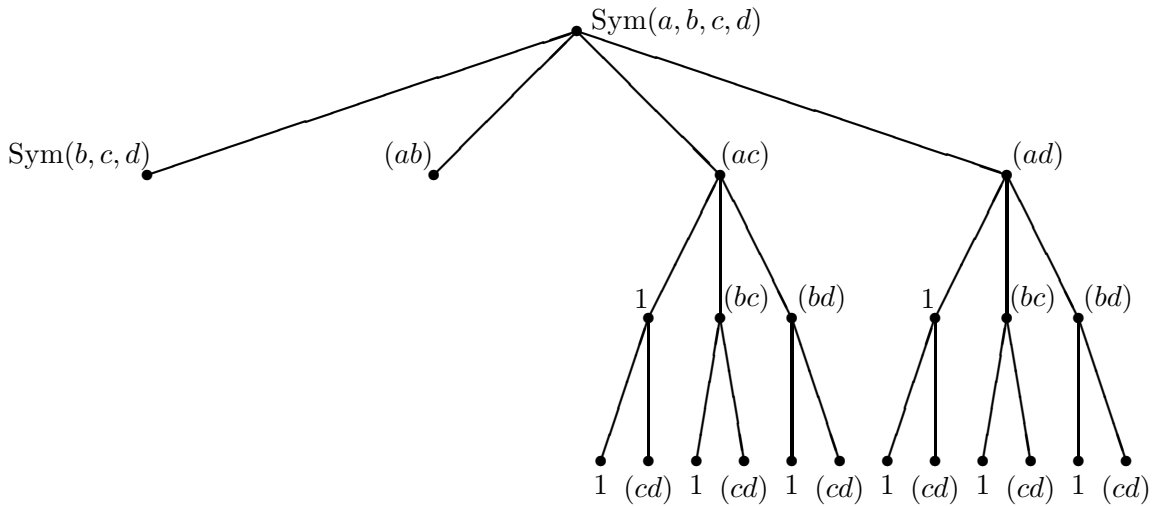

If we map $a$ to $c$, then the first point on the next row corresponds to mapping $b$ to $b$, producing a satisfiable clause. If we map $b$ to $a$ (the next node; $b$ is mapped to $c$ at this node but then $c$ is mapped to $a$ by the permutation $(ac)$ labeling the parent), we also get a satisfiable clause. If we map $b$ to $d$, we will eventually get an unsatisfiable clause, although it is not clear how to recognize that without expanding the two children. The case where $a$ is mapped to $d$ is similar, and the final search tree is:

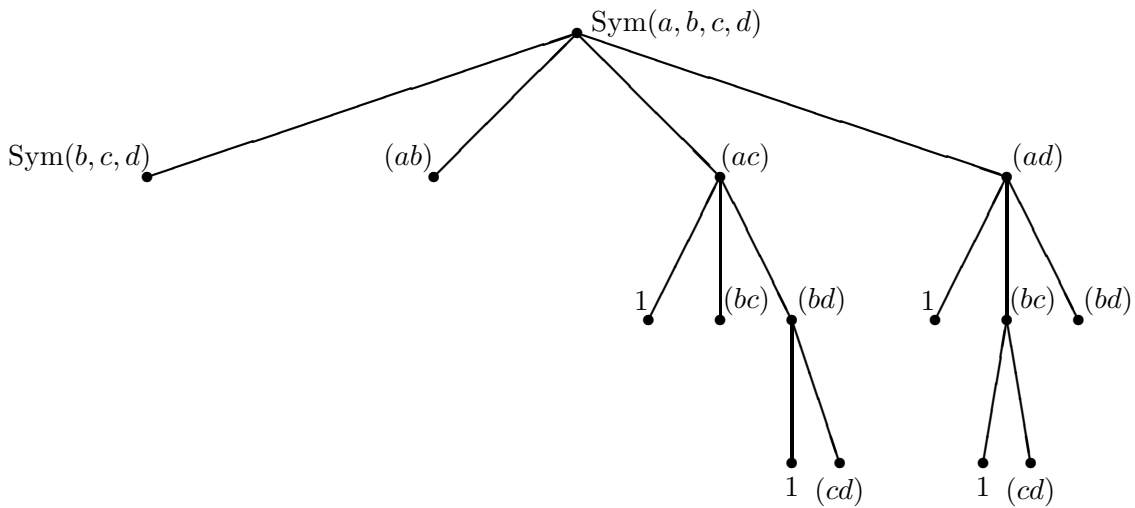

Instead of the six clauses that might need to be examined as instances of the original (4), only four leaf nodes need to be considered. The internal nodes that were pruned above can be pruned without generation, since the only values that need to be considered for $a$ are necessarily $c$ and $d$ (the unsatisfied literals in the theory). At some level, then, the above search space becomes:

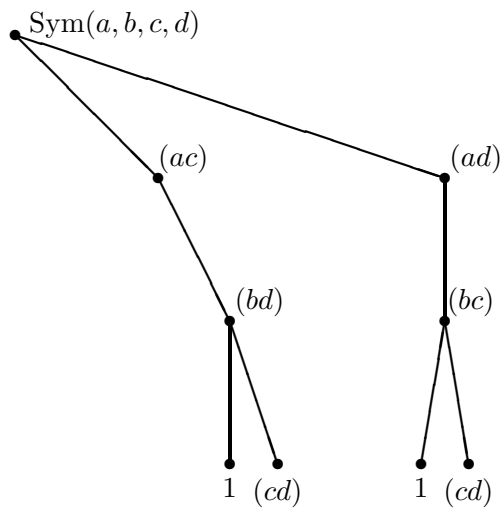

### 3.3 Lex Leaders

Although the remaining search space in this example already examines fewer leaf nodes than the original, there still appears to be some redundancy. To understand one possible simplification, recall that we are searching for a group element $g$ for which $c^g$ is unsatisfied given the current assignment. Since any such group element suffices, we can (if we wish) search for that group element that is smallest under the lexicographic ordering of the group itself:

**Definition 3.11** *Let* $G \leq \mathrm{Sym}(\Omega)$ *be a group, and* $\Omega = \omega_1, \ldots, \omega_n$ *an ordering of the elements of* $\Omega$. *For* $g_1, g_2 \in G$, *we will write* $g_1 < g_2$ *if there is some* $i$ *with* $\omega_j^{g_1} = \omega_j^{g_2}$ *for all* $j < i$ *and* $\omega_i^{g_1} < \omega_i^{g_2}$ .

Since the ordering defined by Definition 3.11 is a total order, we immediately have:

**Lemma 3.12** *If* $S \subseteq \mathrm{Sym}(\Omega)$ *for some ordered set* $\Omega$, *then* $S$ *has a unique minimal element.* □

The minimal element of $S$ is typically called a *lexicographic leader* or *lex leader* of $S$.

In our example, imagine that there were a solution (i.e., a group element corresponding to an unsatisfied instance) under the right hand node at depth three. Now there would necessarily also have been an analogous solution under the preceding node at depth three, since the two search spaces are in some sense identical. The two hypothetical group elements would be identical except the images of $a$ and $b$ would be swapped. Since the group elements under the left hand node precede those under the right hand node in the lexicographic ordering, it follows that the lexicographically least element (which is all that we're looking for) is not under the right hand node, which can therefore be pruned. The search space becomes:

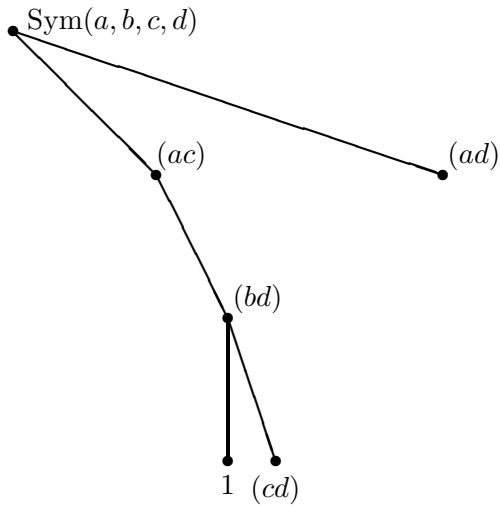

This particular technique is quite general: whenever we are searching for a group element with a particular property, we can restrict our search to lex leaders of the set of all such elements and prune the search space on that basis. Seress (2003) provides a more complete discussion in the context of the problems typically considered by computational group theory; an example in the context of the $k$-transporter problem specifically can be found in Section 5.5.

Finally, we note that the two remaining leaf nodes are equivalent, since they refer to the same instance – once we know the images of $a$ and of $b$, the overall instance is fixed and no further choices are relevant. So assuming that the variables in the problem are ordered so that those in the clause are considered first, we can finally prune the search below depth three to get:

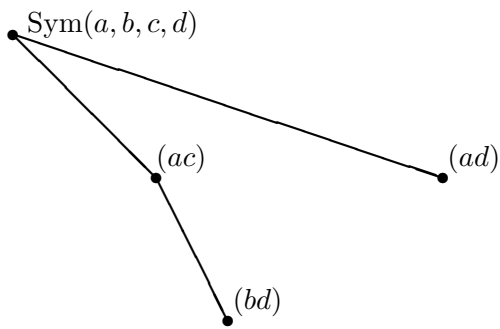

Only a single leaf node need be considered.

Before we return to the application of these ideas in ZAP, we should stress that we have only scratched the surface of computational group theory as a whole. The field is broad and developing rapidly, and the implementation in ZAP is based on ideas that appear in Seress and in the GAP code. Indeed, the name was chosen to reflect ZAP's heritage as an outgrowth of both ZCHAFF and GAP.[6]

## 4. Augmented Resolution

We now turn to our ZAP-specific requirements. First, we have the definition of augmented resolution, which involves computing the group of stable extensions of the groups appearing in the resolvents. Specifically, we have augmented clauses $(c_1, G_1)$ and $(c_2, G_2)$ and need to compute the group $G$ of stable extensions of $G_1$ and $G_2$. Recalling Definition 2.13, this is the group of all permutations $\omega$ with the property that there is some $g_1 \in G_1$ such that

$$\omega|_{c_1^{G_1}} = g_1|_{c_1^{G_1}}$$

and similarly for $g_2$, $G_2$ and $c_2$. We are viewing the clauses $c_i$ as sets, with $c_i^{G_i}$ being the closure of $c_i$ under $G_i$ (recall Definition 2.12).

As an example, consider the two clauses

$$(c_1, G_1) = (a \lor b, \langle (ad), (be), (bf) \rangle)$$

and

$$(c_2, G_2) = (c \lor b, \langle (be), (bg) \rangle)$$

The closure of $c_1$ under $G_1$ is $\{a, b, d, e, f\}$ and $c_2^{G_2} = \{b, c, e, g\}$. We therefore need to find a permutation $\omega$ such that when $\omega$ is restricted to $\{a, b, d, e, f\}$, it is an element of $\langle (ad), (be), (bf) \rangle$, and when restricted to $\{b, c, e, g\}$ is an element of $\langle (be), (bg) \rangle$.

From the second condition, we know that $c$ cannot be moved by $\omega$, and any permutation of $b$, $e$ and $g$ is acceptable because $(be)$ and $(bg)$ generate the symmetric group $\mathrm{Sym}(b, e, g)$. This second restriction does not impact the image of $a$, $d$ or $f$ under $\omega$.

From the first condition, we know that $a$ and $d$ can be swapped or left unchanged, and any permutation of $b$, $e$ and $f$ is acceptable. But recall from the second condition that we must also permute $b$, $e$ and $g$. These conditions combine to imply that we cannot move $f$ or $g$, since to move either would break the condition on the other. We can swap $b$ and $e$ or not, so the group of stable extensions is $\langle (ad), (be) \rangle$, and that is what our construction should return.

**Procedure 4.1** *Given augmented clauses $(c_1, G_1)$ and $(c_2, G_2)$, to compute* `stab`$(c_i, G_i)$:

---

6. The authors of ZCHAFF are Moskewicz, Madigan, Zhao, Zhang and Malik; our selection of only Z to include in our acronym is surely unfair to Moskewicz, Madigan and Malik. ZMAP didn't have quite the same ring to it, however, and we hope that the implicitly excluded authors will accept our apologies for our choice.

1  $\texttt{c\_closure}_1 \leftarrow c_1^{G_1}$, $\texttt{c\_closure}_2 \leftarrow c_2^{G_2}$
2  $\texttt{g\_restrict}_1 \leftarrow G_1|_{\texttt{c\_closure}_1}$, $\texttt{g\_restrict}_2 \leftarrow G_2|_{\texttt{c\_closure}_2}$
3  $C_\cap \leftarrow \texttt{c\_closure}_1 \cap \texttt{c\_closure}_2$
4  $\texttt{g\_stab}_1 \leftarrow \texttt{g\_restrict}_{1\{C_\cap\}}$, $\texttt{g\_stab}_2 \leftarrow \texttt{g\_restrict}_{2\{C_\cap\}}$
5  $\texttt{g\_int} \leftarrow \texttt{g\_stab}_1|_{C_\cap} \cap \texttt{g\_stab}_2|_{C_\cap}$
6  $\{g_i\} \leftarrow \{\text{generators of } \texttt{g\_int}\}$
7  $\{l_{1i}\} \leftarrow \{g_i, \text{lifted to } \texttt{g\_stab}_1\}$, $\{l_{2i}\} \leftarrow \{g_i, \text{lifted to } \texttt{g\_stab}_2\}$
8  $\{l'_{2i}\} \leftarrow \{l_{2i}|_{\texttt{c\_closure}_2 - C_\cap}\}$
9  **return** $\langle \texttt{g\_restrict}_{1C_\cap}, \texttt{g\_restrict}_{2C_\cap}, \{l_{1i} \cdot l'_{2i}\}\rangle$

**Proposition 4.2** *The result returned by Procedure 4.1 is* $\texttt{stab}(c_i, G_i)$.

The proof is in Appendix A; here, we present an example of the computation in use and discuss the computational issues surrounding Procedure 4.1. The example we will use is that with which we began this section, but we modify $G_1$ to be $\langle (ad), (be), (bf), (xy)\rangle$ instead of the earlier $\langle (ad), (be), (bf)\rangle$. The new points $x$ and $y$ don't affect the set of instances in any way, and thus should not affect the resolution computation, either.

1. $\texttt{c\_closure}_i \leftarrow c_1^{G_1}$. This amounts to computing the closures of the $c_i$ under the $G_i$; as described earlier, we have $\texttt{c\_closure}_1 = \{a, b, d, e, f\}$ and $\texttt{c\_closure}_2 = \{b, c, e, g\}$.

2. $\texttt{g\_restrict}_i \leftarrow G_i|_{\texttt{c\_closure}_i}$. Here, we restrict each group to act only on the corresponding $\texttt{c\_closure}_i$. In this example, $\texttt{g\_restrict}_2 = G_2$ but $\texttt{g\_restrict}_1 = \langle (ad), (be), (bf)\rangle$ as the irrelevant points $x$ and $y$ are removed.

   Note that it is not always possible to restrict a group to an arbitrary set; one cannot restrict the permutation $(xy)$ to the set $\{x\}$ because you need to add $y$ as well. But in this case, it is possible to restrict $G_i$ to $\texttt{c\_closure}_i$, since this latter set is closed under the action of the group.

3. $C_\cap \leftarrow \texttt{c\_closure}_1 \cap \texttt{c\_closure}_2$. The construction itself works by considering three separate sets – the intersection of the closures of the two original clauses (where the computation is interesting because the various $\omega$ must agree), and the points in only the closure of $c_1$ or only the closure of $c_2$. The analysis on these latter sets is straightforward; we just need $\omega$ to agree with any element of $G_1$ or $G_2$ on the set in question.

   In this step, we compute the intersection region $C_\cap$. In our example, $C_\cap = \{b, e\}$.

4. $\texttt{g\_stab}_i \leftarrow \texttt{g\_restrict}_{i\{C_\cap\}}$. We find the subgroup of $\texttt{g\_restrict}_i$ that set stabilizes $C_\cap$, in this case the subgroup that set stabilizes the pair $\{b, e\}$. For $\texttt{g\_restrict}_1 = \langle (ad), (be), (bf)\rangle$, this is $\langle (ad), (be)\rangle$ because we can no longer swap $b$ and $f$, while for $\texttt{g\_restrict}_2 = \langle (be), (bg)\rangle$, we get $\texttt{g\_stab}_2 = \langle (be)\rangle$.

5. $\texttt{g\_int} \leftarrow \texttt{g\_stab}_1|_{C_\cap} \cap \texttt{g\_stab}_2|_{C_\cap}$. Since $\omega$ must simultaneously agree with both $G_1$ and $G_2$ when restricted to $C_\cap$ (and thus with $\texttt{g\_restrict}_1$ and $\texttt{g\_restrict}_2$ as well), the restriction of $\omega$ to $C_\cap$ must lie within this intersection. In our example, $\texttt{g\_int} = \langle (be)\rangle$.

6. $\{g_i\} \leftarrow \{$generators of $\texttt{g\_int}\}$. Any element of $\texttt{g\_int}$ will lead to an element of the group of stable extensions provided that we extend it appropriately from $C_{\cap}$ back to the full set $c_1^{G_1} \cup c_2^{G_2}$; this step begins the process of building up these extensions. It suffices to work with just the generators of $\texttt{g\_int}$, and we construct those generators here. We have $\{g_i\} = \{(be)\}$.

7. $\{l_{ki}\} \leftarrow \{g_i, \text{lifted to } \texttt{g\_stab}_k\}$. Our goal is now to build up a permutation on $\texttt{c\_closure}_1 \cup \texttt{c\_closure}_2$ that, when restricted to $C_{\cap}$, matches the generator $g_i$. We do this by lifting $g_i$ separately to $\texttt{c\_closure}_1$ and to $\texttt{c\_closure}_2$. Any such lifting suffices, so we can take (for example)

$$l_{11} = (be)(ad)$$

and

$$l_{21} = (be)$$

In the first case, the inclusion of the swap of $a$ and $d$ is neither precluded nor required; we could just as well have used $l_{11} = (be)$.

8. $\{l'_{2i}\} \leftarrow \{l_{2i}|_{\texttt{c\_closure}_2 - C_{\cap}}\}$. We cannot simply compose $l_{11}$ and $l_{21}$ to get the desired permutation on $\texttt{c\_closure}_1 \cup \texttt{c\_closure}_2$ because the part of the permutations acting on the intersection $\texttt{c\_closure}_1 \cap \texttt{c\_closure}_2$ will have acted twice. In this case, we would get $l_{11} \cdot l_{21} = (ad)$ which no longer captures our freedom to exchange $b$ and $e$.

   We deal with this by restricting $l_{21}$ *away* from $C_{\cap}$ and only then combining with $l_{11}$. In the example, restricting $(be)$ away from $C_{\cap} = \{b, e\}$ produces the trivial permutation $l'_{21} = (\,)$.

9. **Return** $\langle \texttt{g\_restrict}_{1C_{\cap}}, \texttt{g\_restrict}_{2C_{\cap}}, \{l_{1i} \cdot l'_{2i}\} \rangle$. We now compute the final answer from three sources: The combined $l_{1i} \cdot l'_{2i}$ that we have been working to construct, along with elements of $\texttt{g\_restrict}_1$ that fix every point in the closure of $c_2$ and elements of $\texttt{g\_restrict}_2$ that fix every point in the closure of $c_1$. These latter two sets obviously consist of stable extensions. An element of $\texttt{g\_restrict}_1$ point stabilizes the closure of $c_2$ if and only if it point stabilizes the points that are in both the closure of $c_1$ (to which $\texttt{g\_restrict}_1$ has been restricted) and the closure of $c_2$; in other words, if and only if it point stabilizes $C_{\cap}$.

In our example, we have

$$\begin{aligned}
\texttt{g\_restrict}_{1C_{\cap}} &= \langle (ad) \rangle \\
\texttt{g\_restrict}_{2C_{\cap}} &= 1 \\
\{l_{1i} \cdot l'_{2i}\} &= \{(be)(ad)\}
\end{aligned}$$

so that the final group returned is

$$\langle (ad), (be)(ad) \rangle$$

This group is identical to the "obvious"

$$\langle (ad), (be) \rangle$$

461

We can swap either the $(a, d)$ pair or the $(b, e)$ pair, as we see fit. The first swap $(ad)$ is sanctioned for the first "resolvent" $(c_1, G_1) = (a \lor b, \langle (ad), (be), (bf) \rangle)$ and does not mention any relevant variable in the second $(c_2, G_2) = (c \lor b, \langle (be), (bg) \rangle)$. The second swap $(be)$ is sanctioned in both cases.

**Computational issues**    We conclude this section by discussing some of the computational issues that arise when we implement Procedure 4.1, including the complexity of the various operations required.

1.  `c_closure`$_i \leftarrow c_i^{G_i}$. Efficient algorithms exist for computing the closure of a set under a group. The basic method is to use a flood-fill like approach, adding and marking the result of acting on the set with a single generator, and recurring until no new points are added.

2.  `g_restrict`$_i \leftarrow G_i|_{\texttt{c\_closure}_i}$. A group can be restricted to a set that it stabilizes by restricting the generating permutations individually.

3.  $C_\cap \leftarrow$ `c_closure`$_1 \cap$ `c_closure`$_2$. Set intersection is straightforward.

4.  `g_stab`$_i \leftarrow$ `g_restrict`$_{i\{C_\cap\}}$. Set stabilizer is *not* straightforward, and is not known to be polynomial in the total size of the generators of the group being considered (Seress, 2003).[7] The most effective implementations work with a coset decomposition as described in Section 3.2; in computing $G_{\{S\}}$ for some set $S$, a node can be pruned when it maps a point inside of $S$ out of $S$ or vice versa. GAP implements this (but see our comments at the end of Section 10.2).

5.  `g_int` $\leftarrow$ `g_stab`$_1|_{C_\cap} \cap$ `g_stab`$_2|_{C_\cap}$. Group intersection is also not known to be polynomial in the total size of the generators; once again, a coset decomposition is used. Coset decompositions are constructed for each of the groups being combined, and the search spaces are pruned appropriately. GAP implements this as well.

6.  $\{g_i\} \leftarrow \{$generators of `g_int`$\}$. Groups are typically represented in terms of their generators, so reconstructing a list of those generators is trivial. Even if the generators are not known, constructing a strong generating set is known to be polynomial in the number of generators constructed.

7.  $\{l_{ki}\} \leftarrow \{g_i,$ lifted to `g_stab`$_k\}$. Suppose that we have a group $G$ acting on a set $T$, a subset $V \subseteq T$ and a permutation $h$ acting on $V$ such that we know that $h$ is the restriction to $V$ of some $g \in G$, so that $h = g|_V$. To find such a $g$, we first construct a stabilizer chain for $G$ using an ordering that puts the elements of $T - V$ first. Now we are basically looking for a $g \in G$ such that the sifting procedure of Section 3.1 produces $h$ at the point that the points in $T - V$ have all been fixed. We can find such a $g$ in polynomial time by inverting the sifting procedure itself.

8.  $\{l'_{2i}\} \leftarrow \{l_{2i}|_{\texttt{c\_closure}_{2-C_\cap}}\}$. As in line 2, restriction is still easy.

---

7. Unlike the $k$-transporter problem, which was mentioned at the beginning of Section 3.2 to be NP-hard, neither set stabilizer nor group intersection (see step 5) is likely to be NP-hard (Babai & Moran, 1988).

9. **Return** $\langle$`g_restrict`$_{1C_\cap}$, `g_restrict`$_{2C_\cap}$, $\{l_{1i} \cdot l'_{2i}\}\rangle$. Since groups are typically represented by their generators, we need simply take the union of the generators for the three arguments. Point stabilizers (needed for the first two arguments) are straightforward to compute using stabilizer chains.

## 5. Unit Propagation and the (Ir)relevance Test

As we have remarked, the other main computational requirement of an augmented satisfiability engine is the ability to solve the $k$-transporter problem: Given an augmented clause $(c, G)$ where $c$ is once again viewed as a set of literals, and sets $S$ and $U$ of literals and an integer $k$, we want to find a $g \in G$ such that $c^g \cap S = \varnothing$ and $|c^g \cap U| \le k$, if such a $g$ exists.

### 5.1 A Warmup

We begin with a somewhat simpler problem, assuming that $U = \varnothing$ so we are simply looking for a $g$ such that $c^g \cap S = \varnothing$.

We need the following definitions:

**Definition 5.1** *Let $H \le G$ be groups. A* transversal *of $H$ in $G$ is any subset of $G$ that contains one element of each coset of $H$. We will denote such a transversal by $(G : H)$.*

Note that since $H$ itself is one of the cosets, the transversal must contain a (unique) element of $H$. We will generally assume that the identity is this unique element.

**Definition 5.2** *Suppose that $G$ acts on a set $\Omega$ and that $c \subseteq \Omega$. By $c_G$ we will denote the elements of $c$ that are fixed by $G$.*

As the search proceeds, we will gradually fix more and more points of the clause in question. The notation of Definition 5.2 will let us refer easily to the points that have been fixed thus far.

**Procedure 5.3** *Given groups $H \le G$, an element $t \in G$, sets $c$ and $S$, to find a group element $g = $ `map`$(G, H, t, c, S)$ with $g \in H$ and $c^{gt} \cap S = \varnothing$:*

```
1   if c^t_H ∩ S ≠ ∅
2      then return FAILURE
3   if c = c_H
4      then return 1
5   α ← an element of c − c_H
6   for each t' in (H : H_α)
7          do r ← map(G, H_α, t't, c, S)
8             if r ≠ FAILURE
9                then return rt'
10  return FAILURE
```

This is essentially a codification of the example that was presented in Section 3.2. We terminate the search when the clause is fixed by the remaining group $H$, but have not yet included any analog to the lex-leader pruning that we discussed in Section 3.3. In the recursive call in line 7, we retain the original group, for which we will have use in subsequent versions of the procedure.

A more precise description of the procedure would state explicitly that $G$ acts on $c$ and $S$, so that $G \leq \text{Sym}(\Omega)$ with $c, S \subseteq \Omega$. Here and elsewhere, we believe that these conditions are obvious from context and have elected not to clutter the procedural descriptions with them.

**Proposition 5.4** $\text{map}(G, G, 1, c, S)$ *returns an element* $g \in G$ *for which* $c^g \cap S = \varnothing$, *if such an element exists, and returns* FAILURE *otherwise.*

**Proof.** The proof in the Appendix A shows the slightly stronger result that $\text{map}(G, H, t, c, S)$ returns an element $g \in H$ for which $c^{gt} \cap S = \varnothing$ if such an element exists. □

Given that the procedure terminates the search when all elements of $c$ are stabilized by $G$ but does not include lex-leader considerations, the search space examined in the example from Section 3.2 is the following, where we have replaced the variables $a, b, c, d$ with $x_1, x_2, x_3, x_4$ to avoid confusion with our current use of $c$ to represent the clause in question.

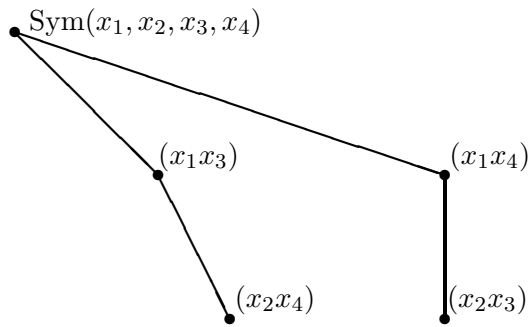

It is still important to prune the node in the lower right, since for a larger problem, this node may be expanded into a significant search subtree. We discuss this pruning in Section 5.5.

In the interests of clarity, let us go through the example explicitly. Recall that the clause $c = x_1 \vee x_2$, $G = \text{Sym}(x_1, x_2, x_3, x_4)$ permutes the $x_i$ arbitrarily, and that $S = \{x_1, x_2\}$.

On the initial pass through the procedure, $c_H = \varnothing$; suppose that we select $x_1$ to stabilize first. Line 6 now selects the point to which $x_1$ should be mapped; if we select $x_1$ or $x_2$, then $x_1$ itself will be mapped into $S$ and the recursive call will fail on line 2. So suppose we pick $x_3$ as the image of $x_1$.

Now $c_H = \{x_1\}$, and we need to fix the image of another point; $x_2$ is all that's left in the original clause $c$. As before, selecting $x_1$ or $x_2$ as the image of $x_2$ leads to failure. $x_3$ is already taken (it's the image of $x_1$), so we have to map $x_2$ into $x_4$. Now every element of $c$ is fixed, and the next recursive call returns the trivial permutation on line 4. This is combined with $(x_2 x_4)$ on line 9 in the caller as we fix $x_4$ as the image of $x_2$. The original invocation then combines with $(x_1 x_3)$ to produce the final answer of $(x_2 x_4)(x_1 x_3)$.

464

## 5.2 The $k$-Transporter Problem

Extending the above algorithm to solve the $k$-transporter problem is straightforward; in addition to requiring that $c_H^t \cap S = \emptyset$ in line 2, we also need to keep track of the number of points that have been (or will be) mapped into the set $U$ and make sure that we won't be forced to exceed the limit $k$.

To understand this, suppose that we are examining a node in the coset decomposition tree labeled with a permutation $t$, so that the node corresponds to keep permutations $gt$ for various $g$ in the subgroup being considered at this level. We want to ensure that there is some $g$ for which $|c^{gt} \cap U| \leq k$. Since $c^{gt}$ is assumed to avoid the set $S$ completely, we can replace this with the slightly stronger

$$|c^{gt} \cap (S \cup U)| \leq k \tag{6}$$

This is in turn equivalent to

$$|c^g \cap (S \cup U)^{t^{-1}}| \leq k \tag{7}$$

since the set in (7) is simply the result of operating on the set in (6) with the permutation $t^{-1}$.

We will present a variety of ways in which the bound of (7) can be approximated; for the moment, we simply introduce an auxiliary function $\mathtt{overlap}(H, c, V)$, which we assume computes a lower bound on $|c^h \cap V|$ for all $h \in H$. Procedure 5.3 becomes:

**Procedure 5.5** *Given groups $H \leq G$, an element $t \in G$, sets $c$, $S$ and $U$ and an integer $k$, to find a group element $g = \mathtt{transport}(G, H, t, c, S, U, k)$ with $g \in H$, $c^{gt} \cap S = \emptyset$ and $|c^{gt} \cap U| \leq k$:*

```
1    if c_H^t ∩ S ≠ ∅
2        then return FAILURE
3    if overlap(H, c, (S ∪ U)^{t^{-1}}) > k
4        then return FAILURE
5    if c = c_H
6        then return 1
7    α ← an element of c − c_H
8    for each t' in (H : H_α)
9        do r ← transport(G, H_α, t't, c, S, U, k)
10           if r ≠ FAILURE
11               then return rt'
12   return FAILURE
```

For convenience, we will denote $\mathtt{transport}(G, G, 1, c, S, U, k)$ by $\mathtt{transport}(G, c, S, U, k)$. This is the "top level" function corresponding to the original invocation of Procedure 5.5.

**Proposition 5.6** *Provided that $|c^h \cap V| \geq \mathtt{overlap}(H, c, V) \geq |c_H \cap V|$ for all $h \in H$, $\mathtt{transport}(G, c, S, U, k)$ as computed by Procedure 5.5 returns an element $g \in G$ for which $c^g \cap S = \emptyset$ and $|c^g \cap U| \leq k$, if such an element exists, and returns FAILURE otherwise.*

The second condition on `overlap` (that $\mathtt{overlap}(H, c, V) \geq |c_H \cap V|$) is needed to ensure that the procedure terminates on line 4 once the overlap limit is reached, rather than succeeding on line 6.

Procedure 5.5 is simplified significantly by the fact that we only need to return a single $g$ with the desired properties, as opposed to all such $g$. In the examples arising in (ir)relevance calculations, a single answer suffices. But if we want to compute the unit consequences of a given literal, we need all of the unit instances of the clause in question. There are other considerations at work in this case, however, and we defer discussion of this topic until Section 6.

Our initial version of `overlap` is:

**Procedure 5.7** *Given a group $H$, and two sets $c, V$, to compute* $\mathtt{overlap}(H, c, V)$, *a lower bound on the overlap of $c^h$ and $V$ for any $h \in H$:*

1   **return** $|c_H \cap V|$

Having defined `overlap`, we may as well use it to replace the test in line 1 of Procedure 5.5 with a check to see if $\mathtt{overlap}(H, c, S^{t^{-1}}) > 0$, indicating that for any $h \in H$, $|c^h \cap S^{t^{-1}}| > 0$ or, equivalently, that $c^{ht} \cap S \neq \emptyset$. For the simple version of `overlap` defined above, there is no difference between the two procedures. But as `overlap` matures, this change will lead to additional pruning in some cases.

### 5.3 Orbit Pruning

There are two general ways in which nodes can be pruned in the $k$-transporter problem. Lexicographic pruning is a bit more difficult, so we defer it until Section 5.5. To understand the other, we begin with the following example.

Consider the clause $c = x_1 \lor x_2 \lor x_3$ and the group $G$ that permutes the variables $\{x_1, x_2, x_3, x_4, x_5, x_6\}$ arbitrarily. If $S = \{x_1, x_2, x_3, x_4\}$, is there a $g \in G$ with $c^g \cap S = \emptyset$?

Clearly not; there isn't enough "room" because the image of $c$ will be of size three, and there is no way that this 3-element set can avoid the 4-element set $S$ in the 6-element universe $\{x_1, x_2, x_3, x_4, x_5, x_6\}$.

We can do a bit better in many cases. Suppose that our group $G$ is $\langle (x_1 x_4), (x_2 x_5), (x_3 x_6) \rangle$ so that we can swap $x_1$ with $x_4$ (or not), $x_2$ with $x_5$, or $x_3$ with $x_6$. Now if $S = \{x_1, x_4\}$, can we find a $g \in G$ with $c^g \cap S = \emptyset$?

Once again, the answer is clearly no. The orbit of $x_1$ in $G$ is $\{x_1, x_4\}$ and since $\{x_1, x_4\} \subseteq S$, $x_1$'s image cannot avoid the set $S$.

In the general case appearing in Procedure 5.5, consider the initial call, where $t$ is the identity permutation. Given the group $G$, consider the orbits of the points in $c$. If there is any such orbit $W$ for which $|W \cap c| > |W - S|$, we can prune the search. The reason is that each of the points in $W \cap c$ must remain in $W$ when acted on by any element of $G$; that is what the definition of an orbit requires. But there are too many points in $W \cap c$ to stay away from $S$, so we will not manage to have $c^g \cap S = \emptyset$.

What about the more general case, where $t \neq 1$ necessarily? For a fixed $\alpha$ in our clause $c$, we will construct the image $\alpha^{gt}$, acting on $\alpha$ first with $g$ and then with $t$. We are interested in whether $\alpha^{gt} \in S$ or, equivalently, if $\alpha^g \in S^{t^{-1}}$. Now $\alpha^g$ is necessarily in the same orbit as $\alpha$, so we can prune if

$$|W \cap c| > |W - S^{t^{-1}}|$$

For similar reasons, we can also prune if

$$|W \cap c| > |W - U^{t^{-1}}| + k$$

In fact, we can prune if

$$|W \cap c| > |W - (S \cup U)^{t^{-1}}| + k$$

because there still is not enough space to fit the image without either intersecting $S$ or putting at least $k$ points into $U$.

We can do better still. As we have seen, for any particular orbit, the number of points that will eventually be mapped into $U$ is at least

$$|W \cap c| - |W - (S \cup U)^{t^{-1}}|$$

In some cases, this expression will be negative; the number of points that will be mapped into $U$ is therefore at least

$$\max(|W \cap c| - |W - (S \cup U)^{t^{-1}}|, 0)$$

and we can prune any node for which

$$\sum_W \max(|W \cap c| - |W - (S \cup U)^{t^{-1}}|, 0) > k \tag{8}$$

where the sum is over the orbits of the group.

It will be somewhat more convenient to rewrite this using the fact that

$$|W \cap c| + |W - c| = |W| = |W \cap (S \cup U)^{t^{-1}}| + |W - (S \cup U)^{t^{-1}}|$$

so that (8) becomes

$$\sum_W \max(|W \cap (S \cup U)^{t^{-1}}| - |W - c|, 0) > k \tag{9}$$

Incorporating this type of analysis into Procedure 5.7 gives:

**Procedure 5.8** *Given a group $H$, and two sets $c, V$, to compute* `overlap`$(H, c, V)$*, a lower bound on the overlap of $c^h$ and $V$ for any $h \in H$:*

1   $m \leftarrow 0$
2   **for** each orbit $W$ of $H$
3        **do** $m \leftarrow m + \max(|W \cap V| - |W - c|, 0)$
4   **return** $m$

**Proposition 5.9** *Let $H$ be a group and $c, V$ sets acted on by $H$. Then for any $h \in H$, $|c^h \cap V| \geq$ `overlap`$(H, c, V) \geq |c_H \cap V|$ where `overlap` is computed by Procedure 5.8.*

467

## 5.4 Block Pruning

The pruning described in the previous section can be improved further. To see why, consider the following example, which might arise in solving an instance of the pigeonhole problem. We have the two cardinality constraints:

$$x_1 + x_2 + x_3 + x_4 \geq 2 \tag{10}$$

$$x_5 + x_6 + x_7 + x_8 \geq 2 \tag{11}$$

presumably saying that at least two of four pigeons are not in hole $m$ and at least two are not in hole $n$ for some $m$ and $n$.[8] Rewriting the individual cardinality constraints as augmented clauses produces

$$(x_1 \lor x_2 \lor x_3, \mathrm{Sym}(x_1, x_2, x_3, x_4))$$

$$(x_5 \lor x_6 \lor x_7, \mathrm{Sym}(x_5, x_6, x_7, x_8))$$

or, in terms of generators,

$$(x_1 \lor x_2 \lor x_3, \langle (x_1 x_2), (x_2 x_3 x_4) \rangle) \tag{12}$$

$$(x_5 \lor x_6 \lor x_7, \langle (x_5 x_6), (x_6 x_7 x_8) \rangle) \tag{13}$$

What we would really like to do, however, is to capture the full symmetry in a single axiom.

We can do this by realizing that we can obtain (13) from (12) by switching $x_1$ and $x_5$, $x_2$ and $x_6$, and $x_3$ and $x_7$ (in which case we want to switch $x_4$ and $x_8$ as well). So we add the generator $(x_1 x_5)(x_2 x_6)(x_3 x_7)(x_4 x_8)$ to the overall group, and modify the permutations $(x_1 x_2)$ and $(x_2 x_3 x_4)$ (which generate $\mathrm{Sym}(x_1, x_2, x_3, x_4)$) so that they permute $x_5, x_6, x_7, x_8$ appropriately as well. The single augmented clause that we obtain is

$$(x_1 \lor x_2 \lor x_3, \langle (x_1 x_2)(x_5 x_6), (x_2 x_3 x_4)(x_6 x_7 x_8), (x_1 x_5)(x_2 x_6)(x_3 x_7)(x_4 x_8) \rangle) \tag{14}$$

and it is not hard to see that this does indeed capture both (12) and (13).

Now suppose that $x_1$ and $x_5$ are false, and the other variables are unvalued. Does (14) have a unit instance?

With regard to the pruning condition in the previous section, the group has a single orbit, and the condition (with $t = 1$) is

$$|W \cap (S \cup U)| - |W - c| > 1 \tag{15}$$

But

$$
\begin{aligned}
W &= \{x_1, x_2, x_3, x_4, x_5, x_6, x_7, x_8\} \\
S &= \emptyset \\
U &= \{x_2, x_3, x_4, x_6, x_7, x_8\} \\
c &= \{x_1, x_2, x_3\}
\end{aligned}
$$

---

8. In an actual pigeonhole instance, all of the variables would be negated. We have dropped the negations for convenience.

so that $|W \cap (S \cup U)| = 6$, $|W - c| = 5$ and (15) fails.

But it *should* be possible to conclude immediately that there are no unit instances of (14). After all, there are no unit instances of (10) or (11) because only one variable in each clause has been set, and three unvalued variables remain. Equivalently, there is no unit instance of (12) because only one of $\{x_1, x_2, x_3, x_4\}$ has been valued, and two need to be valued to make $x_1 \vee x_2 \vee x_3$ or another instance unit. Similarly, there is no unit instance of (13). What went wrong?

What went wrong is that the pruning heuristic thinks that both $x_1$ and $x_5$ can be mapped to the same clause instance, in which case it is indeed possible that the instance in question be unit. The heuristic doesn't realize that $x_1$ and $x_5$ are in separate "blocks" under the action of the group in question.

To formalize this, let us first make the following definition:

**Definition 5.10** *Suppose $G$ acts on a set $T$. We will say that $G$ acts* transitively *on $T$ if $T$ is an orbit of $G$.*

Put somewhat differently, $G$ acts transitively on $T$ just in case for any $x, y \in T$ there is some $g \in G$ such that $x^g = y$.

**Definition 5.11** *Suppose that a group $G$ acts transitively on a set $T$. Then a* block system *for $G$ is a partitioning of $T$ into sets $B_1, \ldots, B_n$ such that $G$ permutes the $B_i$.*

In other words, for each $g \in G$ and each block $B_i$, $B_i^g = B_j$ for some $j$. If $j = i$, then the image of $B_i$ under $g$ is $B_i$ itself. If $j \neq i$, then the image of $B_i$ under $g$ is disjoint from $B_i$, since the blocks partition $T$.

Every group acting transitively and nontrivially on a set $T$ has at least two block systems:

**Definition 5.12** *For a group $G$ acting transitively on a set $T$, a block system $B_1, \ldots, B_n$ will be called* trivial *if either $n = 1$ or $n = |T|$.*

In the former case, there is a single block consisting of the entire set $T$ (which obviously is a block system). If $n = |T|$, each point is in its own block; since $G$ permutes the points, it obviously permutes the blocks.

**Lemma 5.13** *All of the blocks in a block system are of identical size.* ◻

In the example we have been considering, $B_1 = \{x_1, x_2, x_3, x_4\}$ and $B_2 = \{x_5, x_6, x_7, x_8\}$ is also a block system for the action of the group on the set $T = \{x_1, x_2, x_3, x_4, x_5, x_6, x_7, x_8\}$. And while it is conceivable that a clause image is unit within the overall set $T$, it is impossible for it to have fewer than two unvalued literals within each particular block. Instead of looking at the overall expression

$$|W \cap (S \cup U)| - |W - c| > 1 \tag{16}$$

we can work with individual blocks.

The clause $x_1 \vee x_2 \vee x_3$ is in a single block in this block system, and will therefore remain in a single block after being acted on with any $g \in G$. If the clause winds up in block $B_i$, then the condition (16) can be replaced with

$$|B_i \cap (S \cup U)| - |B_i - c| > 1$$

or, in this case,

$$|B_i \cap (S \cup U)| > |B_i - c| + 1 = 2$$

so that we can prune if there are more than two unvalued literals in the block in question. After all, if there are three or more unvalued literals, there must be at least two in the clause instance being considered, and it cannot be unit.

Of course, we don't know exactly which block will eventually contain the image of $c$, but we can still prune if

$$\min(|B_i \cap (S \cup U)|) > 2$$

since in this case *any* target block will generate a prune. And in the example that we have been considering,

$$|B_i \cap (S \cup U)| = 3$$

for each block in the block system.

Generalizing this idea is straightforward. For notational convenience, we introduce:

**Definition 5.14** *Let $T = \{T_1, \ldots, T_k\}$ be sets, and suppose that $T_{i_1}, \ldots, T_{i_n}$ are the $n$ elements of $T$ of smallest size. Then we will denote $\sum_{j=1}^{n} |T_{i_j}|$ by $\Sigma_{i \leq n}^{\min} T_i$.*

**Proposition 5.15** *Let $G$ be a group acting transitively on a set $T$, and let $c, V \subseteq T$. Suppose also that $\{B_1, \ldots, B_k\}$ is a block system for $G$ and that $c \cap B_i \neq \emptyset$ for $n$ of the blocks in $\{B_1, \ldots, B_k\}$. Then if $b$ is the size of an individual block $B_i$ and $g \in G$,*

$$|c^g \cap V| \geq |c| + \Sigma_{i \leq n}^{\min} (B_i \cap V) - nb \tag{17}$$

**Proposition 5.16** *If the block system is trivial (in either sense), (17) is equivalent to*

$$|c^g \cap V| \geq |T \cap V| - |T - c| \tag{18}$$

**Proposition 5.17** *Let $\{B_1, \ldots, B_k\}$ be a block system for a group $G$ acting transitively on a set $T$. Then (17) is never weaker than (18).*

In any event, we have shown that we can strengthen Procedure 5.8 to:

**Procedure 5.18** *Given a group $H$, and two sets $c, V$, to compute* `overlap`$(H, c, V)$, *a lower bound on the overlap of $c^h$ and $V$ for any $h \in H$:*

```
1   m ← 0
2   for each orbit W of H
3       do {B_1, ..., B_k} ← a block system for W under H
4          n = |{i|B_i ∩ c ≠ Ø}|
5          m ← m + max(|c ∩ W| + Σ^{min}_{i≤n} (B_i ∩ V) − n|B_1|, 0)
6   return m
```

Which block system should we use in line 3 of the procedure? There seems to be no general best answer to this question, although we have seen from Proposition 5.17 that any block system is better than one of the trivial ones. In practice, the best choice appears to be a minimal block system (i.e., one with blocks of the smallest size) for which $c$ is contained within a single block. Now Procedure 5.18 becomes:

**Procedure 5.19** *Given a group $H$, and two sets $c, V$, to compute* `overlap(H, c, V)`, *a lower bound on the overlap of $c^h$ and $V$ for any $h \in H$:*

```
1   m ← 0
2   for each orbit W of H
3       do {B_1, ..., B_k} ← a minimal block system for W under H for which
              c ∩ W ⊆ B_i for some i
4          m ← m + max(|c ∩ W| + min(B_i ∩ V) − |B_1|, 0)
5   return m
```

**Proposition 5.20** *Let $H$ be a group and $c, V$ sets acted on by $H$. Then for any $h \in H$, $|c^h \cap V| \geq$* `overlap(H, c, V)` $\geq |c_H \cap V|$ *where* `overlap` *is computed by Procedure 5.19.*  □

Note that the block system being used depends only on the group $H$ and the original clause $c$. This means that in an implementation it is possible to compute these block systems once and then use them even if there are changes in the sets $S$ and $U$ of satisfied and unvalued literals respectively.

GAP includes algorithms for finding minimal block systems for which a given set of elements (called a "seed" in GAP) is contained within a single block. The basic idea is to form an initial block "system" where the points in the seed are in one block and each point outside of the seed is in a block of its own. The algorithm then repeatedly runs through the generators of the group, seeing if any generator $g$ maps elements $x, y$ in one block to $x^g$ and $y^g$ that are in different blocks. If this happens, the blocks containing $x^g$ and $y^g$ are merged. This continues until every generator respects the candidate block system, at which point the procedure is complete.[9]

### 5.5 Lexicographic Pruning

Block pruning will not help us with the example at the end of Section 5.1. The final space being searched is:

---

9. A faster implementation makes use of the procedure designed for testing equivalence of finite automata (Aho, Hopcroft, & Ullman, 1974, chapter 4) and takes $O(snA(n))$ time, where $s$ is the size of the generating set and $A(n)$ is the inverse Ackerman function.

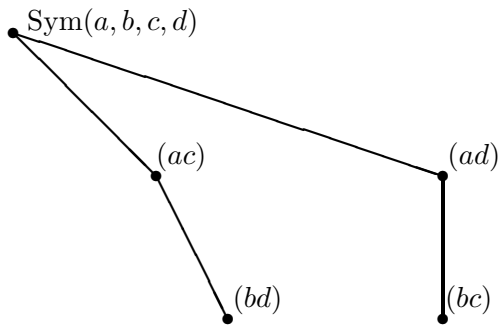

As we have remarked, the first leaf node (where $a$ is mapped to $c$ and $b$ to $d$) is essentially identical to the second (where $a$ is mapped to $d$ and $b$ to $c$). It is important not to expand both since more complicated examples may involve a substantial amount of search below the nodes that are leaf nodes in the above figure.

This is the sort of situation in which lexicographic pruning can generally be applied. We want to identify the two leaf nodes as equivalent in some way, and then expand only the lexicographically least member of each equivalence class. For any particular node $n$, we need a computationally effective way of determining if $n$ is the lexicographically least member of its equivalence class.

We begin by identifying conditions under which two nodes are equivalent. To understand this, recall that we are interested in the image of the clause $c$ under a particular group element $g$. That means that we don't care about where any particular literal $l$ is mapped, because we care only about the image of the entire clause $c$. We also don't care about the image of any literal that isn't in $c$.

From a formal point of view, we begin by extending our set stabilizer notation somewhat:

**Definition 5.21** *For a permutation group $G$ and sets $S_1, \ldots, S_k$ acted on by $G$, by $G_{\{S_1,\ldots,S_k\}}$ we will mean that subgroup of $G$ that simultaneously set stabilizes each of the $S_i$; equivalently, $G_{\{S_1,\ldots,S_k\}} = \cap_i G_{\{S_i\}}$.*

In computing a multiset stabilizer $G_{\{S_1,\ldots,S_k\}} = \cap_i G_{\{S_i\}}$, we need not compute the individual set stabilizers and then take their intersection. Instead, recall that the set stabilizers themselves are computed using coset decomposition; if any stabilized point is moved either into or out of the set in question, the given node can be pruned in the set stabilizer computation. It is straightforward to modify the set stabilizer algorithm so that if any stabilized point is moved into or out of any of the $S_i$, the node in question is pruned. This allows $G_{\{S_1,\ldots,S_k\}}$ to be computed in a single traversal of $G$'s decomposition tree.

Now suppose that $j$ is a permutation in $G$ that stabilizes the set $c$. If $c^g$ satisfies the conditions of the transporter problem, then so will $c^{jg}$. After all, acting with $j$ first doesn't affect the set corresponding to $c$, and the image of the clause under $jg$ is therefore identical to its image under $g$. This means that two permutations $g$ and $h$ are "equivalent" if $h = jg$ for some $j \in G_{\{c\}}$, the set stabilizer of $c$ in $G$. Alternatively, the permutation $g$ is equivalent to any element of the coset $Jg$, where $J = G_{\{c\}}$.

On the other hand, suppose that $k$ is a permutation that simultaneously stabilizes the sets $S$ and $U$ of satisfied and unvalued literals respectively. Now it is possible to show that if we operate with $k$ *after* operating successfully with $g$, we also don't impact the question of whether or not $c^g$ is a solution to the transporter problem. The upshot of this is the following:

**Definition 5.22** *Let $G$ be a group with $J \leq G$ and $K \leq G$, and let $g \in G$. Then the* double coset $JgK$ *is the set of all elements of $G$ of the form $jgk$ for $j \in J$ and $k \in K$.*

**Proposition 5.23** *Let $G$ be a group of permutations, and $c$ a set acted on by $G$. Suppose also that $S$ and $U$ are sets acted on by $G$. Then for any instance $I$ of the $k$-transporter problem and any $g \in G$, either every element of $G_{\{c\}} g G_{\{S,U\}}$ is a solution of $I$, or none is.*

To understand why this is important, imagine that we prune the overall search tree so that the only permutations $g$ remaining are ones that are minimal in their double cosets $JgK$, where $J = G_{\{c\}}$ and $K = G_{\{S,U\}}$ as above. Will this impact the solubility of any instance of the $k$-transporter problem?

It will not. If a particular instance has no solutions, pruning the tree obviously will not introduce any. If the particular instance has a solution $g$, then every element of $JgK$ is also a solution, so specifically the minimal element of $JgK$ is a solution, and this minimal element will not be pruned under our assumptions.

We see, then, that we can prune any node $n$ for which we can show that every permutation $g$ underneath $n$ is *not* minimal in its double coset $JgK$. To state precise conditions under which this lets us prune the node $n$, suppose that we have some coset decomposition of a group $G$, and that $x_j$ is the point fixed at depth $j$ of the tree. Now if $n$ is a node at depth $i$ in the tree, we know that $n$ corresponds to a coset $Ht$ of $G$, where $H$ stabilizes each $x_j$ for $j \leq i$. We will denote the image of $x_j$ under $t$ by $z_j$. If there is no $g \in Ht$ that is minimal in its double coset $JgK$ for $J = G_{\{c\}}$ and $K = G_{\{S,U\}}$ as in Proposition 5.23, then the node $n$ corresponding to $Ht$ can be pruned.

**Lemma 5.24 (Leon, 1991)** *If $x_l \in x_k^{J_{x_1, \ldots, x_{k-1}}}$ for some $k \leq l$ and $z_k > \min(z_l^{K_{z_1, z_2, \ldots, z_{k-1}}})$, then no $g \in Ht$ is the first element of $JgK$.* ☐

**Lemma 5.25 (reported by Seress, 2003)** *Let $s$ be the length of the orbit $x_l^{J_{x_1, \ldots, x_{l-1}}}$. If $z_l$ is among the last $s - 1$ elements of its orbit in $G_{z_1, z_2, \ldots, z_{l-1}}$, then no $g \in Ht$ is the first element of $JgK$.* ☐

Both of these results give conditions under which a node in the coset decomposition can be pruned when searching for a solution to an instance of the $k$-transporter problem. Let us consider an example of each.

We begin with Lemma 5.24. If we return to our example from the end of Section 5.1, we have $G = \mathrm{Sym}(a, b, c, d)$, $c = \{a, b\} = S$, and $U = \emptyset$. Thus $J = K = G_{\{a,b\}} = \mathrm{Sym}(a, b) \times \mathrm{Sym}(c, d) = \langle (ab), (cd) \rangle$.

Consider the node that we have repeatedly remarked can be pruned at depth 1, where we fix the image of $a$ to be $d$. In this case, $x_1 = a$ and $z_1 = d$. If we take $k = l$ in the statement of the lemma, $x_l \in x_l^{J_{x_1, \ldots, x_{l-1}}}$ since $1 \in J_{x_1, \ldots, x_{l-1}}$. Thus we can prune if

$$z_l > \min(z_l^{K_{z_1, z_2, \ldots, z_{l-1}}})$$

473

Further restricting to $l = 1$ gives us

$$z_1 > \min(z_1^K) \tag{19}$$

In this example, $z_1 = d$, so $z_1^K = \{c, d\}$ and (19) holds (assuming that $d > c$ in our ordering). The node can be pruned, and we finally get the reduced search space:

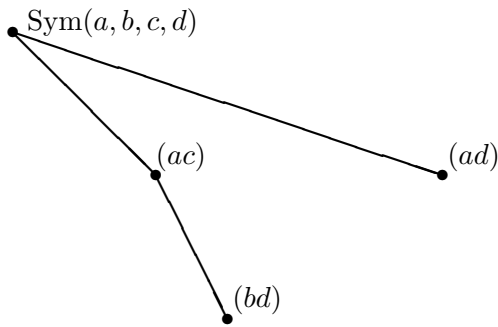

as desired.

This node can be pruned by Lemma 5.25 as well. The conditions of the lemma require that we take $s$ to be the length of the orbit of $a$ under $J$ (since $l = 1$ here), so $s = |\{a, b\}| = 2$. Thus the image of $a$ cannot be among the last $2 - 1 = 1$ points in $a$'s orbit under $G$. Since the orbit of $a$ under $G$ is $\{a, b, c, d\}$, we can once again prune this node. (The previous node, which maps $a$ to $c$, cannot be pruned, of course.)

This particular example is simple. The nodes being examined are at depth one, and there is significant overlap in the groups in question. While the same node is pruned by either lemma here, the lemmas prune different nodes in more complex cases. Note also that the groups $J = G_{\{c\}}$ and $K = G_{\{S, U\}}$ can be computed at the root of the tree, and the group $J$ is independent of the sets $S$ and $U$ and can therefore be cached with the augmented clause $(c, G)$.

Lemmas 5.24 and 5.25 are both well known results in the computational group theory community. We will also have use of the following:

**Lemma 5.26** *Suppose that $t$ is the permutation labeling some node $Ht$ of a coset decomposition tree at depth $k$, so that $x_i^t = z_i$ for $i \leq k$ and $H = G_{x_1,\dots,x_k}$ is the residual group at this level. Let $M$ be the set of points moved by $G_{x_1,\dots,x_k}$. Now if $z_i > \min\left(x_i^{J_{M,x_1,\dots,x_{i-1}}t}\right)$ for any $i \leq k$, then no $g \in Ht$ is the first element of $JgK$.*

As an example, consider the cardinality constraint

$$x_1 + \cdots + x_m \geq n$$

corresponding to the augmented clause $(c, G)$ with

$$c = x_1 \vee \cdots \vee x_{m-n+1}$$

and $G = \text{Sym}(X)$, where $X$ is the set of all of the $x_i$.

Suppose that we fix the images of the $x_i$ in order, and that we are considering a node where the image of $x_1$ is fixed to $z_1$ and the image of $x_2$ is fixed to $z_2$, with $z_2 < z_1$. Now $J = G_{\{c\}} = \mathrm{Sym}(x_1, \ldots, x_{m-n+1}) \times \mathrm{Sym}(x_{m-n+2}, \ldots, x_m)$, so taking $i = 1$ and $k = 2$ in Lemma 5.26 gives us $J_{x_{k+1}, \ldots, x_m} = \mathrm{Sym}(x_1, x_2)$ since we need to fix all of the $x_j$ after $x_2$. But $x_1^{J_{x_{k+1}, \ldots, x_m} t} = \{z_1, z_2\}$, and since $z_1$ is not the smallest element of this set, this is enough to prune this node. See the proof of Proposition 6.9 for another example.

We will refer to Lemmas 5.24–5.26 as the *pruning lemmas*.

Adding lexicographic pruning to our $k$-transporter procedure gives us:

**Procedure 5.27** *Given groups $H \leq G$, an element $t \in G$, sets $c$, $S$ and $U$ and an integer $k$, to find a group element $g = \mathtt{transport}(G, H, t, c, S, U, k)$ with $g \in H$, $c^{gt} \cap S = \emptyset$ and $|c^{gt} \cap U| \leq k$:*

```
 1  if overlap(H, c, S^{t⁻¹}) > 0
 2      then return FAILURE
 3  if overlap(H, c, (S ∪ U)^{t⁻¹}) > k
 4      then return FAILURE
 5  if c = c_H
 6      then return 1
 7  if a pruning lemma can be applied
 8      then return FAILURE
 9  α ← an element of c − c_H
10  for each t′ in (H : H_α)
11      do r ← transport(G, H_α, t′t, c, S, U, k)
12          if r ≠ FAILURE
13              then return rt′
14  return FAILURE
```

Note that the test in line 7 requires access to the groups $J$ and $K$, and therefore to the original group $G$ with which the procedure was called. This is why we retain a copy of this group in the recursive call on line 11.

It might seem that we have brought too much mathematical power to bear on the $k$-transporter problem specifically, but we disagree; recall Figure 1, repeated from ZAP1. High-performance satisfiability engines, running on difficult problems, spend in excess of 90% of their CPU time in unit propagation, which we have seen to be an instance of the $k$-transporter problem. Effort spent on improving the efficiency of Procedure 5.27 (and its predecessors) can be expected to lead to substantial performance improvements in any practical application. See also Figure 8 and the experimental results in Section 9.2.

We do, however, note that while lexicographic pruning is important, it is also expensive. This is why we defer it to line 7 of Procedure 5.27. An earlier lexicographic prune would be independent of the $S$ and $U$ sets, but the count-based pruning is so much faster that we defer the lexicographic check to the extent possible.

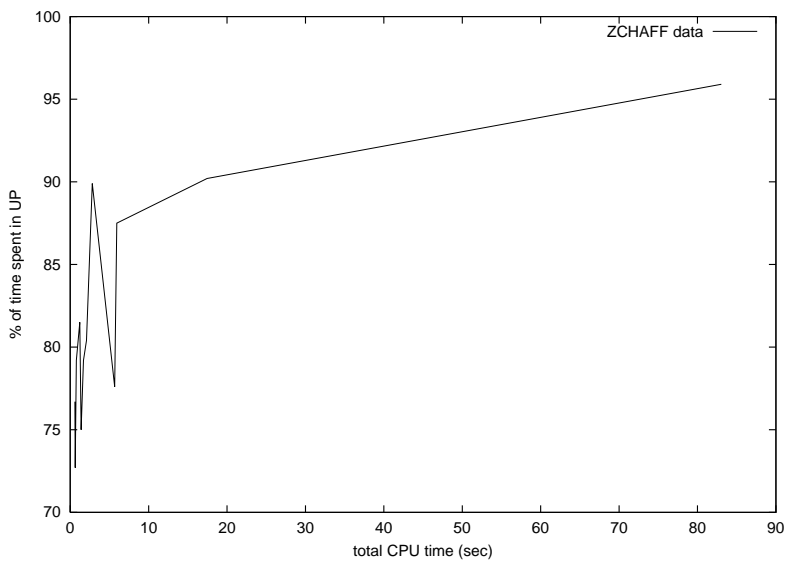

Figure 1: Fraction of CPU time spent in unit propagation

## 6. Unit Propagation

Procedure 5.27 was designed around the need to find a single permutation $g \in G$ satisfying the conditions of the $k$-transporter problem, and this technically suffices for ZAP's needs. In unit propagation, however, it is useful to collect all of the unit consequences of an augmented clause $(c, G)$ at once, as opposed to collecting them via repeated traversals of $G$'s coset decomposition tree.

As we work through the consequences of this observation, it will help to have an example that illustrates the points we are going to be making. To this end, we will consider the augmented clause

$$(a \vee b \vee e, \operatorname{Sym}(a, b, c, d) \times \operatorname{Sym}(e, f)) \tag{20}$$

in a situation where $a$, $b$ and $c$ are false and $d$, $e$ and $f$ are unvalued. The group in (20) allows arbitrary permutations of $\{a, b, c, d\}$ and of $\{e, f\}$, so that both $e$ and $f$ are unit consequences of instances of the given augmented clause.

Note that we cannot simply collect all the group elements associated with each unit instance, since many group elements may correspond to the same clause instance $c^g$ or to the same unit literal $c^g \cap U$. In the above example, both ( ) and ($ab$) correspond to the identical clause $a \vee b \vee e$, and both this clause and $a \vee c \vee e$ lead to the same conclusion $e$ given the current partial assignment.

Our goal will therefore be to compute not a set of permutations, but the associated set of all unit conclusions:

**Definition 6.1** *Let $(c, G)$ be an augmented clause, and $P$ a partial assignment. The* unit consequences *of $(c, G)$ given $P$ is the set of all literals $l$ such that there is a $g \in G$ with $c^g \cap S(P) = \emptyset$ and $c^g \cap U(P) = \{l\}$. For a fixed literal $w$, the* unit $w$-consequences *of $(c, G)$ given $P$ is the set of all literals $l$ such that there is a $g \in G$ with $w \in c^g$, $c^g \cap S(P) = \emptyset$ and $c^g \cap U(P) = \{l\}$.*

The unit $w$-consequences involve an additional requirement that the literal $w$ appear in the clause instance in question. This will be useful when we discuss watched literals in the next section.

In our example, the unit consequences of (20) are $e$ and $f$. The unit $c$-consequences are the same, although we can no longer use the identity permutation ( ), since the needed $c$ is not in the base instance of (20). There are no unit $d$-consequences of (20).

If the partial assignment is to be annotated, we will need not just the unit consequences, but the reasons as well:

**Definition 6.2** *Let $X$ be a set of pairs $\langle l, g \rangle$, where $g \in G$ and $l$ is a literal for each pair. If $X = \{\langle l_1, g_1 \rangle, \ldots, \langle l_n, g_n \rangle\}$, we will denote $\{l_1, \ldots, l_n\}$ by $L(X)$.*

*If $(c, G)$ is an augmented clause and $P$ a partial assignment, $X$ will be called an* annotated set of unit consequences *of $(c, G)$ given $P$ if:*

1. *$c^g \cap S(P) = \varnothing$ and $c^g \cap U(P) = \{l\}$ for every $\langle l, g \rangle \in X$ and*

2. *$L(X)$ is the set of unit consequences of $(c, G)$ given $P$.*

Once again returning to our example, $\langle e, (\,) \rangle$ is an annotated consequence, as is $\langle e, (abc) \rangle$. So are $\langle f, (ef) \rangle$ and $\langle f, (abc)(ef) \rangle$. The set $\{\langle e, (abc) \rangle, \langle f, (ef) \rangle\}$ is an annotated set of unit consequences, as is $\{\langle e, (abc) \rangle, \langle f, (ef) \rangle, \langle f, (abc)(ef) \rangle\}$. But $\{\langle f, (ef) \rangle, \langle f, (abc)(ef) \rangle\}$ is not an annotated set of unit consequences, since $e$ does not appear as a consequence.

We now modify our $k$-transporter procedure so that we search the entire tree while accumulating an annotated set of unit consequences. We need to be careful, however, because the pruning lemmas may prune a node because it includes a permutation $g$ that is not minimal in its double coset $JgK$. This is a problem because $g$ and the minimal element of $JgK$ may correspond to distinct unit consequences. In our running example, it may well be that none of the minimal elements of $JgK$ supports $f$ as a conclusion; if we accumulate only all of the minimal elements, we will not get a full set of unit consequences as a result.

Given a successful $g$ that *is* minimal in its double coset, reconstructing the relevant orbits under $J$ and $K$ is easy, so we begin by introducing some definitions that cater to this. The basic idea is that we want the minimal $g$ to "entail", in some sense, the conclusions that can be drawn from other permutations in the double coset $JgK$.

In our example, the subgroup of $G$ that simultaneously stabilizes $S$ and $U$ is $G_{\{S,U\}} = \mathrm{Sym}(a, b, c) \times \mathrm{Sym}(e, f)$. Once we have a permutation $g_1$ that allows us to conclude $e$, we can operate with $g_1 \cdot (ef) \in g_1 G_{\{S,U\}}$ to conclude $f$ as well. We formalize this as follows:

**Definition 6.3** *Given a group $G$, we will say that $\langle l_1, g_1 \rangle$ $G$-entails $\langle l_2, g_2 \rangle$, to be denoted $\langle l_1, g_1 \rangle \models_G \langle l_2, g_2 \rangle$, if there is some $g \in G$ such that $l_2 = l_1^g$ and $g_2 = g_1 g$. We will say that a set of pairs $X$ $G$-entails a set of pairs $Y$, writing $X \models_G Y$, if every pair in $Y$ is $G$-entailed by some pair in $X$.*

A skeletal set of unit consequences of $(c, G)$ given $P$ is any set $X$ of unit consequences that $G_{\{S(P), U(P)\}}$-entails an annotated set of unit consequences of $(c, G)$ given $P$.

In our running example, we have $l_1 = e$ and $g = (ef)$ in the first paragraph, allowing (for example) $\langle e, (\,) \rangle$ to $G_{\{S,U\}}$-entail $\langle f, (ef) \rangle$. Thus we see that $\{\langle e, (\,) \rangle\}$ is a skeletal set of unit consequences of (20) given the partial assignment $\{\neg a, \neg b, \neg c\}$.

**Lemma 6.4** *If $X \models_G Y$, then $L(Y) \subseteq L(X)^G$.*

**Proof.** Every pair in $Y$ is of the form $\langle l_1^g, g_1 g \rangle$ for $\langle l_1, g_1 \rangle \in X$ and $g \in G$. Thus the associated literal is in $L(X)^G$. □

To construct a full set of unit consequences from a skeletal set, we repeatedly find new unit conclusions until no more are possible:

**Procedure 6.5** *Given a set $X$ of pairs $\langle l, g \rangle$ and a group $G$, to compute $\texttt{complete}(X, G)$, where $X \models_G \texttt{complete}(X, G)$ and $L(\texttt{complete}(X, G)) = L(X)^G$:*

```
1  Y ← ∅
2  for each ⟨l, g⟩ ∈ X
3       do for each l' ∈ l^G − L(Y)
4            do select h ∈ G such that l^h = l'
5                 Y ← Y ∪ ⟨l', gh⟩
6  return Y
```

**Proposition 6.6** *$X \models_G \texttt{complete}(X, G)$ and $L(\texttt{complete}(X, G)) = L(X)^G$.*

Now we can apply the pruning lemmas as the search proceeds, eventually returning a skeletal set of unit consequences for the clause in question. In addition, if there is a unit instance that is in fact unsatisfiable, we should return a failure marker of some sort. We handle this by returning two values. The first indicates whether or not a contradiction was found, and the second is the skeletal set of unit consequences.

**Procedure 6.7** *Given groups $H \leq G$, an element $t \in G$, sets $c$, $S$ and $U$, to find $\texttt{Transport}(G, H, t, c, S, U)$, a skeletal set of unit consequences for $(c, G)$ given $P$:*

```
1   if overlap(H, c, S^{t^{-1}}) > 0
2      then return ⟨false, ∅⟩
3   if overlap(H, c, (S ∪ U)^{t^{-1}}) > 1
4      then return ⟨false, ∅⟩
5   if c = c_H
6      then if c^t ∩ U = ∅
7              then return ⟨true, 1⟩
8              else  return ⟨false, ⟨c^t ∩ U, 1⟩⟩
9   if a pruning lemma can be applied
10     then return ⟨false, ∅⟩
11  Y ← ∅
12  α ← an element of c − c_H
13  for each t' in (H : H_α)
14       do ⟨u, V⟩ ← Transport(G, H_α, t't, c, S, U)
15          if u = true
16          then return ⟨true, Vt'⟩
17          else Y ← Y ∪ {⟨l, gt'⟩|⟨l, g⟩ ∈ V}
18  return ⟨false, Y⟩
```

**Proposition 6.8** *Assume that $|c^h \cap V| \geq \mathtt{overlap}(H, c, V) \geq |c_H \cap V|$ for all $h \in H$, and let $\mathtt{Transport}(G, c, S, U)$ be computed by Procedure 6.7. Then if there is a $g \in G$ such that $c^g \cap S = c^g \cap U = \emptyset$, $\mathtt{Transport}(G, c, S, U) = \langle \mathtt{true}, g \rangle$ for such a $g$. If there is no such $g$, $\mathtt{Transport}(G, c, S, U) = \langle \mathtt{false}, Z \rangle$, where $Z$ is a skeletal set of unit consequences for $(c, G)$ given $P$.*

As an application of the pruning lemmas, we have:

**Proposition 6.9** *Let $(c, G)$ be an augmented clause corresponding to a cardinality constraint. Then for any sets $S$ and $U$, Procedure 6.7 will expand at most a linear number of nodes in finding a skeletal set of unit consequences of $(c, G)$.*

In the original formulation of cardinality constraints (as in ZAP1), determining if a particular constraint is unit (and finding the implied literals if so) takes time linear in the length of the constraint, since it involves a simple walk along the constraint itself. It therefore seems appropriate for a linear number of nodes to be expanded in this case.

## 7. Watched Literals

There is one pruning technique that we have not yet considered, and that is the possibility of finding an analog in our setting to Zhang and Stickel's (2000) watched literal idea.

To understand the basic idea, suppose that we are checking to see if the clause $a \vee b \vee c$ is unit in a situation where $a$ and $b$ are unvalued. It follows that the clause cannot be unit, independent of the value assigned to $c$.

At this point, we can *watch* the literals $a$ and $b$; as long as they remain unvalued, the clause cannot be unit. In practice, the data structures representing $a$ and $b$ include a pointer to the clause in question, and the unit test needs only be performed for clauses pointed to by literals that are changing value.

As we continue to discuss these ideas, it will be useful to distinguish among three different types of clauses: those that are satisfied given the current partial assignment, those that are unit, and those that are neither:

**Definition 7.1** *Let $C$ be a clause, and $P$ a (possibly annotated) partial assignment. We will say that $C$ is settled by $P$ if it is either satisfied or unit; otherwise it is unsettled.*

We now have:

**Definition 7.2** *Let $C$ be a clause, and $P$ a (possibly annotated) partial assignment. If $C$ is unsettled by $P$, then a watching set for $C$ under $P$ is any set of literals $W$ such that $|W \cap C \cap U(P)| > 1$.*

In other words, $W$ contains at least two unvalued literals in $C$ if $C$ is unsettled by the current partial assignment.

What about if $C$ is satisfied or unit? What should the watching set be in this case?

In some sense, it doesn't matter. Assuming that we notice when a clause changes from unsettled to unit (so that we can either unit propagate or detect a potential contradiction), settled clauses are uninteresting from this perspective, since they can never generate a second unit propagation. So we can watch a settled clause or not, as we see fit.

In another sense, however, it does matter. One of the properties that we would like the watching sets to have is that they remain valid during a backtrack. That means that if a settled clause $C$ becomes unsettled during a backtrack, there must be two watched and unvalued variables after that backtrack.

In order to discuss backtracking in a formal way, we introduce:

**Definition 7.3** *Let $P$ be a partial assignment for a set $T$ of (possibly augmented) clauses. We will say that $P$ is $T$-closed if no clause $C \in T$ has a unit consequence given $P$. A $T$-closure of $P$ is any minimal, sound and $T$-closed extension of $P$, and will be denoted by either $\overline{P_T}$ or by simply $\overline{P}$ if $T$ is clear from context.*

The definition of closure makes sense because the intersection of two closed partial assignments is closed as well. To compute the closure, we simply add unit consequences one at a time until no more are available. Note that there is still some ambiguity; if there is more than one unit consequence that can be added at some point, we can add the unit consequences in any order.

**Definition 7.4** *Let $P = \langle l_1, \ldots, l_n \rangle$ be a partial assignment. A* subassignment *of $P$ is any initial subsequence $\langle l_1, \ldots, l_j \rangle$ for $j \leq n$. We will say that a subassignment $P'$ of $P$ is a* backtrack point *for $P$ if either $P' = P$ or $P' = \overline{P'}$. We will denote by $P_-$ the largest backtrack point for $P$ that is not $P$ itself.*

*If $C$ is a clause, we will say that the $P$-retraction of $C$, to be denoted $P_{\neg C}$, is the largest backtrack point for $P$ for which $C$ is unsettled.*

Note that we require a backtrack to the point that $C$ is *unsettled*, as opposed to simply unsatisfied. If $P$ is closed, there is no difference because Definition 7.4 does not permit a backtrack to a point where $C$ is unit. But if $C$ is unit under $P$, we can only "retract" $C$ by reverting to a point before $C$ became unit. Otherwise, $C$ will simply be reasserted when unit propagation computes $\overline{P}$.

Since $P$ itself is a backtrack point for $P$, we immediately have:

**Lemma 7.5** *If $C$ is unsettled by $P$, then $P_{\neg C} = P$.*  ◻

As an example, suppose that we have the following annotated partial assignment $P$:

| literal | reason |
|:---:|:---:|
| $a$ | `true` |
| $\neg b$ | `true` |
| $c$ | $\neg a \vee b \vee c$ |
| $d$ | `true` |
| $e$ | $b \vee \neg d \vee e$ |

If our clause $C$ is $b \vee e \vee f$, the $P$-retraction of $C$ is $\langle a, \neg b, c \rangle$. Removing $e$ is sufficient to make $C$ unsettled, but $\langle a, \neg b, c, d \rangle$ is not closed and is therefore not a legal backtrack point. If $b \vee e$ is in our theory, the retraction is in fact $\langle a \rangle$ because $\langle a, \neg b, c \rangle$ is not a backtrack point because the unit conclusion $e$ has not been drawn.

We can now generalize Definition 7.2 to include settled clauses:

**Definition 7.6** *Let $C$ be a clause, and $P$ an annotated partial assignment. A* watching set *for $C$ under $P$ is any set of literals $W$ such that $|W \cap C \cap U(P_{\neg C})| > 1$.*

In other words, $W$ will contain at least two unvalued literals in $C$ if we replace $P$ with the $P$-retraction of $C$. As discussed earlier, this is the first point to which we could backtrack so that $C$ was no longer satisfied or unit. Continuing our earlier example, $\{e, f\}$ is a watching set for $b \vee e \vee f$, and $\{\neg b, e\}$ is a watching set for $\neg b \vee e$. A watching set for $b \vee e$ is $\{b, e\}$; recall that the definition forces us to backtrack all the way to $\langle a \rangle$.

**Lemma 7.7** *If $W$ is a watching set for $C$ under $P$, then so is any superset of $W$.* □

In order for watching sets to be useful, of course, we must maintain them as the search proceeds. Ideally, this maintenance would involve modifying the watching sets as infrequently as possible, so that we could adjust them only as required when variables take new values, and not during backtracking at all. Recall the example at the beginning of this section, where $a$ and $b$ are unvalued and constitute a watching set for the clause $a \vee b \vee c$. If $a$ or $b$ becomes satisfied, we need do nothing since the clause is now satisfied and $\{a, b\}$ is still a watching set. Note that if $a$ (for example) becomes satisfied, we can't remove $b$ from the watching set, since we would then need to replace it if we backtrack to the point that $a$ is unvalued once again. Leaving $b$ in the watching set is required to satisfy Definition 7.6 and needed to ensure that the sets need not be adjusted after a backtrack.

On the other hand, if $a$ (for example) becomes *un*satisfied, we need to check the clause to see whether or not it has become unit. If the clause is unit, then $b$ should be set to true by unit propagation, so no maintenance is required. If the clause is unsettled, then $c$ must be unvalued, so we can replace $a$ with $c$ in the set of literals watching the clause. Finally, if the clause is already satisfied, then $a$ will be unvalued in the $P$-retraction of the clause and the watching set need not be modified.

In general, we have:

**Proposition 7.8** *Suppose that $W$ is a watching set for $C$ under $P$ and $l$ is a literal. Then:*

1. *$W$ is a watching set for $C$ under any backtrack point for $P$.*

2. *If $C$ is settled by $\langle P, l \rangle$, then $W$ is a watching set for $C$ under $\langle P, l \rangle$.*

3. *If $C$ is settled by $\langle P, l \rangle$, and $|(W - \{\neg l\}) \cap C \cap U(P_{\neg C})| > 1$, then $W - \{\neg l\}$ is a watching set for $C$ under $\langle P, l \rangle$.*

4. *If $\neg l \notin W \cap C$, then $W$ is a watching set for $C$ under $\langle P, l \rangle$.*

The proposition tells us how to modify the watching sets as the search proceeds. No modification is required during a backtrack (claim 1). No modification is required if the clause is satisfied or unit (claim 2), and we can also remove a newly valued literal from a watching set if enough other unvalued variables are present (claim 3). No modification is required unless we add the negation of an already watched literal (claim 4).

In sum, modification to the watching sets is only required when we add the negation of a watched literal to our partial assignment and the watched clause is not settled; in this case, we have to add one of the remaining unvalued literals to the watching set. In addition, we can remove literals from the watching set if enough unvalued literals are already in it. Since this last possibility is not used in ZCHAFF or other ground systems, here is an example of it.

Suppose that we are, as usual, watching $a$ and $b$ in $a \vee b \vee c$. At some point, $a$ becomes true. We can either leave the watching set alone by virtue of condition 4, or we can extend the watching set to include $c$ (extending a watching set is always admissible, by virtue of Lemma 7.7), and then remove $a$ from the watching set. This change is unneeded in a ground prover, but will be useful in the augmented version 7.10 of the proposition below.

To lift these ideas to an augmented setting, we begin by modifying Definition 7.6 in the obvious way to get:

**Definition 7.9** *Let $(c, G)$ be an augmented clause, and $P$ an annotated partial assignment. A watching set for $(c, G)$ under $P$ is any set of literals $W$ that is a watching set for every instance $c^g$ of $(c, G)$ under $P$.*

This leads to the following augmented analog of Proposition 7.8. (Although there are four clauses in Proposition 7.8 and four in the following proposition, there is no clause-for-clause correspondence between the two results.)

**Proposition 7.10** *Suppose that $W$ is a watching set for $(c, G)$ under $P$ and $l$ is a literal. Then:*

1. *$W$ is a watching set for $(c, G)$ under any backtrack point for $P$.*

2. *If $\neg l \notin W \cap c^G$, then $W$ is a watching set for $(c, G)$ under $\langle P, l \rangle$.*

3. *If $|(W \cup V) \cap c^g \cap U(\langle P, l \rangle)| > 1$ for every $g \in G$ such that $c^g$ is unsettled by $\langle P, l \rangle$, then $W \cup V$ is a watching set for $(c, G)$ under $\langle P, l \rangle$.*

4. *If $|(W \cup V) \cap c^g \cap [U(\langle P, l \rangle) \cup (S(P) - S(P_-))]| > 1$ for every $g \in G$, then $W \cup V - \{\neg l\}$ is a watching set for $(c, G)$ under $\langle P, l \rangle$.*

As an example, suppose that we return to the augmented clause we considered in the previous section, $(a \vee b \vee e, \operatorname{Sym}(a, b, c, d) \times \operatorname{Sym}(e, f))$. Suppose that we are initially watching $a$, $b$, $c$ and $d$, and that $e$ is false, and now imagine that $a$ becomes false as well.

We need to augment $W$ so that $|W \cap c^g \cap U(P)| > 1$ for every unsettled instance $c^g$ of $(c, G)$ that contains $a$. Those instances are $a \vee b \vee f$, $a \vee c \vee f$ and $a \vee d \vee f$. Since $b$, $c$ and $d$ are already in $W$, we need to add $f$. If $f$ had been in the watching set but not $b$, $c$ and $d$, we would have had to add those three points instead.

In this case, since the clause has a unit instance ($a \vee b \vee e$, for example), we cannot remove $a$ from the watching set. The reason is that if we do so and later backtrack past this point, we are in danger of watching only $b$ for this unsatisfied clause.

Suppose, however, that $e$ had been unvalued when $a$ became false. Now we would have to add both $e$ and $f$ to the watching set and we would be free to remove $a$. This is sanctioned by Proposition 7.10, since $(c, G)$ has no settled instances and if $c^g \cap S(P) = \emptyset$ for all $g \in G$ as well, the conditions of claims three and four are equivalent.

What if $e$, instead of being false or unvalued, had been true? Now we add $f$ to the watching set, but can we remove $a$ from the new watching set $\{a, b, c, d, f\}$? We cannot: the instance $a \vee b \vee e$ would have only one watched literal if we did.

In some cases, however, we can remove the literal that just became false from the watching set. We can surely do so if every clause instance still has two unvalued literals in the watching set. This would correspond to the requirement that

$$|(W \cup V) \cap c^g \cap U(\langle P, l \rangle)| > 1$$

for every instance. The stronger condition in claim four of the proposition allows us to do slightly better in cases where the satisfied literal in the clause became satisfied sufficiently recently that we know that any backtrack will unvalue it.

The fourth conclusion in Proposition 7.10 is essential to the effective functioning of our overall prover; when we replace a watched literal $l$ that has become false with a new and unvalued literal, it is important that we *stop* watching the original watched literal $l$. It is the last claim in the proposition that allows us to do this in most (although not all) practical cases. Without this fourth conclusion, the watching sets would only get larger as the search proceeded. Eventually, every literal in every clause would be watched and the computational power of the idea would be lost.

We can now use the watching sets to reduce the number of clauses that must be examined in line 1 of the unit propagation procedure 2.7. Each augmented clause needs to be associated with a watching set that is initialized and updated as sanctioned by Proposition 7.10.

Initialization is straightforward; for any clause $(c, G)$ with $c$ of length at least two, we need to define an associated watching set $W$ with the property that $|W \cap c^g| > 1$ for every $g \in G$. In fact, we take $W$ to be simply $c^G$, the union of all of the instances $c^g$, and rely on subsequent unit tests to gradually reduce the size of $W$. (Once again, using the fourth clause of Proposition 7.10.) The challenge is to modify Procedure 6.7 in a way that facilitates the maintenance of the watching sets.

Before doing this, let us understand in a bit more detail how the watching sets are used in searching for unit instances of a particular augmented clause. Consider the augmented clause corresponding to the quantified clause

$$\forall xy \, . \, [q(x) \wedge r(y) \rightarrow s]$$

If $Q$ is the set of instances of $q(x)$ and $R$ the set of instances of $r(y)$, this becomes the augmented clause

$$(\neg q(0) \vee \neg r(0) \vee s, \mathrm{Sym}(Q) \times \mathrm{Sym}(R)) \tag{21}$$

where $q(0)$ and $r(0)$ are elements of $Q$ and $R$ respectively.

Now suppose that $r(y)$ is true for all $y$, but $q(x)$ is unvalued, as is $s$, so that the clause (21) has no unit instances. Suppose also that we search for unit instances of (21) by first stabilizing the image of $r$ and then of $q$ ($s$ is stabilized by the group $\mathrm{Sym}(Q) \times \mathrm{Sym}(R)$ itself). If there are four possible bindings for $y$ (which we will denote $0, 1, 2, 3$) and three for $x$ ($0, 1, 2$), the search space looks like this:

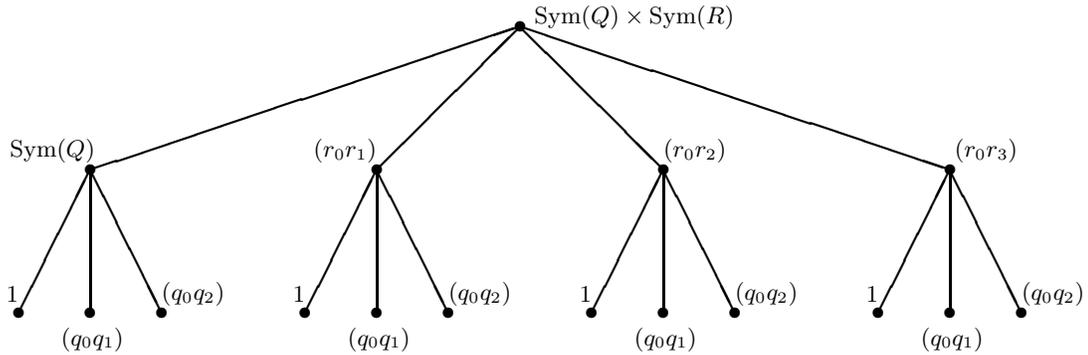

In the interests of conserving space, we have written $q_i$ instead of $q(i)$ and similarly for $r_j$.

Each of the leaf nodes fails because both $s$ and the relevant instance of $q(x)$ are unvalued, and we now construct a new watching set for the entire clause (21) that watches $s$ and all of the $q(x)$.

Note that this causes us to lose significant amounts of information regarding portions of the search space that need not be reexamined. In this example, the responsible literals at each leaf node are as follows:

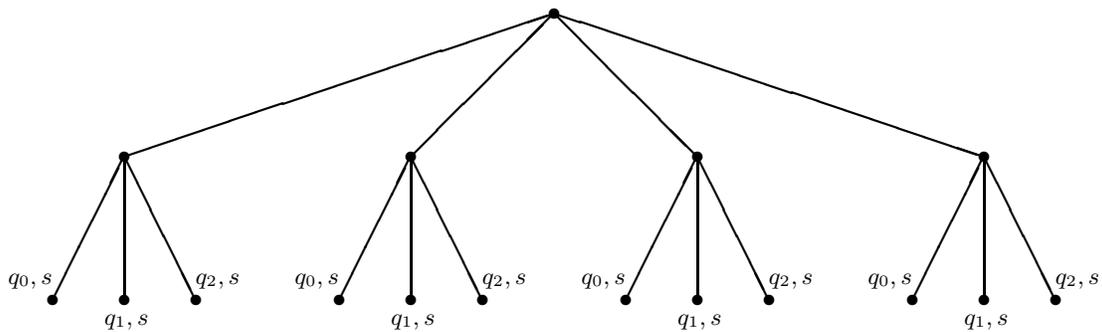

When we simply accumulate these literals at the root of the search tree, we conclude that the reason for the failure is the watching set $\{q_0, q_1, q_2, s\}$. If any of these watched literals changes value, we potentially have to reexamine the entire search tree.

We can address this by changing the order of variable stabilization, replacing the search space depicted above with the following one:

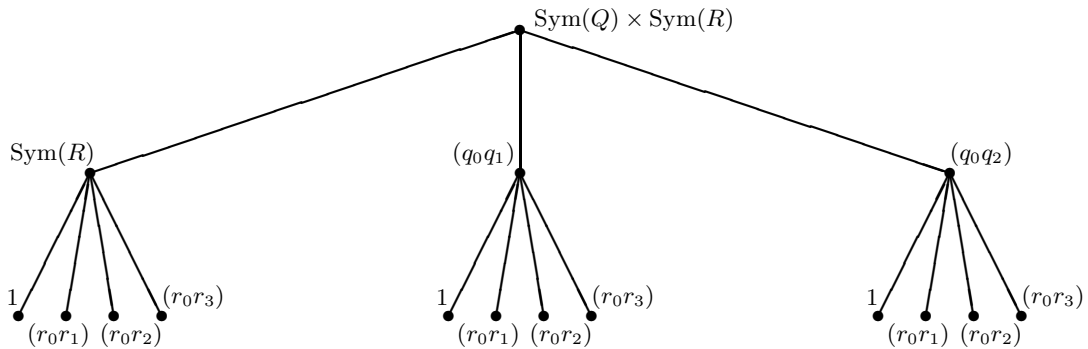

Now only the center node needs reexpansion if the value of $q_1$ changes, since it is only at this node that $q_1$ appears. The search space becomes simply:

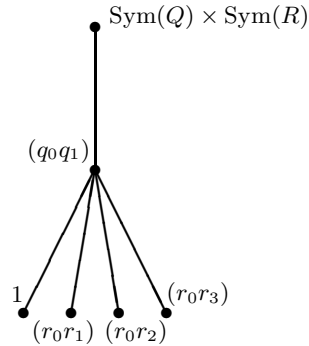

which is what one would expect if $q_1$ changes value.

The upshot of this is that while we collect a new watching set for the original augmented clause corresponding to (21), we also need to modify our unit propagation procedure so that we first stabilize points that can be mapped to a specific watched literal that has become unsatisfied.

To see how to keep the watching set updated, consider Proposition 7.10. When searching for the unit instances of an augmented clause $(c, G)$, we need to compute some set $W$ such that $|W \cap c^g \cap U(P)| > 1$ for every unsettled instance $c^g$ of $(c, G)$ that contains a fixed literal $w$. How are we to do this?

The solution lies in Procedure 6.7, which describes our search for unit instances. If all of the remaining clause instances below some particular search node are determined to be nonunit in the test on line 3, instead of simply recognizing that every instance under this node is nonunit, we need to be able to identify a set of unvalued literals that meets every unsettled instance of $c^g$ at least twice. We modify the `overlap` procedure 5.19 to become:

**Procedure 7.11** *Given a group $H$, two sets $c, V$ acted on by $H$, and a bound $k \geq 0$, to compute* `overlap`$(H, c, V, k)$, *a collection of elements of $V$ sufficient to guarantee that for any $h \in H$, $|c^h \cap V| > k$, or $\emptyset$ if no such collection exists:*

```
1   m ← 0
2   W ← ∅
3   for each orbit X of H
4       do {B₁, . . . , Bₖ} ← a minimal block system for W under H for which
                 c ∩ W ⊆ Bᵢ for some i
5          Δ = |c ∩ X| + min(Bᵢ ∩ V) − |B₁|
6          if Δ > 0
7             then m ← m + Δ
8                  W ← W ∪ (X ∩ V)
9                  if m > k
10                    then return W
11  return ∅
```

**Proposition 7.12** *Procedure 7.11 returns a nonempty set $W$ if and only if Procedure 5.19 returns a value in excess of $k$. In this case, $|c^h \cap W| > k$ for every $h \in H$.*

We are finally in a position to replace Procedure 6.7 with a version that uses watched literals:

**Procedure 7.13** *Given groups $H \leq G$, an element $t \in G$, sets $c$, $S$ and $U$, and optionally a watched element $w$, to find $\mathtt{Transport}(G, H, t, c, S, U, w)$, a skeletal set of unit $w$-consequences for $(c, G)$ given $P$:*

```
 1  if w is supplied and w^{t^{-1}} ∉ c^H
 2      then return ⟨false, ∅, ∅⟩
 3  V ← overlap(H, c, S^{t^{-1}}, 0)
 4  if V ≠ ∅
 5      then return ⟨false, ∅, ∅⟩
 6  V ← overlap(H, c, (S ∪ U)^{t^{-1}}, 1)
 7  if V ≠ ∅
 8      then return ⟨false, ∅, V^t⟩
 9  if c = c_H
10      then if c^t ∩ U = ∅
11              then return ⟨true, 1, ∅⟩
12              else return ⟨false, ⟨c^t ∩ U, 1⟩, ∅⟩
13  if a pruning lemma can be applied
14      then return ⟨false, ∅, ∅⟩
15  α ← an element of c − c_H. If w is supplied and w ∉ c_H^t, choose α so that w^{t^{-1}} ∈ α^H.
16  Y ← ∅
17  W ← ∅
18  for each t' in (H : H_α)
19      do ⟨u, V, X⟩ ← Transport(G, H_α, t't, c, S, U, w)
20          if u = true
21              then return ⟨true, Vt', ∅⟩
22              else W ← W ∪ X
23                  Y ← Y ∪ {⟨l, gt'⟩ | ⟨l, g⟩ ∈ V}
24  return ⟨false, Y, W⟩
```

In the application of the pruning lemmas in line 13, we need to use the restricted group $G_{\{S, U, \{w\}\}}$, so that we do not prune a group element $g$ with $w \in c^g$ on the basis of another group element $jgk$ for which $w \notin c^{jgk}$, since $jgk$ might itself then be pruned on line 2.

**Proposition 7.14** *Suppose that $\mathtt{overlap}(H, c, V, k)$ is computed using Procedure 7.11, or otherwise satisfies the conclusion of Proposition 7.12. Then if there is a $g \in G$ such that $w \in c^g$ and $c^g \cap S = c^g \cap U = \emptyset$, $\mathtt{Transport}(G, c, S, U, w)$ as computed by Procedure 7.13 returns $\langle \mathtt{true}, g, \emptyset \rangle$ for such a $g$. If there is no such $g$, Procedure 7.13 returns $\langle \mathtt{false}, Z, W \rangle$, where $Z$ is a skeletal set of unit $w$-consequences of $(c, G)$ given $P$, and $W$ is such that $|W^{G_{\{S, U, \{w\}\}}} \cap c^h \cap U| > 1$ for every $h \in H$ such that $w \in c^h$ and $c^h$ is unsettled by $P$.*

Note that the pruning lemmas are applied relatively late in the procedure (line 13) even though a successful application prunes the space without increasing the size of the watching set. It might seem that the pruning lemmas should be applied earlier.

This appears not to be the case. As discussed at the end of Section 5, the pruning lemmas are relatively complex to check; moving the test earlier (to precede line 6, presumably) actually slows the unit propagation procedure by a factor of approximately two, primarily due to the need to compute the set stabilizer $G_{\{S,U\}}$ even in cases where a simple counting argument suffices. In addition, the absolute impact on the watching sets can be expected to be quite small.

To understand why, suppose that we are executing the procedure for an instance where it will eventually fail. Now if $n$ is a node that can be pruned either by a counting argument (with the new contribution $W_n$ to the set of watched literals) or by a lexicographic argument using another node $n'$, then since the node $n'$ will eventually fail, it will contribute its own watching set $W_{n'}$ to the eventually returned value. While it is possible that $W_n \neq W_{n'}$ (different elements can be selected by the `overlap` function in line 6, for example), we expect that in the vast majority of cases we will have $W_n = W_{n'}$ and the non-lexicographic prune will have no impact on the eventual watching set computed.

Proposition 7.14 implies that the watching set returned by Procedure 7.13 can be used to update the watching set as in the third claim of Proposition 7.10. For the fourth claim, where we hope to remove $\neg l$ from the new watching set, we need to check to see if

$$|W \cap c^g \cap [U(\langle P, l\rangle) \cup (S(P) - S(P_-))]| > 1$$

for each $g \in G$, where $W$ is the new watching set. This can be determined by a single call to `transport`; if there is no $g \in G$ for which

$$|c^g \cap [W \cap (U(\langle P, l\rangle) \cup (S(P) - S(P_-)))]| \leq 1 \tag{22}$$

we can remove $\neg l$ from $W$. In some cases, we can save the call to `transport` by exploiting the fact (as shown in the proof of Proposition 7.10) that (22) cannot be satisfied if $\langle P, l\rangle$ has a unit consequence.

We are finally in a position to describe watched literals in an augmented setting. As a start, we have:

**Definition 7.15** *A* watched augmented clause *is a pair $\langle (c, G), W \rangle$ where $(c, G)$ is an augmented clause and $W$ is a watching set for $(c, G)$.*

**Procedure 7.16 (Unit propagation)** *To compute* Unit-Propagate$(C, P, L)$ *where $C$ is a set of watched augmented clauses, $P$ is an annotated partial assignment, and $L$ is a set of pairs $\langle l, r\rangle$ of literals $l$ and reasons $r$:*

```
1   while L ≠ ∅
2       do ⟨l, r⟩ ← an element of L
3           L ← L − ⟨l, r⟩
4           P ← ⟨P, ⟨l, r⟩⟩
5           for each ⟨(c, G), W⟩ ∈ C
6               do if ¬l ∈ W
7                   then ⟨r, H, V⟩ ← Transport(G, c, S(P), U(P), ¬l)
8                       if r = true
9                           then lᵢ ← the literal in c^H with the highest index in P
10                              return ⟨true, resolve((c^H, G), cᵢ)⟩
11                      H' ← complete(H, G_{{S(P),U(P),{l}}})
12                      for each h ∈ H'
13                          do z ← the literal in c^h unassigned by P
14                              if there is no ⟨z, r'⟩ in L
15                                  then L ← L ∪ ⟨z, c^h⟩
16                      W ← W ∪ (U(P) ∩ V^{G_{{S(P),U(P),{l}}}})
17                      U ← U(P) ∪ (S(P) − S(P_−))
18                      if H = ∅ ∧ transport(G, c, ∅, W ∩ U, 1, ¬l) = FAILURE
19                          then W ← W − {¬l}
20  return ⟨false, P⟩
```

On line 18, we invoke a version of the `transport` function that accepts as an additional argument a literal that is required to be included in the clause instance being sought. This modification is similar to the introduction of such a literal $w$ in the `Transport` procedure 7.13.

**Proposition 7.17** *Let $P$ be an annotated partial assignment, and $C$ a set of watched augmented clauses, where for every $\langle (c, G), W \rangle \in C$, $W$ is a watching set for $(c, G)$ under $P$. Let $L$ be the set of unit consequences of clauses in $C$. If* Unit-Propagate$(C, P, L)$ *returns* $\langle \text{true}, c \rangle$ *for an augmented clause $c$, then $c$ is a nogood for $P$, and any modified watching sets in $C$ are still watching sets under $P$. Otherwise, the value returned is* $\langle \text{false}, \overline{P} \rangle$ *and the watching sets in $C$ will all have been replaced with watching sets under $\overline{P}$.*

Procedure 7.16 can be modified and incorporated in a fairly obvious way into Procedure 2.8, where the literal most recently added to the partial assignment is added to $L$ and thereby passed into the unit propagation procedure.

# 8. Resolution Revisited

There is one additional theoretical point that we need to discuss before turning our attention to experimental matters.

The goal in augmented resolution is to produce many (if not all) of the resolvents sanctioned by instances of the augmented clauses being resolved. As we showed in ZAP2, however, it is not always possible to produce all such resolvents. Here is another example of that phenomenon.

Suppose that we are resolving the two clauses

$$(a \vee c, (ab)) \tag{23}$$

and

$$(b \vee \neg c, (ab)) \tag{24}$$

The result is[10]

$$(a \vee b, (ab)) \tag{25}$$

But consider the example. The instances of (23) are $a \vee c$ and $b \vee c$; those of (24) are $b \vee \neg c$ and $a \vee \neg c$. Surely it is better to have the resolvent be $(a, (ab))$ instead of (25). In general, we never want to conclude $(c, G)$ when it is possible to conclude $(c', G)$ for $c' \subset c$ where the set inclusion is proper. The resolvent with $c'$ is properly stronger than that with $c$.

There is an additional consideration as well. Suppose that we are resolving two augmented clauses, and can choose instances of the resolving clauses so that the resolvent is $(a \vee c, G)$ or $(b \vee c, G)$, where $a$ and $b$ are literals and the two possible resolvents are distinct because $(ab) \notin G$. Which should we select?

We know of no general answer, but a reasonable heuristic is to make the choice based on the order in which literals were added to the current partial assignment. Assuming that the resolvent is a nogood, presumably $a$ and $b$ are both false for the current partial assignment $P$. We should select the resolvent that allows a larger backjump; in this case, the resolvent involving the literal that was added to $P$ first.

All of these considerations have no direct analog in a conventional Boolean satisfiability engine. For any particular literal $l$, the resolvent of the reasons for $l$ and for $\neg l$ is just that; there is no flexibility possible.[11]

**Definition 8.1** *Let $(\alpha, G)$ and $(\beta, H)$ be two augmented clauses resolving on a literal $l$, so that $l \in \alpha$ and $\neg l \in \beta$. An $l$-resolvent for $(\alpha, G)$ and $(\beta, H)$ will be any clause obtained by resolving $\alpha^g$ and $\beta^h$ for $g \in G$ and $h \in H$ such that $l \in \alpha^g$ and $\neg l \in \beta^h$.*

Note that the group $Z$ in the resolvent clause $(\mathtt{resolve}(\alpha^g, \beta^h), Z)$ is independent of the resolvent selected, so we can focus our attention strictly on the syntactic properties of the resolvent.

We next formalize the fact that the partial assignment $P$ induces a natural lexicographic ordering on the set of nogoods for a given theory:

**Definition 8.2** *Let $P$ be a partial assignment, and $c$ a ground clause. If $l$ is the literal in $c$ whose negation has maximal index in $P$, we will say that the* falsification depth *of $c$ is the position in $P$ of the literal $\neg l$. The falsification depth is zero if there is no such literal in $c$; in any event, the falsification depth of $c$ will be denoted by $c^{?P}$.*

*If $c_1$ and $c_2$ are two nogoods, we will say that $c_1$ is falsified earlier than $c_2$ by $P$, writing $c_1 <_P c_2$, if either $c_1^{?P} < c_2^{?P}$, or $c_1^{?P} = c_2^{?P}$ and $c_1 - \neg l_{c_1^{?P}} <_P c_2 - \neg l_{c_2^{?P}}$.*

---

10. The result can be obtained by direct computation or by applying the resolution stability property discussed in ZAP2, since the groups are identical.

11. A weak analog is present in ZCHAFF, which can replace one nogood $n$ with another $n'$ if $n'$ leads to a greater backjump than $n$ does. This functionality is part of the ZCHAFF code but does not appear to have been documented.

As an example, suppose that $P$ is $\langle a, b, c, d, e \rangle$. The falsification depth of $\neg a \vee \neg c$ is three, since $c$ is the third variable assigned in $P$. The falsification depth of $\neg b \vee \neg d$ is four. Thus $\neg a \vee \neg c <_P \neg b \vee \neg d$; we would rather learn $\neg a \vee \neg c$ because it allows us to backjump to $c$ instead of to $d$. Similarly $\neg a \vee \neg c \vee \neg e <_P \neg b \vee \neg d \vee \neg e$; once the common element $\neg e$ is eliminated, we would still rather backtrack to $c$ than to $d$. In general, our goal when resolving two augmented clauses is to select a resolvent that is minimal under $<_P$. Note that we have:

**Lemma 8.3** *If $c_1 \subset c_2$ are two nogoods for $P$, then $c_1 <_P c_2$.*

**Procedure 8.4** *Suppose we are given two augmented clauses $(\alpha, G)$ and $(\beta, H)$ that are unit for a partial assignment $P = \langle l_1, \ldots, l_n \rangle$, with $l \in \alpha$ and $\neg l \in \beta$. To find a $<_P$-minimal $l$-resolvent of $(\alpha, G)$ and $(\beta, H)$:*

```
 1   U ← {l, ¬l}          ▷ literals you can't avoid
 2   α_f ← α
 3   β_f ← β
 4   p ← [(α ∪ β) − U]^{?P}
 5   while p > 0
 6       do g ← transport(G, α, {¬l_p, . . . , ¬l_n}) − U, ∅, 0, l)
 7          h ← transport(H, β, {¬l_p, . . . , ¬l_n}) − U, ∅, 0, ¬l)
 8          if g = FAILURE ∨ h = FAILURE
 9             then U ← U ∪ {¬l_p}
10             else  α_f ← α^g
11                   β_f ← β^h
12          p ← [(α_f ∪ β_f) − U]^{?P}
13   return resolve(α_f, β_f)
```

The basic idea is that we gradually force the two clause instances away from the end of the partial assignment; as we back up, we keep track of literals that are unavoidable because an associated call to `transport` failed. The unavoidable literals are accumulated in the set $U$ above, and as we continue to call the transporter function, we have no objection if one or both of the clause instances includes elements of $U$. At each point, we refocus our attention on the deepest variable that is not yet known to be either avoidable or unavoidable; when we reach the root of the partial assignment, we return the instances found.

Here is an example. Suppose that $P = \langle a, b, c, d, e \rangle$ as before, and that $(\alpha, G)$ has instances $\neg c \vee \neg d \vee l$ and $\neg a \vee \neg e \vee l$. The second clause $(\beta, H)$ has the single instance $\neg b \vee \neg e \vee \neg l$.

If we resolve the $<_P$-minimal instances of the two augmented clauses, we will resolve $\neg c \vee \neg d \vee l$ with $\neg b \vee \neg e \vee \neg l$ to get $\neg b \vee \neg c \vee \neg d \vee \neg e$. We do better if we resolve $\neg a \vee \neg e \vee l$ and $\neg b \vee \neg e \vee \neg l$ instead to get $\neg a \vee \neg b \vee \neg e$. The literals $\neg b$ and $\neg e$ appear in any case, but we're better off with $\neg a$ than with $\neg c \vee \neg d$.

Suppose that we follow this example through the procedure, with $U$ initially set to $\{l, \neg l\}$ and (say) $\alpha$ and therefore $\alpha_f$ set to $\neg c \vee \neg d \vee l$. Both $\beta$ and $\beta_f$ are set to $\neg b \vee \neg e \vee \neg l$, since this is the only instance of $(\beta, H)$. The initial value for $p$ is five, since the last literal in $\alpha \cup \beta - U = \{\neg b, \neg c, \neg d, \neg e\}$ is $\neg e$.

We now try to find a way to avoid having $\neg e$ appear in the final resolvent. We do this by looking for an instance of $(\alpha, G)$ that includes $l$ (the literal on which we're resolving) and avoids $\neg e$ (and any subsequent literal, but there aren't any). Such an instance is given by $\alpha$ itself. But there is no instance of $(\beta, H)$ that avoids $\neg e$, so the call in line 7 fails. We therefore add $\neg e$ to $U$ and leave the clauses $\alpha_f$ and $\beta_f$ unchanged. We decrement $p$ to four, since $\neg e$ is no longer in $(\alpha_f \cup \beta_f) - U$.

On the next pass through the loop, we are looking for clause instances that avoid $\{\neg d, \neg e\} - U = \{\neg d\}$. We know that we'll be forced to include $\neg e$ in the final result, so we don't worry about it. All we hope to do at this point is to exclude $\neg d$.

Here, we are successful in finding such instances. The existing instance $\beta$ suffices, as does the other instance $\neg a \vee \neg e \vee l$ of $(\alpha, G)$. This becomes the new $\alpha_f$ and $p$ gets reduced to two, since we now have $(\alpha_f \cup \beta_f) - U = \{\neg a, \neg b\}$.

The next pass through the loop tries to avoid $\neg b$ while continuing to avoid $\neg c$ and $\neg d$ (which we know we can avoid because the current $\alpha_f$ and $\beta_f$ do so). This turns out to be impossible, so $\neg b$ is added to $U$ and $p$ is decremented to one. Avoiding $\neg a$ is impossible as well, so $p$ is decremented to zero and the procedure correctly returns $\neg a \vee \neg b \vee \neg e$.

**Proposition 8.5** *Suppose that we are given two augmented clauses $(\alpha, G)$ and $(\beta, H)$ such that $\alpha$ and $\beta$ are unit for a partial assignment $P$, with $l \in \alpha$ and $\neg l \in \beta$. Then the value returned by Procedure 8.4 is a $<_P$-minimal $l$-resolvent of $(\alpha, G)$ and $(\beta, H)$.*

The procedure can be implemented somewhat more efficiently than described above; if $\alpha_f$, for example, already satisfies the condition implicit in line 6, there is no need to reinvoke the `transport` function for $g$.

More important than this relatively slender improvement, however, is the fact that resolution now involves repeated calls to the `transport` function. In general, Boolean satisfiability engines need not worry about the time used by the resolution function, since unit propagation dominates the running time. A naive implementation of Procedure 8.4, however, involves more calls to `transport` than does the unit propagation procedure, so that resolution comes to dominate ZAP's overall runtime.

To correct this, remember the point of Procedure 8.4. The procedure is not needed for correctness; it is only needed to find improved resolution instances. The amount of time spent looking for such instances should be less than the computational savings achieved by having them. Put slightly differently, there is no requirement that we produce a resolvent that is *absolutely* minimal under the $<_P$ ordering. A resolvent that is nearly minimal will suffice, especially if producing the truly minimal instance involves large computational cost.

We achieve this goal by working with a modified `transport` function on lines 6 and 7 of Procedure 8.4. Instead of expanding the coset decomposition tree completely, a limited number of nodes are examined. ZAP's current implementation prunes the transporter search after 100 nodes have been examined; in solving the pigeonhole problem, for example, this turns out to be sufficient to ensure that the resulting proof length is the same as it would have been had strictly $<_P$-minimal resolvents been found. We also modify the pruning computation, pruning with $K = G_{S \cup U}$ instead of the more difficult to compute $G_{\{S, U\}}$. Since $G_{S \cup U} \leq G_{\{S, U\}}$ (stabilizing every element of a set surely stabilizes the set itself), this approximation saves time but reduces the amount of possible pruning. This is appropriate given the artificially reduced size of the overall search tree and the need to produce an answer quickly.

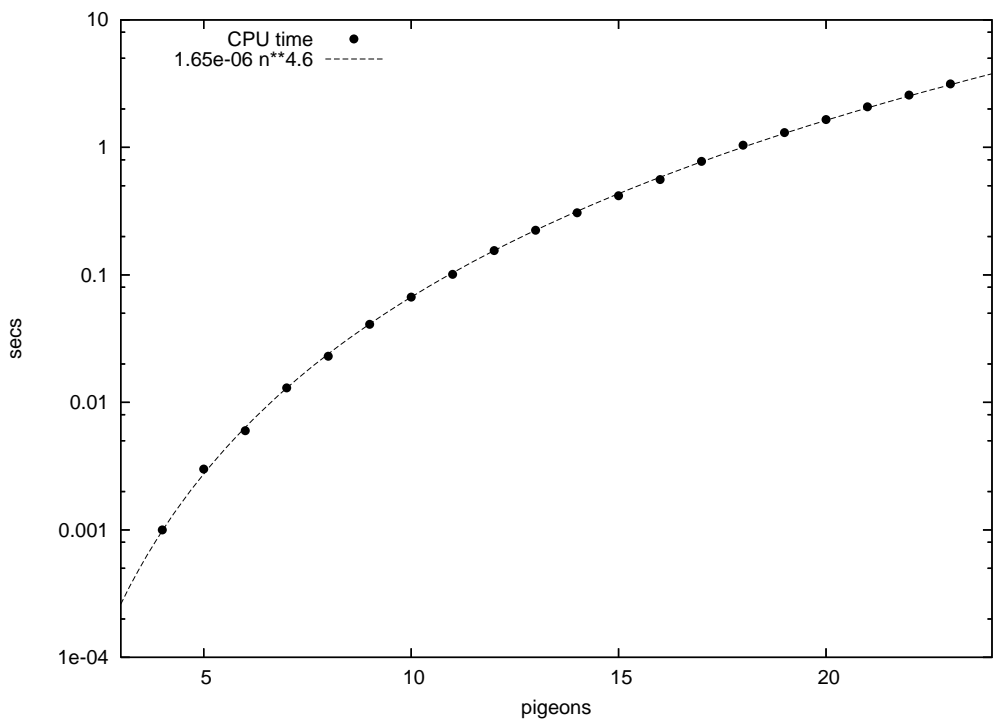

Figure 2: CPU time for a resolution in the pigeonhole problem

## 9. Experimental Results: Components

We are finally in a position to describe the experimental performance of the algorithms that we have presented. As remarked in the introduction, we begin by describing the performance of ZAP's algorithmic components, its resolution and unit propagation algorithms. Performance results for a complete inference tool build using our ideas are in the next section. All experiments were performed on a 2GHz Pentium-M with 1GB of main memory.

### 9.1 Resolution

We have implemented the resolution procedure described in Section 4, and the results for the pigeonhole problem are shown in Figure 2. This particular example involves resolving the two basic axioms in a pigeonhole problem containing $n$ pigeons and $n - 1$ holes:

$$(p_{11} \vee \cdots \vee p_{1,n-1}, G)$$
$$(\neg p_{11} \vee \neg p_{12}, G)$$

The first axiom says that pigeon 1 must be in some hole; the second, that the first two pigeons cannot both be in the first hole. The group $G$ corresponds to a global symmetry group where pigeons and holes can be interchanged freely.

The resolvent of the above two axioms can in fact be computed without any group-theoretic computation at all, using the result from ZAP2 that the group of stable extensions of $(c_1, G)$ and $(c_2, G)$ is always a superset of the group $G$. The algorithm in Section 4 for computing augmented resolvents does not include a check to see if the groups are identical, but the implementation does include such a check. This test was disabled to produce the data in Figure 2.

We plot the observed time (in seconds) for the resolution as a function of the number of pigeons involved, with time plotted on a log scale. Memory usage was typically approximately 5MB; the CPU usage was dominated by the need to compute stabilizer chains for the groups in question. The algorithms used for doing so take time $O(d^5)$ where $d$ is the size of the domain on which the group is operating (Furst, Hopcroft, & Luks, 1980; Knuth, 1991). In this case, the symmetries over pigeons and over holes can be stabilized independently and we therefore expect the stabilizer chain computation to take time $O(n^5)$, where $n$ is the number of pigeons. We fit the data to the curve $ax^b$, with the best fit occurring for $b \approx 4.6$. This is consistent with the stabilizer chain computation dominating the runtime.

If we reinsert the check to see if the groups are the same, the running times are reduced uniformly by approximately 35%. Testing group equality involves checking to see if each generator of $G_1$ is a member of $G_2$ and vice versa, and therefore still involves computing stabilizer chains for the groups in question. Once again, the need to compute the stabilizer chains dominates the computation.

## 9.2 Unit Propagation

In Figure 3 we give data showing the average time needed for a unit test in the pigeonhole problem. These are the "naturally occurring" unit tests that arise in a run of the prover on the problem in question. The memory used by the program remained far less than the 1GB available; as an example, maximum usage was approximately 20MB for 13 pigeons.[12]

Since the unit test is NP-complete, it is customary to give both mean and median running times; we present only means in Figure 3 because the mean running times appear to be growing polynomially (compare the two lines of best fit in the figure), and because the medians appear to be only modestly smaller than the means. This is shown in Figure 4, where it appears that the ratio of the mean to median running times is growing only linearly with problem size.

The earlier figure 3 also shows the average CPU time for "failed" tests (where the clause in question has no unit instances) and "successful" tests (where unit instances exist); as can be seen, failed unit tests generally complete far more quickly than their successful counterparts of similar size as the various pruning heuristics come into play. In both cases, however, the scaling continues to appear to be polynomial in the problem size.

---

12. Accurately measuring peak memory usage is difficult because the group operations regularly allocate and free relatively large blocks of memory. We measured the usage by simply starting a system monitor and observing it, which was not practical for problem instances that took extended amounts of time to complete. This is the reason that we report memory usage only approximately, and only for one problem instance.

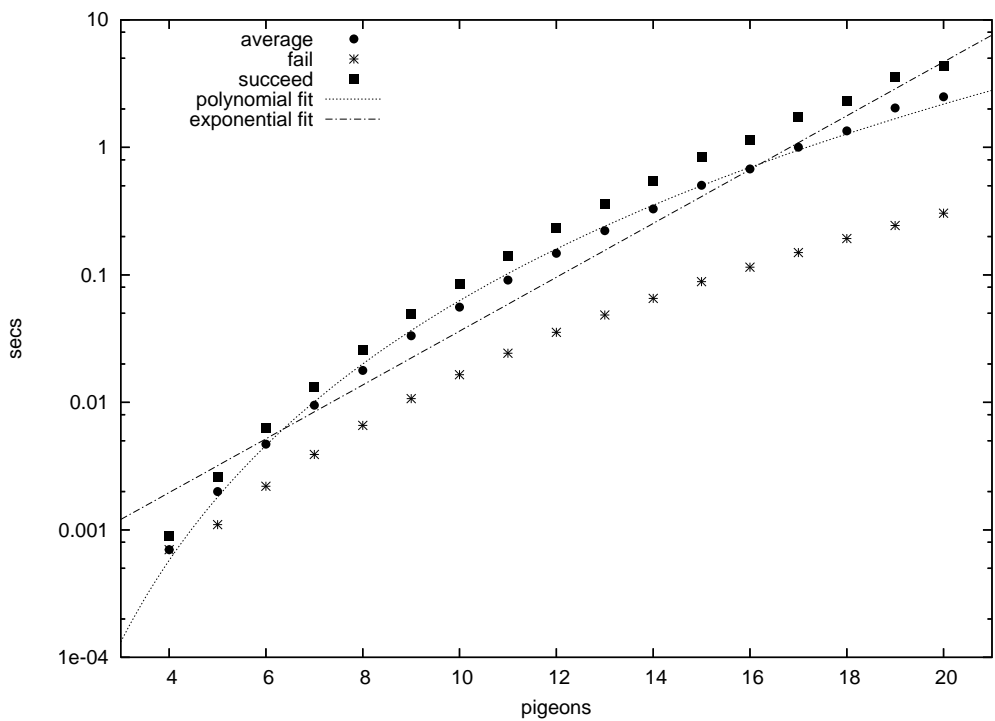

Figure 3: CPU time for a unit test in the pigeonhole problem

## 10. Experimental Results: ZAP

We conclude our discussion of ZAP's experimental performance with results on problem instances in their entirety, as opposed to the performance of individual algorithmic components. Before presenting the results, however, let us describe both the domains being considered and our expectations with regard to performance of both ZAP and of existing systems in these areas.

We will be examining performance in three domains:

1. In a *pigeonhole problem*, the goal is to show that you cannot put $n + 1$ pigeons into $n$ holes if each pigeon is to get its own hole.

2. In a *parity problem*, the goal is to show that $\sum_{i \in I} x_i + \sum_{i \in J} x_i$ cannot be odd if the sets $I$ and $J$ are equal (Tseitin, 1970).

3. In a *clique-coloring problem*, the goal is to show that a map containing an $m$-clique cannot be colored in $n$ colors if $n < m$.

The reasons that we have chosen these particular problem classes are as follows:

1. They all should be easy. It's "obvious" that you can't put $n + 1$ pigeons into $n$ holes, and that $\sum_{i \in I} x_i + \sum_{i \in J} x_i$ is even if each $x_i$ appears exactly twice. It's also obvious that you can't color a graph containing an $m$-clique user fewer than $m$ colors.

   In this last case especially, note that we are solving an easy problem. It is not the case that we are trying to color a *specific* graph containing an $m$-clique; the goal is

494

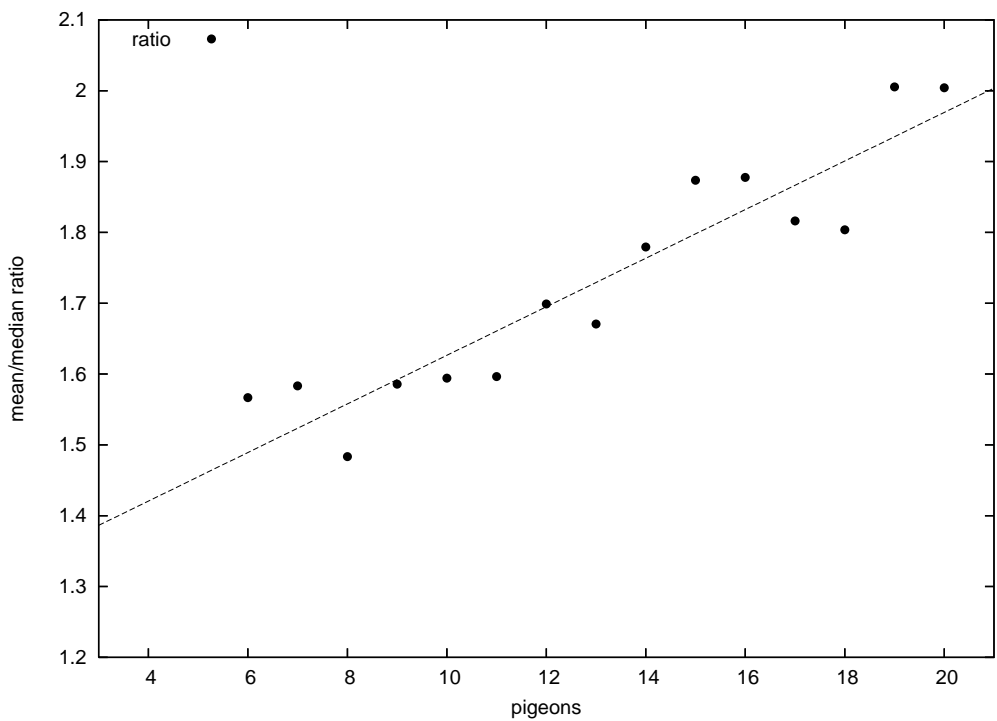

Figure 4: Mean vs. median CPU time for a unit test in the pigeonhole problem

to show that any graph containing an *m*-clique *anywhere* cannot be colored. This is very different from graph coloring generally.

Put somewhat differently, all of the problems that we will be examining are in P. Given suitable representations, they should all be easy.

2. All of the problems are known to be exponentially difficult for resolution-based methods. This was shown for pigeonhole problems by Haken (1985) and for parity problems by Tseitin (1970). Clique-coloring problems are known to be exponentially difficult not only for resolution, but for linear programming methods as well (Pudlak, 1997). In fact, we know of *no* implemented system that scales polynomially on this class of problem.

3. Finally, all of these problems involve structure that can be captured in a group-based setting.

The data that we will present compares ZAP's performance to that of zCHAFF; Section 10.4 discusses the performance of some other Boolean tools on the problem classes that we will be discussing. We chose zCHAFF for comparison partly because it has been discussed throughout this series of papers, and partly because it appears to have the best overall performance on the three problem classes that we will be considering. (Once again, see Section 10.4 for additional details.)

**ZAP expectations**  Before proceeding, let us point out that on a theoretical basis, it is known that short group-based proofs exist for all of these problems. We showed in ZAP2 that group-based pigeonhole proofs can be expected to be short, and also that all parity problems have short group-based proofs that mimic Gaussian elimination. We also showed that short group-based proofs existed for clique coloring, although the proof was fairly intricate. Our goal here is to determine whether an implementation of our ideas can discover these short proofs in practice, or whether the control of group-based inference will require additional theoretical ideas that we do not yet understand.

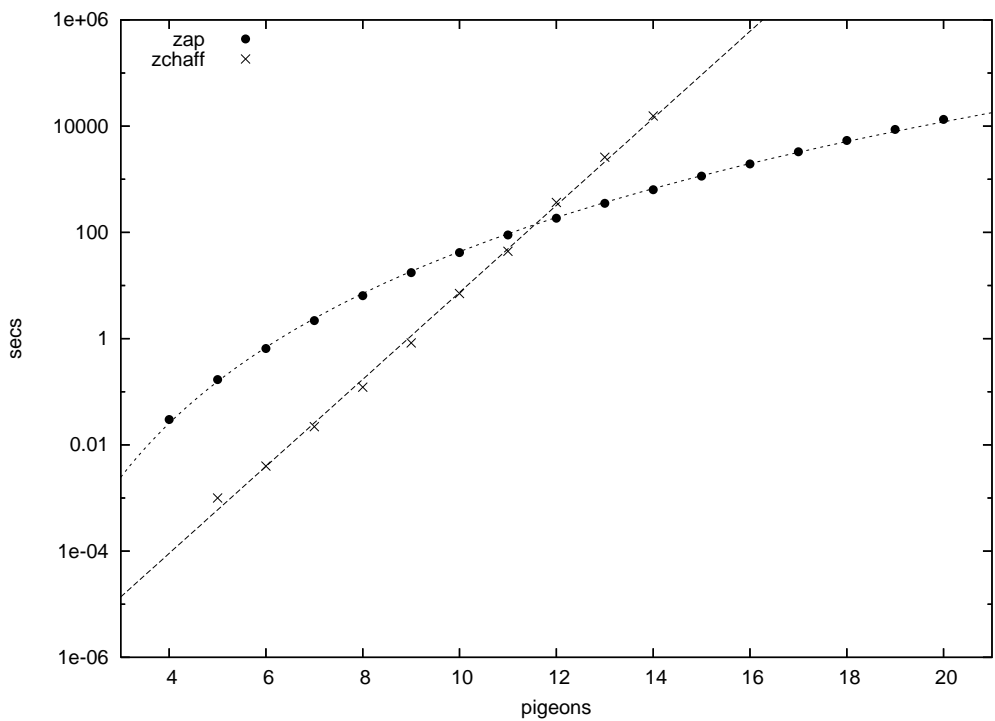

Figure 5: CPU time for pigeonhole instances, ZAP and ZCHAFF

Please understand that our goal at this point is *not* to test ZAP on standard NP-complete search problems in Boolean form, such as graph coloring or quasigroup completion problems (Gomes & Selman, 1997). Doing so involves a significant effort in ensuring that ZAP's constant factors and data structures are comparable to those of other systems; while preliminary indications are that this will be possible with only modest impact on performance (approximately a factor of two), the work is not yet complete and will be reported elsewhere.

## 10.1 Pigeonhole Results

Figure 5 shows running times for both ZAP and for ZCHAFF on pigeonhole instances. Figure 6 repeats the ZAP data, also including best exponential and polynomial fits for the time spent. The overall running time appears to be polynomial, varying as approximately $n^{8.1}$ where $n$ is the number of pigeons. In very rough terms, there is a factor of $O(n^5)$ needed for the stabilizer chain constructions. If we branch only on positive literals, we know (see ZAP2) that there will be $O(n)$ resolutions needed to solve the problem, and each resolution will lead to $O(n^2)$ unit propagations. The total time can thus be expected to be approximately $O(n^8)$, assuming that each unit propagation involves only stabilizer chain computations and no actual search. Our observed performance is close to this theoretical value.

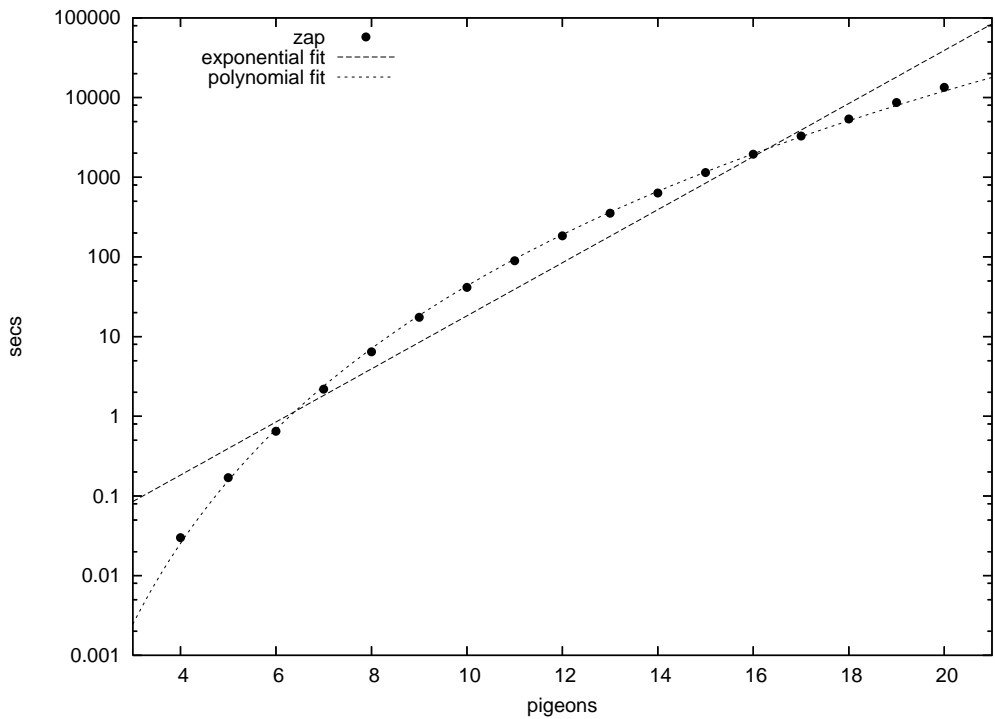

Figure 6: zap scaling for pigeonhole instances

In practice, zap branches not on positive literals, but on negative ones. The reason is that the negative literals appear in far more clauses than the positive ones ($O(n)$ clauses for each negative literal as opposed to a single clause for a positive literal), and the usual branching heuristic in the Boolean satisfiability community initially assigns to a variable the value that satisfies as many clauses as possible.

The number of nodes expanded by zap in solving any particular instance of the pigeonhole problem is shown in Figure 7, which also presents similar data for zChaff. The number of nodes expanded by zap is in fact exactly $n^2 - 3n + 1$; curiously, this is also the *depth* of the zChaff search for the next smaller instance with $n - 1$ pigeons. We do not know if the small size of the pigeonhole proofs found by zap is the result of the effectiveness of the use of $<_P$-optimal resolvents, or if some fundamental argument can be made that all zap proofs of the pigeonhole problem will be short.

Before moving on to parity problems, allow us to comment on the importance of the various algorithmic techniques that we have described. We recognize that many of the algorithms we have presented are quite involved, and it is important to demonstrate that the associated algorithmic complexity leads to legitimate computational gains.

Figure 8 shows the time needed to solve pigeonhole instances if we either abandon the pruning lemmas or avoid the search for $<_P$-optimal resolvents. As should be clear from the data, both of these techniques are essential to obtaining the overall performance exhibited by the system.

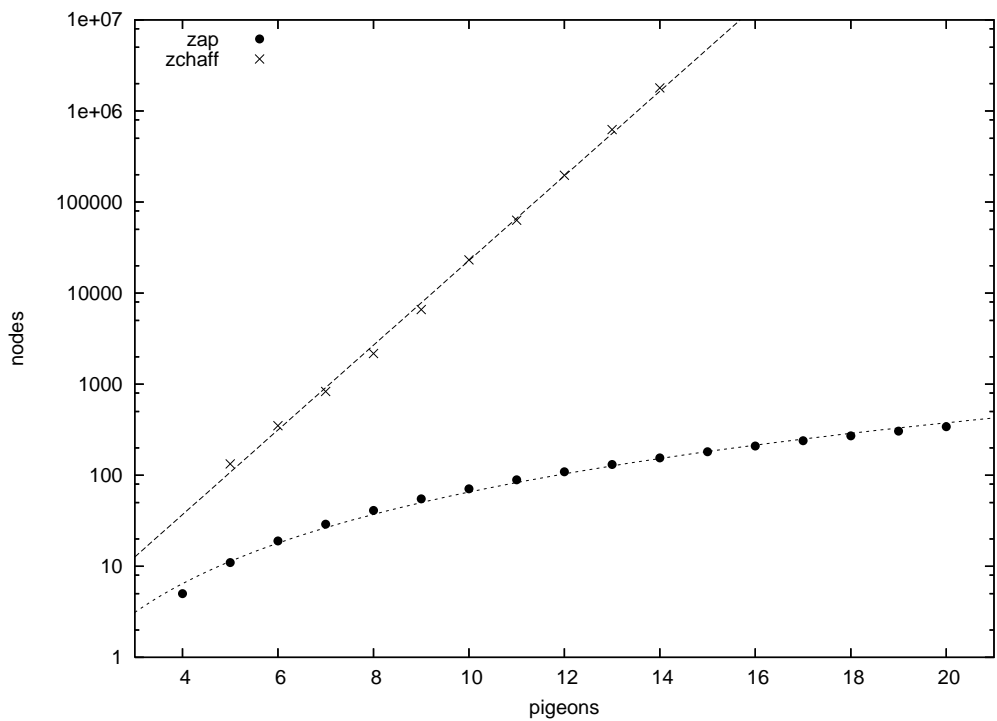

Figure 7: Nodes expanded in the pigeonhole problem

If we abandon the search for $<_P$-optimal resolvents, the proof lengths increase significantly but appear to remain polynomial in $n$. The length increase in the learned axioms leads to increased running times for unit propagation, and this appears to be the primary reason for the performance degradation in the figure. The overall running times scale exponentially.

Abandoning the pruning lemmas also leads to exponential running times. This is to be expected at some level; there are still exponentially many learned ground axioms and if we cannot prune the search for unit instances, exponential behavior is to be expected.

There were other ways that we could have reduced ZAP's algorithmic complexity as well. We could, for example, have removed watched literals and the computational machinery needed to maintain them. As it turns out, this change has virtually no impact on ZAP's pigeonhole performance because the prover's behavior here is typically backtrack-free (Dixon et al., 2004a). In general, however, watched literals can be expected to play as important a role in ZAP as they do in any other DPLL-style prover. Our overall focus in this series of papers has been to show that group-based augmentations could be implemented *without* sacrificing the ability to use any of the recent techniques that have made Boolean satisfiability engines so effective in practice, and watched literals can certainly be numbered amongst those techniques.

We also did not evaluate the possibility of not learning augmented clauses at all, perhaps learning instead only their ground versions. This would avoid the need to implement Procedure 4.1, but would also avoid all of the computational gains to which ZAP theoretically has access. It is only by learning augmented clauses that theoretical reductions in proof size can be obtained; otherwise, the proof itself would necessarily be unchanged from any other DPLL-style approach.

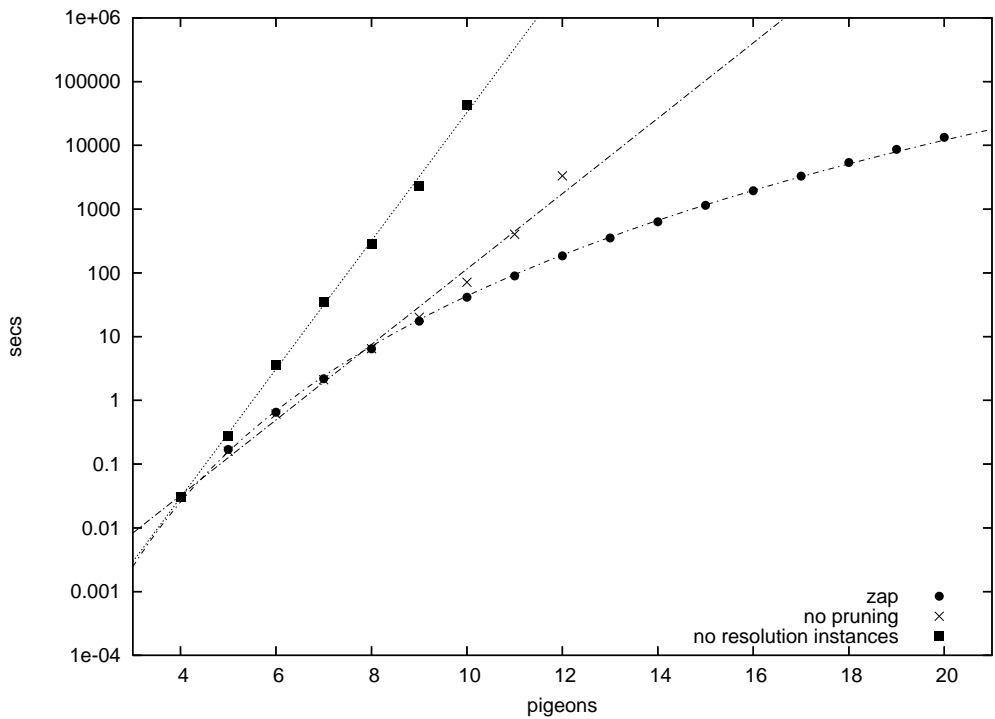

Figure 8: Improvement due to pruning lemmas and $<_P$-optimal resolution instances. The circles mark ZAP's performance. The x's indicate performance with the pruning lemmas disabled during unit propagation, and the boxes give performance if resolutions use the original base instances of the clauses being resolved, as opposed to searching for and resolving $<_P$-optimal instances.

## 10.2 Tseitin Results

The next problem class for which we present experimental data is one due to Tseitin (1970) that was shown by Urquhart (1987) to require resolution proofs of exponential length. Each problem is based on a graph $G$. We associate a Boolean variable with each edge in $G$, and every vertex $v$ in $G$ has an associated charge of 0 or 1 that is equal to the sum mod 2 of the variables adjacent to $v$. The charge of the entire graph $G$ is the sum mod 2 of the charges of its vertices. If we require that a connected graph $G$ have a charge of one, then the set of constraints associated with its vertices is unsatisfiable (Tseitin, 1970). Here is the graph for a problem of size four, together with its associated constraints:

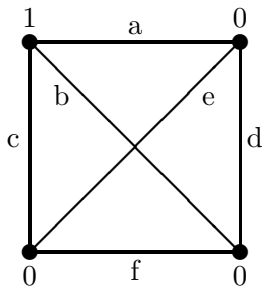

$$a + b + c \equiv 1$$
$$d + e + a \equiv 0$$
$$f + b + d \equiv 0$$
$$c + e + f \equiv 0$$

In the language of ZAP (see Appendix B), we have

```
a b c %2= 1 ;
d e a %2= 0 ;
f b d %2= 0 ;
c e f %2= 0 ;
```

The axiom set is unsatisfiable because adding all of the above axioms gives us

$$2a + 2b + 2c + 2d + 2e + 2f \equiv 1$$

These problems are known to be exponentially difficult for resolution-based methods (Urquhart, 1987).

Times to solution for ZAP and ZCHAFF are shown in Figure 9. ZCHAFF is clearly scaling exponentially; the best fit for the ZAP times is $0.00043n^{1.60 \log(n)}$, where $n$ is the problem size.

Figure 10 shows the number of nodes expanded by the two systems. The number of search nodes expanded by ZAP appears to be growing polynomially with the size of the problem ($O(n^{2.6})$, give or take), in keeping with a result from ZAP2 showing that ZAP proofs of polynomial length always exist for parity problems. As with the pigeonhole instances, we see that short proofs exist not only in theory, but apparently in practice as well.

Given that a polynomial number of nodes are expanded but a super-polynomial amount of time is consumed, it seems likely that the unit propagation procedure is the culprit, taking a super-polynomial amount of time per unit propagation. As shown in Figure 11, this is in fact the case. But the unit test should be easy here – after all, the groups are all simply those that flip an even number of the variables in question. If we want to know if an augmented clause has a unit instance, we find the unvalued variables it contains. If more than one, the clause is not unit. If exactly one, the clause is *always* unit – the variable must be valued so as the make the parity of the sum take the desired value. So there seems to be no reason for the unit tests to be scaling as $n^{\log(n)}$.

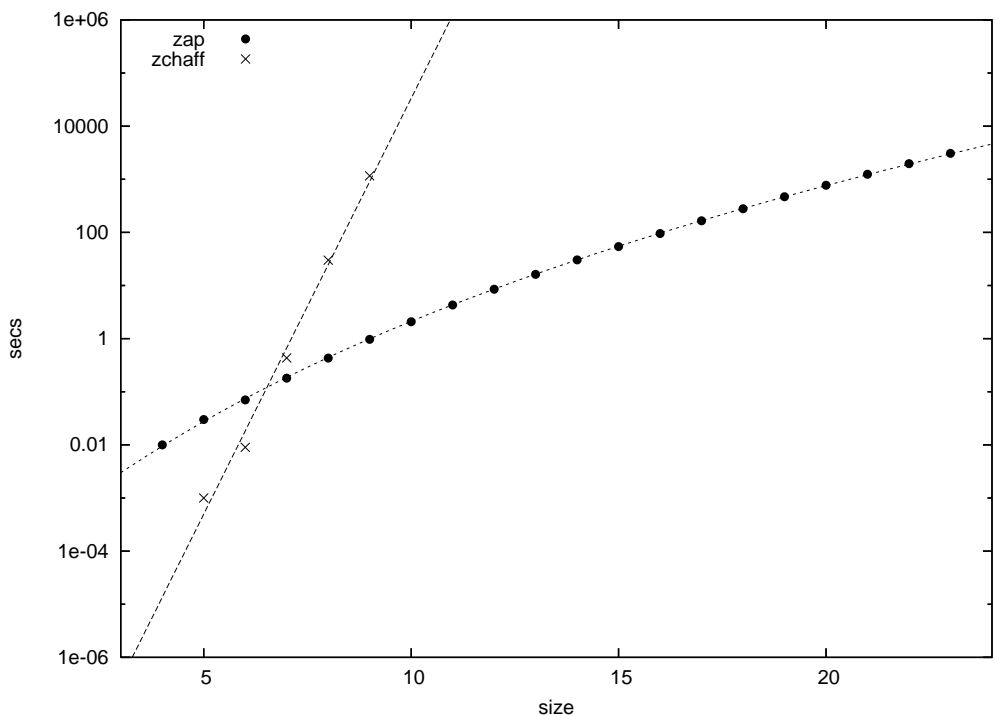

Figure 9: CPU time for Tseitin instances, ZAP and ZCHAFF. ZCHAFF is scaling exponentially; ZAP is scaling as $O(n^{1.6 \log(n)})$.

The $n^{\log(n)}$ scaling itself appears to be a consequence of the multiset stabilizer computation that underlies the $k$-transporter pruning. Here, too, the scaling should be polynomial, since we can show that polytime $(O(n^3))$ methods exist for set stabilizer for the groups in question.[13] The general methods implemented by GAP and by ZAP do not exploit the Abelian nature of the parity groups, however, and the scaling is as shown. An obvious extension of the existing implementation would include more efficient set stabilizer algorithms for these groups.

## 10.3 Clique Coloring

The final problem class for which we present experimental data is that of clique coloring. This class of problems is related to the pigeonhole problem but far more difficult.

As mentioned previously, the domain is that of graph coloring, where two nodes connected by an edge must be assigned different colors. If the graph is a clique of size $m$, then it is obvious that the graph cannot be colored in $m - 1$ colors. This is equivalent to an instance of the pigeonhole problem. But in the clique coloring problem, we are not told that the graph *is* a clique of size $m$, only that it *contains* a clique of size $m$. The fact that we do not know the exact location of the clique widens the search considerably.

---

13. The argument can be made either from the fact that the groups are Abelian, or from the fact that the group orbits are all of size two, and the set stabilizer problem can thus be converted to one of linear algebra over $\mathbb{Z}_2$.

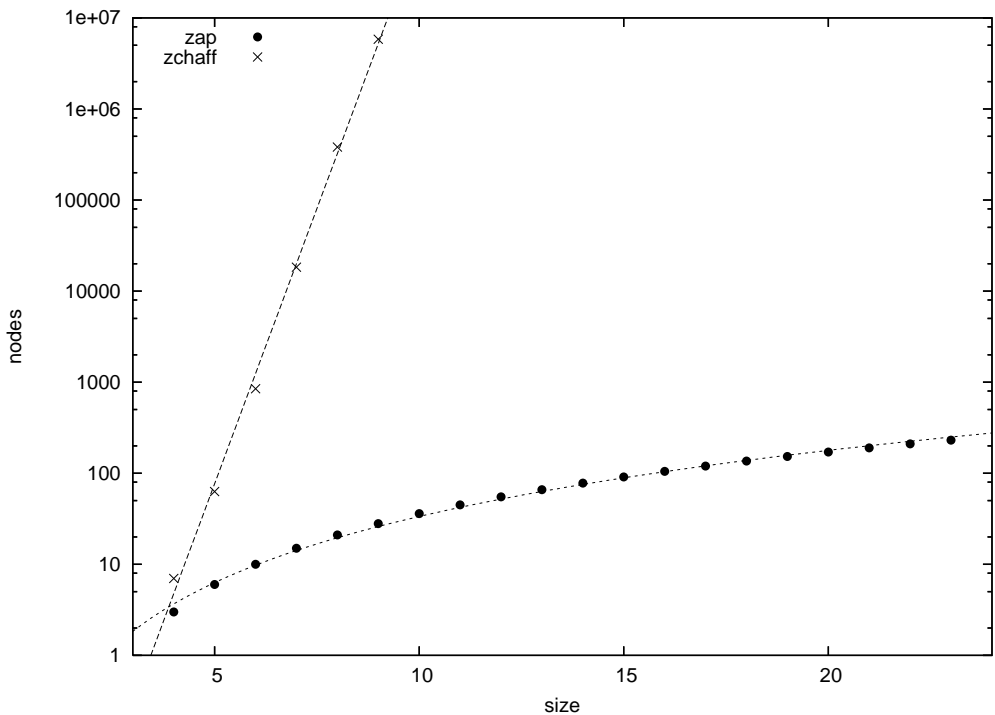

Figure 10: Nodes expanded in the Tseitin problems. ZChaff is scaling exponentially; zap is scaling polynomially as $O(n^{2.6})$.

We know of no (other) automated proof system that scales polynomially on problems in this class; both resolution and linear programming methods inevitably scale exponentially (Pudlak, 1997). We showed in zap2 that zap could produce polynomial-length proofs in theory, but no suggestions were made that such proofs would be easy to find in practice.

Before we present the details of zap's performance on this problem class, let us reiterate our observation that clique-coloring problems should *not* be thought of as unsatisfiable instances of graph-coloring problems generally. A particular instance of this problem class does not describe a specific graph that needs to be colored; it says only that the graph contains an $m$-clique and needs to be colored in $m - 1$ colors.

An axiomatization of this problem is as follows. We use $e_{ij}$ to describe the graph, $c_{ij}$ to describe the coloring of the graph, and $q_{ij}$ to describe the embedding of the clique into the graph. The graph has $m$ nodes, the clique is of size $n + 1$, and there are $n$ colors available.

$$c_{i1} \vee \cdots \vee c_{in} \qquad \text{for } i = 1, \ldots, m \tag{26}$$

$$q_{i1} \vee \cdots \vee q_{im} \qquad \text{for } i = 1, \ldots, n + 1 \tag{27}$$

$$\neg e_{ij} \vee \neg c_{il} \vee \neg c_{jl} \qquad \text{for } 1 \leq i < j \leq m, \, l = 1, \ldots, n \tag{28}$$

$$\neg q_{ij} \vee \neg q_{kj} \qquad \text{for } 1 \leq i < k \leq n + 1, \, j = 1, \ldots, m \tag{29}$$

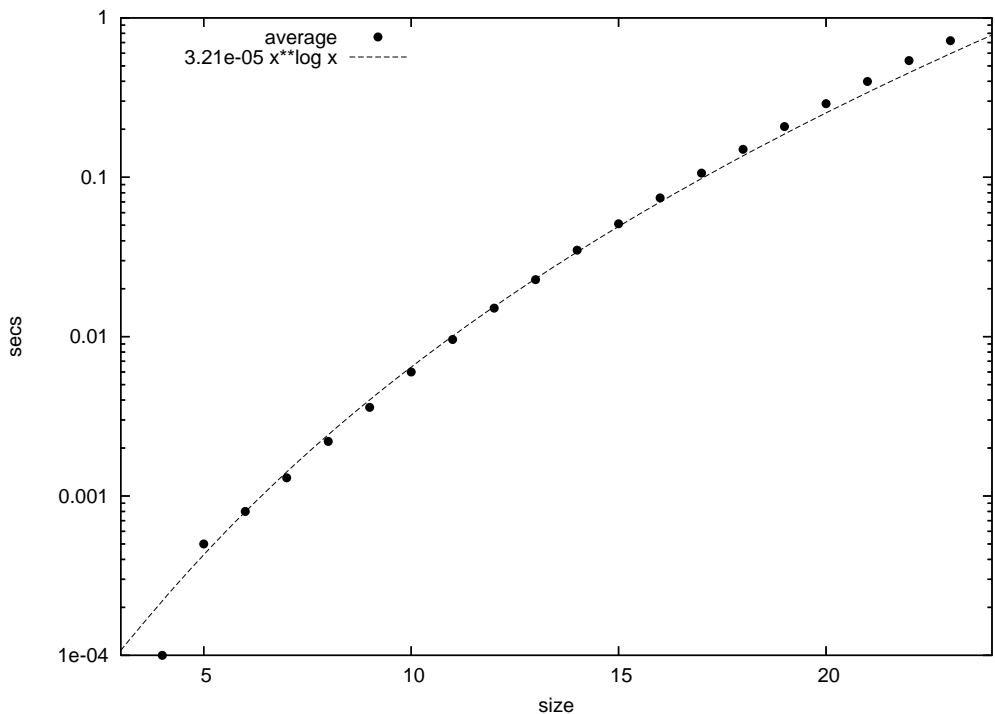

Figure 11: CPU time for a unit test in the Tseitin problems. ZAP is scaling as approximately $O(n^{\log(n)})$.

$$e_{ij} \vee \neg q_{ki} \vee \neg q_{lj} \qquad \text{for } 1 \le i < j \le m,\, 1 \le k \ne l \le n+1 \qquad (30)$$

Here $e_{ij}$ means that there is an edge between graph nodes $i$ and $j$, $c_{ij}$ means that graph node $i$ is colored with the $j$th color, and $q_{ij}$ means that the $i$th element of the clique is mapped to graph node $j$. Thus the first axiom (26) says that every graph node has a color. (27) says that every element of the clique appears in the graph. (28) says that two of the $m$ nodes in the graph cannot be the same color (of the $n$ colors available) if they are connected by an edge. (29) says that no two elements of the clique map to the same node in the graph. Finally, (30) says that the clique is indeed a clique – no two clique elements can map to disconnected nodes in the graph.

The encoding passed to ZAP was group-based, as follows:

```
SORT color 2 ;
SORT node 4 ;
SORT clique 3 ;

PREDICATE edge( node node ) ;
PREDICATE color( node color ) ;
PREDICATE clique( clique node ) ;

GROUP COLOR <
(( color[1 1] color[1 2])
```

```
  ( color[2 1] color[2 2])
  ( color[3 1] color[3 2])
  ( color[4 1] color[4 2]))
> ;
GROUP CLIQUE <
(( clique[1 1] clique[2 1])
 ( clique[1 2] clique[2 2])
 ( clique[1 3] clique[2 3])
 ( clique[1 4] clique[2 4]))
(( clique[2 1] clique[3 1])
 ( clique[2 2] clique[3 2])
 ( clique[2 3] clique[3 3])
 ( clique[2 4] clique[3 4]))
> ;
GROUP NODES <
(( edge[1 3] edge[2 3])
 ( edge[1 4] edge[2 4])
 ( color[1 1] color[2 1])
 ( color[1 2] color[2 2])
 ( clique[1 1] clique[2 1])
 ( clique[2 1] clique[2 2])
 ( clique[3 1] clique[3 2]))
(( color[2 1] color[3 1] color[4 1])
 ( color[2 2] color[3 2] color[4 2])
 ( edge[1 2] edge[1 3] edge[1 4])
 ( edge[2 3] edge[3 4] edge[2 4])
 ( clique[1 2] clique[1 3] clique[1 4])
 ( clique[2 2] clique[2 3] clique[2 4])
 ( clique[3 2] clique[3 3] clique[3 4]))
> ;

color[1 1] color[1 2] GROUP NODES ;
clique[1 1] clique[1 2] clique[1 3] GROUP CLIQUE ;
-edge[1 2] -color[1 1] -color[2 1] GROUP NODES COLOR ;
-clique[1 1] -clique[2 1] GROUP NODES CLIQUE ;
-clique[1 1] -clique[2 2] edge[1 2] GROUP NODES CLIQUE ;
```

This is the version where there is a 3-clique in a graph of size four, and we are trying to use just two colors. The first group is the symmetry over colors alone, the second that over the elements of the clique, and the third the symmetry over nodes. The axiomatization is identical to that presented earlier. Note that although there is a common symmetry in this problem, the axiomatization obscures that in some sense, since we have only included the relevant symmetry or symmetries in any particular axiom.

Times to solution for ZAP and ZCHAFF are shown in Figure 12. As might be expected, ZCHAFF is scaling exponentially; ZAP appears to be scaling as $n^{8.5}$. In order to allow the data to be presented along a single axis, these problem instances were selected so that the clique size was one smaller than the graph size.

Figure 13 shows the number of nodes expanded by the two systems. Once again, the number of nodes expanded by ZCHAFF is growing exponentially with problem size, while the number expanded by ZAP is growing polynomially. As with the pigeonhole problem,

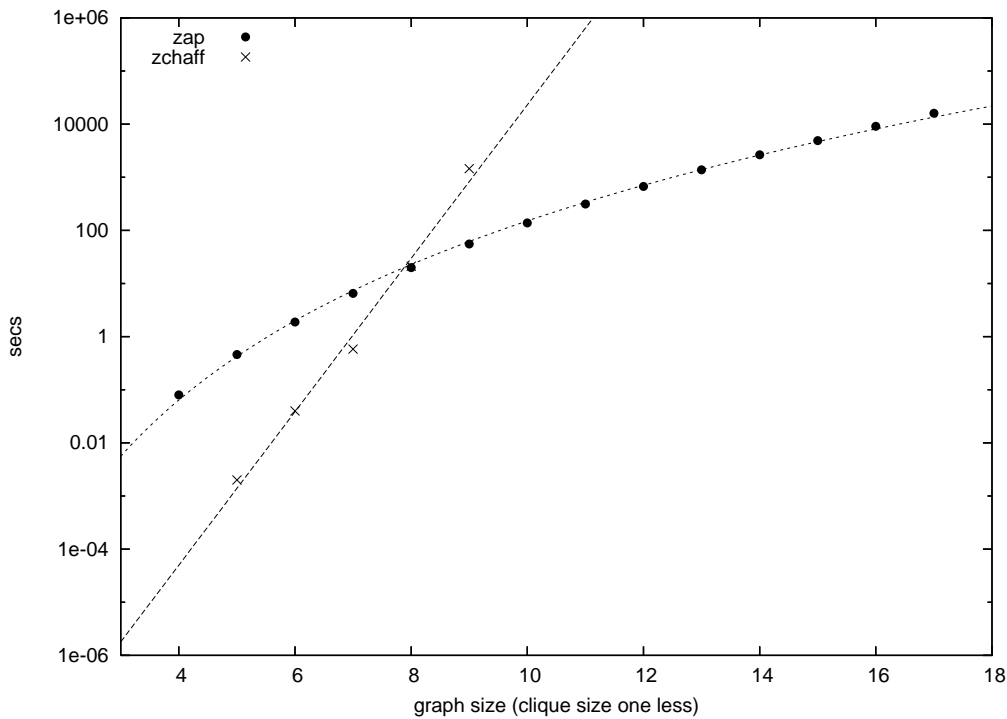

Figure 12: CPU time for clique instances, ZAP and zChaff

we see that the short proofs whose existence is guaranteed by the theory can be found in practice.

Figures 14 and 15 display ZAP's performance on a somewhat wider range of problem instances where the clique and graph sizes are allowed to vary independently. The number of nodes expanded was in general

$$\frac{(c+g)^2 - 13c - g + 14}{2}$$

where $c$ is the size of the clique and $g$ the size of the graph. There were a handful of outliers, most notably the $c = 11, g = 13$ instance which expanded a larger number of nodes. The other exceptions all expanded fewer nodes.

With regard to total CPU time (Figure 15), the time appears to be scaling as $(cg)^{3.89}$. Once again, $c = 11, g = 13$ is an outlier but polynomial performance is observed generally. To the best of our knowledge, ZAP is the first system to exhibit polynomial performance on this problem class; as we have remarked, most other approaches have been proven to scale exponentially.

## 10.4 Related Work

Finally, we compare our experimental results to those obtained using other systems that attempt to exploit problem structure to improve the performance of satisfiability solvers. This section provides a high-level summary of experimental results for a number of these efforts and compares these results with ZAP on the benchmark problems described in the previous sections.

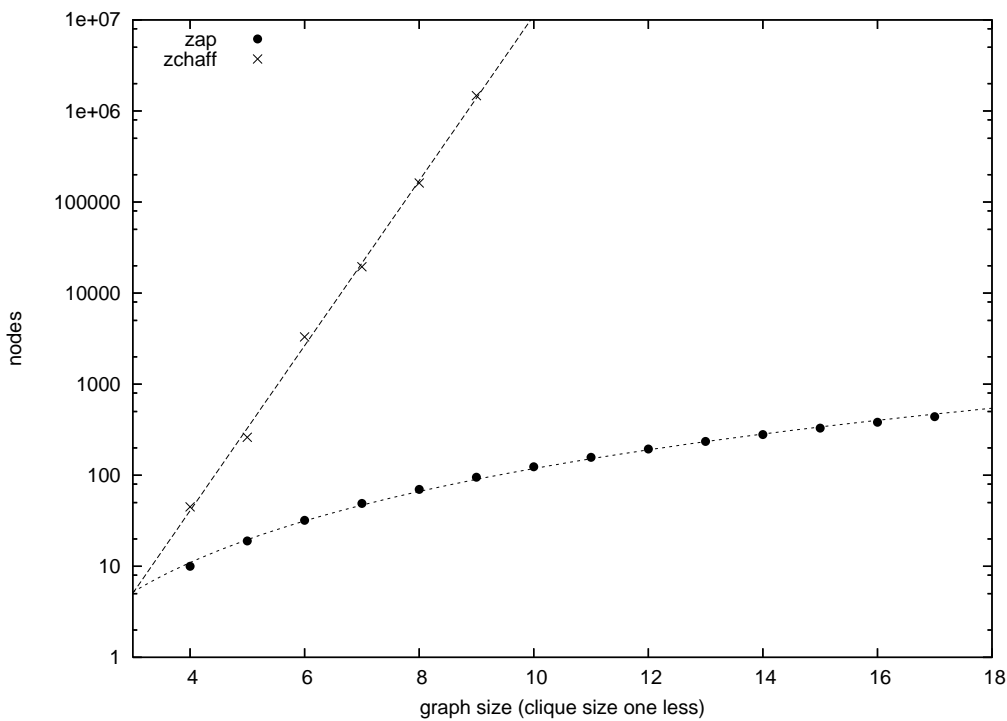

Figure 13: Nodes expanded in the clique problems

Recall that our benchmark problems are all highly structured, but each has a very different type of structure. Theoretically, these problems all allow polynomial-time solutions, but they are provably hard for conventional solvers. A solver that solves all of these problems efficiently has the ability to exploit a range of different types of problem structure and automates a strong proof system. Of course, to be interesting, a solver must also be a practical general purpose solver. For example, Tseitin problems can be solved in polynomial time by a form of Gaussian elimination (Schaefer, 1978), and pigeonhole problems can be solved in polynomial time by a linear programming method such as the simplex method. However, neither of these solutions constitutes a practical general purpose solver.

We ran a number of solvers on the benchmark problems, obtaining the following results:

|  | **pigeonhole** | **Tseitin** | **clique coloring** |
|---|---|---|---|
| ZAP | P | $n^{\log n}$ | P |
| zCHAFF | E | E | E |
| PBCHAFF | P | E | E |
| EQSATZ | E | E | E |
| MARCH_EQ | E | E (P) | E |
| resolution | E | E (?) | E |
| cutting-planes or integer programming | P | ? | E |

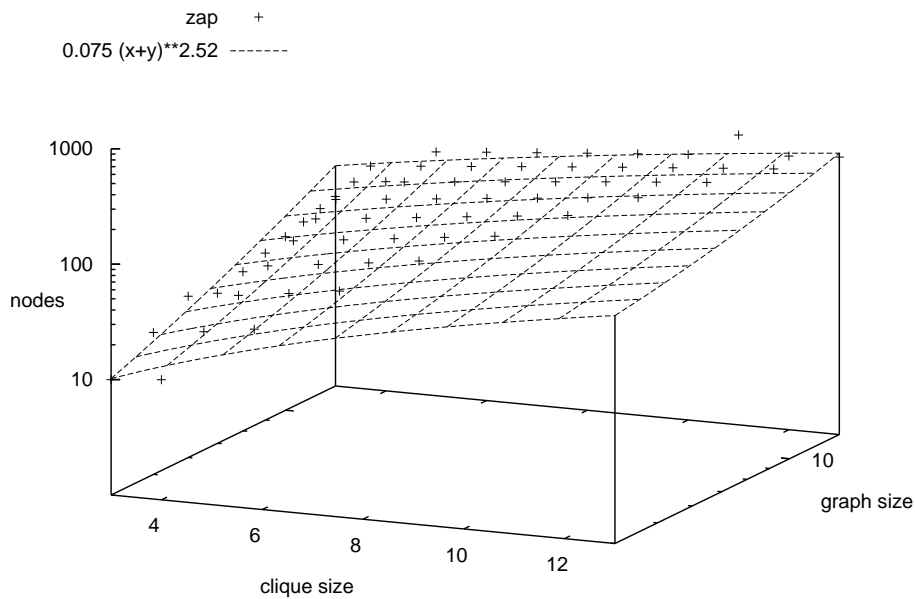

Figure 14: Nodes expanded in the clique problems

Rather than presenting numerous graphs, we summarize our results above, simply reporting the overall scaling of each solver on each problem class. Polynomial-time scaling is indicated with a P and exponential-time scaling with an E. Scaling is shown for the three problem classes we have discussed, with two separate encodings considered for the Tseitin problems. The first encoding is the Booleanization of the encoding of Section 10.2; the second involves the introduction of new variables to reduce clause length and is described below. If the performance is improved by this introduction, the new scaling is given parenthetically. The final two rows give known proof complexity results for the resolution and cutting-planes proof systems and thus provide lower bounds on the corresponding rows above them.

Reducing performance results to exponential or polynomial scaling omits valuable information. Clearly the difference between $n^{100}$ and $n^2$ scaling is something we care about, although both are polynomial. The details of specific scaling factors will be included in the discussion that follows; our goal in the table is merely to summarize the strength of each solver's underlying proof system.

Details of the solvers appearing in the table are as follows:

- PBCHAFF is a pseudo-Boolean version of the DPLL algorithm. It represents problems in pseudo-Boolean form and automates a cutting-planes proof system. The cutting-planes proof system allows polynomial-length proofs of the pigeonhole problem and PBCHAFF is able to solve these problems efficiently. Scaling for PBCHAFF on pigeonhole instances was as $n^{4.8}$, where $n$ is the number of pigeons. This is an improvement over the $n^{8.1}$ scaling seen for ZAP. However, the performance of PBCHAFF on Tseitin and clique coloring problems is exponential, since cutting-planes inference is not able to capture and exploit the structure of these problems.

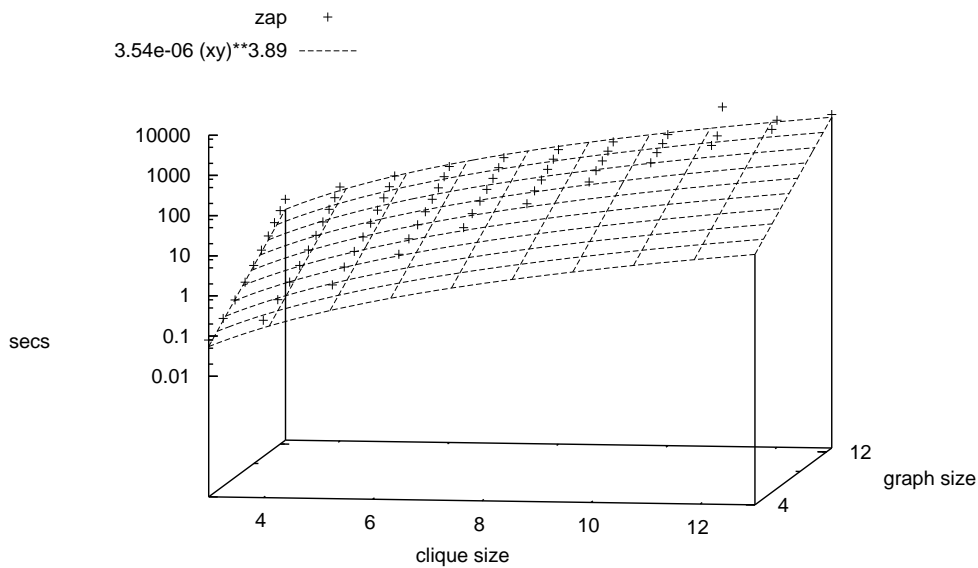

Figure 15: CPU time expended in the clique problems

- EQSATZ (Li, 2000) and MARCH_EQ (Heule & van Maaren, 2004) are DPLL-based solvers that have been modified to incorporate equivalence reasoning, which should enable them to solve parity problems efficiently. As expected, both EQSATZ and MARCH_EQ exhibited exponential scaling on pigeonhole and clique coloring problems, since these solvers are not designed to recognize the structure of these problems. More surprising was the exponential scaling observed for both EQSATZ and MARCH_EQ on our initial encoding of the Tseitin problems.

Eqsatz scales exponentially because it does not recognize the structure present in our encoding of the parity problems.[14] This performance can be improved by modifying the CNF encoding to reduce its size and make the structure more apparent to the solver. Doing so involves the introduction of a significant number of new auxiliary variables, and experimental results for this new encoding are discussed below. MARCH_EQ does recognize the structure in our original encoding, and solves it during a preprocessing phase. The exponential scaling here is due simply to the fact that the *size* of the Boolean encoding is growing exponentially as a function of graph size (see Section 10.2).

---

14. Li, personal communication (2005).

Any parity constraint can be rewritten as a set of parity constraints, each of length at most three (Li, 2000). A parity constraint of the form

$$x_1 + x_2 + \ldots + x_n \equiv k \qquad (31)$$

is equivalent to the set of parity constraints

$$
\begin{aligned}
x_1 + A_1 &\equiv k \\
A_1 + x_2 + A_2 &\equiv 0 \\
A_2 + x_3 + A_3 &\equiv 0 \\
&\vdots \\
A_{n-2} + x_{n-1} + A_{n-1} &\equiv 0 \\
A_{n-1} + x_n &\equiv 0
\end{aligned}
$$

Summing over this set of parity constraints gives

$$2A_1 + 2A_2 + \cdots + 2A_{n-1} + x_1 + \cdots + x_n \equiv k$$

which is equivalent to (31). If the Tseitin encoding from Section 10.2 is translated into parity constraints in this way and then converted to CNF, the exponential blowup in the size of our existing CNF encoding can be avoided. (It is not clear, however, if resolution can then produce a polynomially sized proof of the unsatisfiability of the resulting theory.)

EQSATZ, MARCH_EQ and ZAP all exhibit improved performance if this new encoding is used; these results are shown parenthetically in the **Tseitin** column of the table. MARCH_EQ solves this encoding of the Tseitin problems virtually instantaneously. EQSATZ now substantially outperforms zCHAFF, as reported by Li (2003). The running times for EQSATZ, however, remain exponential and the system is unable to solve the instance of size ten within 10,000 seconds. The performance of ZAP is improved as well, but the overall scaling is unchanged.

The introduction of new variables is accepted practice for reducing the size of CNF encodings, and also has the potential to reduce the length of proofs constructed by solvers. Indeed, there are no classes of problems known to be hard for extended resolution, a version of resolution in which the introduction of new variables is permitted. In general, however, introducing new variables in order to reduce proof length is considered "cheating" from a proof complexity perspective; once new variables are introduced, most proof systems are essentially equivalent. In addition, no general method for introducing variables is known and we know of no implemented system that does so. One advantage of ZAP is that group-based annotations avoid the need for syntactic reworkings of this sort.

Another approach to solving highly symmetric problems is seen in the solver sSATZ (Li, Jurkowiak, & Purdom, Jr., 2002). This solver is also based on the DPLL algorithm, and accepts as input both a problem in CNF and a set of matrices describing a global symmetry on the variables. The global symmetry is then used to partition the set of variable assignments into equivalence classes. In addition to the normal pruning techniques used in DPLL, search can now also be pruned by eliminating any partial assignment that is not minimal under the equivalence corresponding to the global symmetry. sSATZ scales polynomially on pigeonhole problems; however, the class of input symmetry groups allowed by sSATZ is currently too limited to be applied to Tseitin or clique coloring problems. It is not clear whether this is a limitation that can be overcome as the work matures, which is why we have not included sSATZ in our table.

Of all the solvers tested, ZAP is the only solver to provide efficient solutions on all the test problems, and it is the only solver that scales polynomially on clique coloring. PBCHAFF has better scaling on pigeonhole problems, and MARCH_EQ has better scaling on Tseitin problems; however, both solvers exploit a narrowly defined type of problem structure and therefore perform poorly in the other domains. The performance of ZAP is also likely to improve as the basic group primitives underlying ZAP's procedures are optimized.

## 11. Conclusion and Future Work

ZAP represents what appears to be a new synthesis between two very distant fields: computational group theory and Boolean satisfiability. From an algorithmic point of view, each of these fields is fairly mature and complex, and our synthesis inherits significant algorithmic complexity as a result. Our goal in this paper has been to present initial versions of the algorithms that a group-based theorem prover will need, and to describe the performance of a prototype implementation of these ideas. As we have seen, ZAP easily outperforms its conventional counterparts on difficult problem instances where there is group structure concealed by the Boolean axiomatization.

That said, it is important to realize that our results only scratch the surface of what ZAP's underlying representational shift allows. On the Tseitin problems, for example, it seems likely that incorporation of more sophisticated set stabilizer algorithms will allow us to improve ZAP's performance; the fact that only polynomially many nodes are expanded in solving these problems bodes well for the eventual performance of the system.

Other improvements are also possible. In the pigeonhole and clique coloring problems, computational performance is dominated by the $O(n^5)$ stabilizer chain computations on the groups in question; these groups are products of full symmetry groups. It is well known that full symmetry groups are extremely difficult for the usual stabilizer chain algorithms, but in cases such as these it is possible to produce the stabilizer chains directly, taking time $O(n^3)$ or even $O(n^2)$ if the stabilizer chain data structure is modified (Jerrum, 1986). Such modifications can be expected to improve ZAP's performance significantly in this domain.

There is simply too much to do. The above extensions are only the beginning; we also obviously need to experiment with ZAP on a wide range of other problem instances. There are also two general points that we would like to make regarding future work in this area.

First, we have left unmentioned the problem of discovering group structure in existing clausal databases. The practical impact here would be substantial, for several reasons. It would make it possible to apply ZAP directly to problems that have already been encoded using Boolean axioms, and it would also make it possible to discover "emergent" group structure that only appears after search has begun. As an example, perhaps a symmetry exists for a particular problem but is hidden by the existing axiomatization; after a few inferences, the symmetry may become apparent but still needs to be noticed.

Second, and perhaps most important, ZAP provides us with a very broad stage on which to work. Progress in computational group theory can be expected to lead to performance improvements in inference; dually, applying ZAP to a wide range of reasoning problems should provide a new set of examples that the computational group theorists can use to test their ideas. Lifting heuristics from one area of AI to a group-based setting may make analogs of those heuristics available in other, more practical domains. As with all new syntheses, it seems reasonable to hope that ZAP will allow ideas from Boolean satisfiability, computational group theory and search-based AI to be combined, leading to new insights and levels of performance in all of these areas.

## Acknowledgments

We would like to thank the members of CIRL and the technical staff of On Time Systems for their assistance with the ideas in this series of papers. We would also like to thank the implementers and maintainers of GAP; many elements of the ZAP implementation are based directly on either the implementations that appear in GAP or the descriptions in Seress' book (2003). Finally, we would especially like to thank the anonymous reviewers of all of the ZAP papers for the care and effort they put into reviewing a series of papers that span some 200 journal pages in their entirety. All three of the papers were substantially improved through their efforts.

This work was sponsored in part by grants from the Air Force Office of Scientific Research (AFOSR) number F49620-92-J-0384, the Air Force Research Laboratory (AFRL) number F30602-97-0294, Small Business Technology Transfer Research, Advanced Technology Institute (STTR-ATI) number 20000766, the Office of Naval Research (ONR) number N00014-00-C-0233, the Defense Advanced Research Projects Agency (DARPA) and the Air Force Research Laboratory, Rome, NY, under agreements numbered F30602-95-1-0023, F30602-97-1-0294, F30602-98-2-0181, F30602-00-2-0534, and F33615-02-C-4032, and by DARPA under agreement number HR0011-05-C-0039. The views expressed are those of the authors.

## Appendix A. Proofs

**Procedure 4.1** *Given augmented clauses* $(c_1, G_1)$ *and* $(c_2, G_2)$, *to compute* $\mathtt{stab}(c_i, G_i)$:

1   $\mathtt{c\_closure}_1 \leftarrow c_1^{G_1}$, $\mathtt{c\_closure}_2 \leftarrow c_2^{G_2}$
2   $\mathtt{g\_restrict}_1 \leftarrow G_1|_{\mathtt{c\_closure}_1}$, $\mathtt{g\_restrict}_2 \leftarrow G_2|_{\mathtt{c\_closure}_2}$
3   $C_\cap \leftarrow \mathtt{c\_closure}_1 \cap \mathtt{c\_closure}_2$
4   $\mathtt{g\_stab}_1 \leftarrow \mathtt{g\_restrict}_{1\{C_\cap\}}$, $\mathtt{g\_stab}_2 \leftarrow \mathtt{g\_restrict}_{2\{C_\cap\}}$
5   $\mathtt{g\_int} \leftarrow \mathtt{g\_stab}_1|_{C_\cap} \cap \mathtt{g\_stab}_2|_{C_\cap}$
6   $\{g_i\} \leftarrow \{\text{generators of } \mathtt{g\_int}\}$
7   $\{l_{1i}\} \leftarrow \{g_i, \text{lifted to } \mathtt{g\_stab}_1\}$, $\{l_{2i}\} \leftarrow \{g_i, \text{lifted to } \mathtt{g\_stab}_2\}$
8   $\{l'_{2i}\} \leftarrow \{l_{2i}|_{\mathtt{c\_closure}_2 - C_\cap}\}$
9   **return** $\langle \mathtt{g\_restrict}_{1 C_\cap}, \mathtt{g\_restrict}_{2 C_\cap}, \{l_{1i} \cdot l'_{2i}\} \rangle$

**Proposition 4.2** *The result returned by Procedure 4.1 is* $\mathtt{stab}(c_i, G_i)$.
**Proof.** We show that every element of the group returned is a stable extension by showing that the generators in line 9 are all stable extensions; recall that the set of stable extensions is a subgroup. We show that every stable extension is returned by showing that they can all be constructed via the above procedure.

For the first claim, we argued in the main text that the elements of $\texttt{g\_restrict}_{iC_\cap}$ are stable; we must only show that the elements of $\{l_{1i} \cdot l'_{2i}\}$ are as well. For such an element $\omega$, however, note that $\omega|_{c_1^{G_1}} = l_{1i}|_{c_1^{G_1}} = g_i$ and similarly for $\omega|_{c_2^{G_2}}$, since $\omega$ agrees with $l_{1i} = l_{2i} = g_i$ on $C_\cap$ and with $l_{2i}$ outside of $C_\cap$. Thus $l_{1i} \cdot l'_{2i}$ is stable.

For the second claim, suppose that we have a stable extension $\omega$; consider its restriction to $c_1^{G_1} \cup c_2^{G_2}$. Now on the intersection $c_1^{G_1} \cap c_2^{G_2}$, $\omega$ must agree with elements of both $G_1$ and $G_2$; call the elements with which it agrees $l_1$ and $l_2$. Restricting $l_2$ away from the intersection to get $l'_2$, we see that there will be some element $l$ of the group generated by $\{l_{1i} \cdot l'_{2i}\}$ that matches $\omega$ on $c_1^{G_1} \cap c_2^{G_2}$.

Now consider $\omega \cdot l^{-1}$. This is the identity on $c_1^{G_1} \cap c_2^{G_2}$. Restricting to either $c_1^{G_1}$ or $c_2^{G_2}$ we get an element of $G_1$ or $G_2$ that point stabilizes $c_1^{G_1} \cap c_2^{G_2}$, and all such elements are included directly in line 9 of the resolution procedure. It follows that $\omega \cdot l^{-1}$ is an element of $\langle \texttt{g\_restrict}_{1C_\cap}, \texttt{g\_restrict}_{2C_\cap} \rangle$, so that

$$\omega \in \langle \texttt{g\_restrict}_{1C_\cap}, \texttt{g\_restrict}_{2C_\cap}, \{l_{1i} \cdot l'_{2i}\} \rangle \qquad \square$$

**Procedure 5.3** *Given groups $H \leq G$, an element $t \in G$, sets $c$ and $S$, to find a group element $g = \texttt{map}(G, H, t, c, S)$ with $g \in H$ and $c^{gt} \cap S = \emptyset$:*

1   **if** $c_H^t \cap S \neq \emptyset$
2     **then return** FAILURE
3   **if** $c = c_H$
4     **then return** 1
5   $\alpha \leftarrow$ an element of $c - c_H$
6   **for** each $t'$ in $(H : H_\alpha)$
7      **do** $r \leftarrow \texttt{map}(G, H_\alpha, t't, c, S)$
8        **if** $r \neq$ FAILURE
9          **then return** $rt'$
10 **return** FAILURE

**Proposition 5.4** $\texttt{map}(G, G, 1, c, S)$ *returns an element $g \in G$ for which $c^g \cap S = \emptyset$, if such an element exists, and returns* FAILURE *otherwise.*

**Proof.** As we remarked in the main text, we will prove the slightly stronger result that $\texttt{map}(G, H, t, c, S)$ returns an element $g \in H$ for which $c^{gt} \cap S = \emptyset$ if such an element exists. The proposition as stated is then the special case $t = 1$.

The proof proceeds by induction on the number of elements of $c$ that are moved by $H$. If none are, then either $c^t \cap S \neq \emptyset$ and the procedure will return failure on line 2, or $c^t \cap S = \emptyset$ and it will return 1 on line 4.

For the inductive step, assume that $H$ moves at least one point in $c$. Lines 1–4 don't affect the correctness of the procedure at this point, other than to allow an early termination if some already fixed point is moved inside of $S$ by $t$. In the interesting case, we form a transversal at line 6. Every element of $H$ can be represented as $gt'$ for some $g \in H_\alpha$ and $t'$ in the transversal. If some such $gt'$ should be returned as a solution, we know by the inductive hypothesis that $g$ will be found by the recursive call in line 7.   $\square$

**Procedure 5.5** *Given groups $H \leq G$, an element $t \in G$, sets $c$, $S$ and $U$ and an integer $k$, to find a group element $g = \texttt{transport}(G, H, t, c, S, U, k)$ with $g \in H$, $c^{gt} \cap S = \emptyset$ and $|c^{gt} \cap U| \leq k$:*

1   **if** $c_H^t \cap S \neq \emptyset$
2      **then return** FAILURE
3   **if** `overlap`$(H, c, (S \cup U)^{t^{-1}}) > k$
4      **then return** FAILURE
5   **if** $c = c_H$
6      **then return** 1
7   $\alpha \leftarrow$ an element of $c - c_H$
8   **for** each $t'$ in $(H : H_\alpha)$
9      **do** $r \leftarrow$ `transport`$(G, H_\alpha, t't, c, S, U, k)$
10        **if** $r \neq$ FAILURE
11          **then return** $rt'$
12  **return** FAILURE

**Proposition 5.6** *Provided that $|c^h \cap V| \geq$ `overlap`$(H, c, V) \geq |c_H \cap V|$ for all $h \in H$, `transport`$(G, c, S, U, k)$ as computed by Procedure 5.5 returns an element $g \in G$ for which $c^g \cap S = \emptyset$ and $|c^g \cap U| \leq k$, if such an element exists, and returns* FAILURE *otherwise.*

**Proof.** As remarked in the main text, $|c \cap (S \cup U)^{t^{-1}}| = |c^t \cap (S \cup U)|$. But since $c^t \cap S$ is required to be empty, $|c^t \cap (S \cup U)| = |c^t \cap U|$. The proof now proceeds essentially unchanged from that of Proposition 5.4.

The two conditions on the `overlap` function are both necessary. We need to know that $|c^h \cap V| \geq$ `overlap`$(H, c, V)$ in order to avoid terminating the search early on line 3. We need `overlap`$(H, c, V) \geq |c_H \cap V|$ to ensure that once we have fixed every element of $c$, line 3 will identify a failure if $|c^t \cap U| > k$ so that we don't return successfully on line 6 in this case.   □

**Procedure 5.8** *Given a group $H$, and two sets $c, V$, to compute* `overlap`$(H, c, V)$, *a lower bound on the overlap of $c^h$ and $V$ for any $h \in H$:*

1   $m \leftarrow 0$
2   **for** each orbit $W$ of $H$
3      **do** $m \leftarrow m + \max(|W \cap V| - |W - c|, 0)$
4   **return** $m$

**Proposition 5.9** *Let $H$ be a group and $c, V$ sets acted on by $H$. Then for any $h \in H$, $|c^h \cap V| \geq$ `overlap`$(H, c, V) \geq |c_H \cap V|$ where* `overlap` *is computed by Procedure 5.8.*

**Proof.** The only subtlety involves the contribution that the fixed points in the clause make to the sum. But since each fixed point is in its own orbit, the fixed points contribute either 1 or 0 to the sum depending on whether or not they are already in $V$.   □

**Proposition 5.15** *Let $G$ be a group acting transitively on a set $T$, and let $c, V \subseteq T$. Suppose also that $\{B_1, \ldots, B_k\}$ is a block system for $G$ and that $c \cap B_i \neq \emptyset$ for $n$ of the blocks in $\{B_1, \ldots, B_k\}$. Then if $b$ is the size of an individual block $B_i$ and $g \in G$,*

$$|c^g \cap V| \geq |c| + \Sigma_{i \leq n}^{\min}\ (B_i \cap V) - nb \tag{32}$$

**Proof.** For any $g \in G$, there will be a set of $n$ blocks that collectively contain the image $c^g$. We can therefore use the usual counting argument. Within those $n$ blocks, $c$ will contain $|c|$ points, and the set $V$ will contain at least $\Sigma_{i\leq n}^{\min} \ (B_i \cap V)$ points. But there are only $nb$ points available, so the result follows. □

**Proposition 5.16** *If the block system is trivial (in either sense), (32) is equivalent to*

$$|c^g \cap V| \geq |T \cap V| - |T - c| \tag{33}$$

**Proof.** Suppose first that there is a single block. Now $n = 1$, $b = |T|$ and there is only one set over which to take the minimum in (32), which therefore becomes

$$\begin{aligned}
|c^g \cap V| &\geq |c| + |T \cap V| - |T| \\
&= |T \cap V| - |T - c|
\end{aligned}$$

If, on the other hand, the block system is trivial in that each point is in its own block, then $n = |c|$, $b = 1$ and

$$\Sigma_{i\leq n}^{\min} \ (B_i \cap V)$$

is the smallest number of points in $V$ that must be in a set of size $n$, so

$$\Sigma_{i\leq n}^{\min} \ (B_i \cap V) = n + |T \cap V| - |T|$$

Now (32) becomes

$$\begin{aligned}
|c^g \cap V| &\geq |c| + |c| + |T \cap V| - |T| - |c| \\
&= |c| + |T \cap V| - |T| \\
&= |T \cap V| - |T - c| \quad □
\end{aligned}$$

**Proposition 5.17** *Let $\{B_1, \ldots, B_k\}$ be a block system for a group $G$ acting transitively on a set $T$. Then (32) is never weaker than (33).*

**Proof.** Comparing (32) and (33), we see that we are trying to show that

$$\begin{aligned}
|c| + \Sigma_{i\leq n}^{\min} \ (B_i \cap V) - nb &\geq |T \cap V| - |T - c| \\
&= |c| + |T \cap V| - |T|
\end{aligned}$$

or

$$\Sigma_{i\leq n}^{\min} \ (B_i \cap V) - nb \geq |T \cap V| - |T|$$

If there are $q$ blocks in the block system, then this is equivalent to

$$\Sigma_{i\leq n}^{\min} \ (B_i \cap V) - nb \geq \Sigma_{i\leq q}^{\min} \ (B_i \cap V) - bq$$

or

$$bq - nb \geq \Sigma_{i\leq q}^{\min} \ (B_i \cap V) - \Sigma_{i\leq n}^{\min} \ (B_i \cap V) \tag{34}$$

But the lefthand side of (34) is the total amount of space in the $q - b$ blocks not included in $\Sigma_{i\leq n}^{\min} \ (B_i \cap V)$, and the righthand side is the amount of space used by $V$ within these $q - b$ blocks. Thus (34) follows and the result is proved. □

**Lemma A.1** *Let $G$ be a group of permutations, and $c$ a set acted on by $G$. Suppose also that $S$ and $U$ are sets acted on by $G$. Now if $j \in G_{\{c\}}$ and $g \in G$ is any permutation in $G$, then*

$$|c^g \cap S| = |c^{jg} \cap S|$$

*and*

$$|c^g \cap U| = |c^{jg} \cap U|$$

**Proof.** This is immediate, since $c^j = c$. □

**Lemma A.2** *Let $G$ be a group of permutations, and $c$ a set acted on by $G$. Suppose also that $S$ and $U$ are sets acted on by $G$. Now if $k \in G_{\{S,U\}}$ and $g \in G$ is any permutation in $G$, then*

$$|c^g \cap S| = |c^{gk} \cap S|$$

*and*

$$|c^g \cap U| = |c^{gk} \cap U|$$

**Proof.** It clearly suffices to show the result for $S$; $U$ is equivalent. But

$$
\begin{aligned}
|c^{gk} \cap S| &= |c^g \cap S^{k^{-1}}| \\
&= |c^g \cap S|
\end{aligned}
$$

where $S^{k^{-1}} = S$ because $k$ is in the set stabilizer of $S$ and therefore $k^{-1}$ is as well (because the set stabilizer of $S$ is a group). □

**Proposition 5.23** *Let $G$ be a group of permutations, and $c$ a set acted on by $G$. Suppose also that $S$ and $U$ are sets acted on by $G$. Then for any instance $I$ of the $k$-transporter problem and any $g \in G$, either every element of $G_{\{c\}}gG_{\{S,U\}}$ is a solution of $I$, or none is.*

**Proof.** Combine lemmas A.1 and A.2. □

**Lemma A.3** *Let $G, J \leq \mathrm{Sym}(\Omega)$ where $\Omega$ is the (ordered) set $\{x_1, \ldots, x_n\}$ and suppose $t \in \mathrm{Sym}(\Omega)$ satisfies $x_l^t = z_l$ for $1 \leq l \leq k$ where $k \leq n$. Suppose that we have fixed $i$ with $i \leq k$ and set $Z = J_{\{x_i, \ldots, x_k\}}$. Suppose finally that*

$$z_i > \min\left(x_i^{Z_{x_1, \ldots, x_{i-1}}t}\right)$$

*Then no $h \in G_{x_1, \ldots, x_k}t$ is the first element of $Jh$.*

**Proof.** We are given the existence of $j \in Z_{x_1, \ldots, x_{i-1}}$ such that $z_i > x_i^{jt}$. Consider any $h = gt$ with $g \in G_{x_1, \ldots, x_k}$. Since $j \in Z$, $j$ stabilizes the set $\{x_i, \ldots, x_k\}$. Since $g$ stabilizes every point in this set, it fixes both $x_i$ and $x_i^j$. Thus $x_i^{gt} = x_i^t$ and $x_i^{jgt} = x_i^{jt}$, and

$$x_i^{gt} = x_i^t = z_i > x_i^{jt} = x_i^{jgt}$$

On the other hand, for $l < i$, both $g$ and $j$ fix $x_l$, so that $x_l^{gt} = x_l^{jgt}$. Since $jgt$ thus precedes $gt$, $gt$ is not minimal in $Jgt$. ◻

**Lemma 5.26** *Suppose that $t$ is the permutation labeling some node $Ht$ of a coset decomposition tree at depth $k$, so that $x_i^t = z_i$ for $i \leq k$ and $H = G_{x_1,\ldots,x_k}$ is the residual group at this level. Let $M$ be the set of points moved by $G_{x_1,\ldots,x_k}$. Now if $z_i > \min\left(x_i^{J_{M,x_1,\ldots,x_{i-1}}t}\right)$ for any $i \leq k$, then no $g \in Ht$ is the first element of $JgK$.*

**Proof.** This is a direct consequence of Lemma A.3. Let $\rho$ be a permutation in $J_{M,x_1,\ldots,x_{i-1}}$. Since $\rho$ fixes every point moved by $G_{x_1,\ldots,x_k}$, and $\rho$ also fixes $x_1,\ldots,x_{i-1}$, it follows that $\rho$ must only permute the remaining points $x_i,\ldots,x_k$. Thus $J_{M,x_1,\ldots,x_{i-1}} \leq Z_{x_1,\ldots,x_{i-1}}$ where $Z$ is the set stabilizer in the statement of Lemma A.3, and therefore no $g \in T$ is the first element of $Jg$. Since $Jg \subseteq JgK$, the result follows. ◻

**Procedure 6.5** *Given a set $X$ of pairs $\langle l, g \rangle$ and a group $G$, to compute* complete$(X, G)$, *where $X \models_G$ complete$(X, G)$ and $L(\text{complete}(X, G)) = L(X)^G$:*

1   $Y \leftarrow \emptyset$
2   **for** each $\langle l, g \rangle \in X$
3           **do for** each $l' \in l^G - L(Y)$
4                   **do** select $h \in G$ such that $l^h = l'$
5                       $Y \leftarrow Y \cup \langle l', gh \rangle$
6   **return** $Y$

**Proposition 6.6** $X \models_G$ complete$(X, G)$ *and* $L(\text{complete}(X, G)) = L(X)^G$.

**Proof.** $X \models_G$ complete$(X, G)$ because every entry added to $Y$ is clearly $G$-entailed by $X$. $L(\text{complete}(X, G)) = L(X)^G$ because the entire image of $L(X)$ under $G$ is eventually added. ◻

**Procedure 6.7** *Given groups $H \leq G$, an element $t \in G$, sets $c$, $S$ and $U$, to find* Transport$(G, H, t, c, S, U)$, *a skeletal set of unit consequences for $(c, G)$ given $P$:*

```
1  if overlap(H, c, S^{t^{-1}}) > 0
2      then return ⟨false, ∅⟩
3  if overlap(H, c, (S ∪ U)^{t^{-1}}) > 1
4      then return ⟨false, ∅⟩
5  if c = c_H
6      then if c^t ∩ U = ∅
7              then return ⟨true, 1⟩
8              else  return ⟨false, ⟨c^t ∩ U, 1⟩⟩
9  if a pruning lemma can be applied
10     then return ⟨false, ∅⟩
11 Y ← ∅
12 α ← an element of c − c_H
13 for each t' in (H : H_α)
14     do ⟨u, V⟩ ← Transport(G, H_α, t't, c, S, U)
15         if u = true
16             then return ⟨true, Vt'⟩
17             else  Y ← Y ∪ {⟨l, gt'⟩|⟨l, g⟩ ∈ V}
18 return ⟨false, Y⟩
```

**Proposition 6.8** *Assume that $|c^h ∩ V| ≥ \text{overlap}(H, c, V) ≥ |c_H ∩ V|$ for all $h ∈ H$, and let* Transport$(G, c, S, U)$ *be computed by Procedure 6.7. Then if there is a $g ∈ G$ such that $c^g ∩ S = c^g ∩ U = ∅$,* Transport$(G, c, S, U) = ⟨\text{true}, g⟩$ *for such a $g$. If there is no such $g$,* Transport$(G, c, S, U) = ⟨\text{false}, Z⟩$, *where $Z$ is a skeletal set of unit consequences for $(c, G)$ given $P$.*

**Proof.** Procedure 6.7 is identical to Procedure 5.27 with $k = 1$ except for the value returned. If there is a $g$ with $c^g ∩ S = ∅$ and $c^g ∩ U = ∅$ as well, $⟨\text{true}, g⟩$ will be returned on line 7, and this will cause $⟨\text{true}, gt'⟩$ to be returned from the recursive call(s) on line 16 also.

If there is no $g$ with $c^g ∩ S = c^g ∩ U = ∅$, then the argument proceeds as usual by induction on the number of points of $c$ moved by $H$. If none, we know that the correct answer is returned on line 8 for the usual reasons; it remains to consider the recursive case on line 18. We know that for every $g$ such that $c^g$ is unit, we will accumulate a result from that $g'$ that is minimal in $JgK$ where $J = G_{\{c\}}$ and $K = G_{\{S,U\}}$ as usual. We only need to show that the set of $⟨l, g⟩$ collected is indeed a skeletal set of unit consequences.

To see this, suppose that $⟨l, g⟩$ is any annotated unit consequence. Then there is some minimal $jgk$ that will be accumulated when the set of pairs is accumulated on line 17, with the associated literal $l' = c^{jgk} ∩ U$. But since $j ∈ G_{\{c\}}$ set stabilizes the clause $c$, $c^j = c$ and $l' = c^{gk} ∩ U$. Thus taking the given $k ∈ G_{\{S,U\}}$ produces the given unit consequence from the element of the proposed skeleton, and $Y$ as returned by Procedure 6.7 is indeed a skeletal set of unit consequences. □

**Proposition 6.9** *Let $(c, G)$ be an augmented clause corresponding to a cardinality constraint. Then for any sets $S$ and $U$, Procedure 6.7 will expand at most a linear number of nodes in finding a skeletal set of unit consequences of $(c, G)$.*

**Proof.** If the original cardinality constraint was

$$x_1 + \cdots + x_m \geq n$$

then $G$ will be $\mathrm{Sym}(X)$ where $X$ is the set of $x_i$ and $c$ will be

$$x_1 \vee \cdots \vee x_{m-n+1}$$

We will first show that Leon's pruning lemma 5.24 suffices to reduce the search to quadratic size. The basic idea of this part of the proof is as follows.

Suppose that we are expanding a particular node, corresponding to the selection of an image for a point $x_i$ in $c$. If the image of $x_i$ is selected to be in $S$, we can prune immediately. If the image is selected to be in either $U$ or $X - S - U$, the image will have to be the smallest available point in the set in question for lexicographic reasons. In addition, the original symmetry on the literals in $c$ can be used to require that the literals that are neither satisfied nor unvalued are selected "in order" during the expansion.

To make this argument formally, note first that $J = G_{\{c\}} = \mathrm{Sym}(c) \times \mathrm{Sym}(X - c)$ and $K = G_{\{S,U\}} = \mathrm{Sym}(S) \times \mathrm{Sym}(U) \times \mathrm{Sym}(X - S - U)$. We assume without loss of generality that the points fixed in the coset decomposition tree are the $x_i$ in order for $i \leq m - n + 1$, and will continue to denote the fixed image of $x_i$ at any particular search node by $z_i$. We denote by $\Gamma$ the sequence of all $z_i$ for $i$ less than the depth of the node in question, so $\Gamma$ is the fixed part of the image of the clause $c$. We also set $l = |X - S - U|$, the total number of points that are valued but unsatisfied.

We can now prune any node for which:

**1. $\Gamma \cap S \neq \emptyset$.** These nodes can be pruned because the image of $c$ meets the set $S$ of satisfied literals.

**2. $|\Gamma \cap U| > 1$.** As above, these nodes will be pruned because the image of $c$ includes two or more unsatisfied literals.

**3. $\Gamma = \langle y_1, ..., y_j, u \rangle$, where each $y_i \in X - U - S$, and $u \in U$ is not minimal in U.** Leon's lemma 5.24 with $k = l$ requires that $u = z_{j+1} \leq \min(u^{K_{y_1,...,y_j}})$. But since all of the $y_i$ are outside of $U$, $K_{y_1,...,y_j} \geq \mathrm{Sym}(U)$ and $u^{K_{y_1,...,y_j}}$ is all of $U$. Since $u$ is assumed nonminimal in $U$, the node can be pruned.

**4. $\Gamma \cap (X - S - U) = \langle y_1, ..., y_j \rangle$, where $y_1, ..., y_{j-1}$ are the first $j - 1$ elements of $X - U - S$ and are in order, but $y_j \in X - U - S$ is not the next element of $X - U - S$.** An argument identical to that in the previous paragraph can be used, since $K_{y_1,...,y_{j-1}}$ includes the full symmetry group on the remaining elements of $X - U - S$.

It follows from this that the only unpruned nodes in the search are those for which either $\Gamma = \langle y_1, ..., y_k \rangle$ for $k \leq \min(l, m - n + 1)$, or

$$\Gamma = \langle y_1, ..., y_j, u, y_{j+1}, ..., y_k \rangle \tag{35}$$

for $k \leq \min(l, m - n)$, $u$ the minimal element of $U$, and the $y_i$ the smallest elements of $X - U - S$ in order. We need $k \leq l$ because there are only that many possible $y$ values, and $k \leq m - n + 1$ or $k \leq m - n$ because that is the depth of the tree when the clause $c$ has been completely instantiated. There is a linear number of nodes of the first type but a quadratic number of nodes of the second.

To reduce the total number of nodes being searched to linear, we repeat the argument used in the discussion of the example following Lemma 5.26. There, we considered a node where the image of $x_1$ was $z_1$ and that of $x_2$ was $z_2$, with $z_1 > z_2$. Here, we consider a slightly more general case, where $\Gamma = \langle z_1, \ldots, z_{k-1}, z_k \rangle$, with all of the $z_i$ in sequence except $z_{k-1} > z_k$.

In Lemma 5.26, $G_{x_1, \ldots, x_k}$ will be the full symmetry group on the remaining $x_i$, so that $M = \{x_{k+1}, \ldots, x_m\}$. We also have $J = \text{Sym}(x_1, \ldots, x_{m-n+1}) \times \text{Sym}(x_{m-n+2}, \ldots, x_m)$. Now since $k \leq m - n + 1$, taking $i = k - 1$ in the statement of the lemma gives us

$$J_{M, x_1, \ldots, x_{i-1}} = J_{M, x_1, \ldots, x_{k-2}} \geq \text{Sym}(x_{k-1}, x_k)$$

As a result,

$$z_{k-1} > x_{k-1}^{(x_{k-1} x_k) \cdot t} = x_k^t = z_k$$

and the node can be pruned.

This fixes $u$'s position in the list to be at the point where it is in sequence among the $y_i$ and thus reduces the number of search nodes to linear. □

**Proposition 7.8** *Suppose that $W$ is a watching set for $C$ under $P$ and $l$ is a literal. Then:*

1. *$W$ is a watching set for $C$ under any backtrack point for $P$.*

2. *If $C$ is settled by $\langle P, l \rangle$, then $W$ is a watching set for $C$ under $\langle P, l \rangle$.*

3. *If $C$ is settled by $\langle P, l \rangle$, and $|(W - \{\neg l\}) \cap C \cap U(P_{\neg C})| > 1$, then $W - \{\neg l\}$ is a watching set for $C$ under $\langle P, l \rangle$.*

4. *If $\neg l \notin W \cap C$, then $W$ is a watching set for $C$ under $\langle P, l \rangle$.*

**Proof.** None of these is hard. First, note that if $P'$ is a backtrack point for $P$, then $P'_{\neg c}$ will be a subassignment of $P_{\neg c}$, so a watching set for $C$ under $P$ will also be a watching set for $C$ under $P'$.

For the second claim, if $C$ is settled by $\langle P, l \rangle$, there are two possibilities:

1. If $C$ is unsettled by $P$ (so that the addition of $l$ to $P$ caused $C$ to be settled), then $\langle P, l \rangle_{\neg C}$ is a subassignment of $P$ (the subassignment will be proper if $P \neq \overline{P}$). Since $C$ is unsettled by $P$, $P_{\neg C} = P$. Thus $U(\langle P, l \rangle_{\neg C}) \supseteq U(P_{\neg C})$, and $W$ is still a watching set.

2. If $C$ is settled by $P$, then $\langle P, l \rangle_{\neg C} = P_{\neg C}$, and $W$ is once again still a watching set.

The third claim follows from the second, since $W - \{\neg l\}$ is assumed to be a watching set for $C$ under $P$.

For the fourth claim, suppose that $l \notin C$ and $\neg l \notin C$. Now $C \cap U(P) = C \cap U(\langle P, l \rangle)$, and $W$ remains a watching set. If $l \in C$, then $C$ will be satisfied (and therefore settled) after $l$ is added to $P$. So $W$ continues to be a watching set by virtue of the second claim.

In the remaining case, $\neg l \in C$ and $C \cap U$ is potentially smaller because $\neg l$ is removed after $l$ is adjoined to $P$. But this can only impact the intersection with $W$ if $\neg l$ is itself in $W$; otherwise, $W$ will still be a watching set. So $W$ is still a watching set unless $\neg l$ is in both $C$ and $W$, which proves the final claim. □

**Proposition 7.10** *Suppose that $W$ is a watching set for $(c, G)$ under $P$ and $l$ is a literal. Then:*

1. *$W$ is a watching set for $(c, G)$ under any backtrack point for $P$.*

2. *If $\neg l \notin W \cap c^G$, then $W$ is a watching set for $(c, G)$ under $\langle P, l \rangle$.*

3. *If $|(W \cup V) \cap c^g \cap U(\langle P, l \rangle)| > 1$ for every $g \in G$ such that $c^g$ is unsettled by $\langle P, l \rangle$, then $W \cup V$ is a watching set for $(c, G)$ under $\langle P, l \rangle$.*

4. *If $|(W \cup V) \cap c^g \cap [U(\langle P, l \rangle) \cup (S(P) - S(P_-))]| > 1$ for every $g \in G$, then $W \cup V - \{\neg l\}$ is a watching set for $(c, G)$ under $\langle P, l \rangle$.*

**Proof.** We know that $W$ is a watching set for every instance of $(c, G)$ under $P$, and use Proposition 7.8 to show that each of the above claims follows.

First, Proposition 7.8 states directly that $W$ is a watching set for every instance of $(c, G)$ under a backtrack point for $P$.

Second, if $\neg l \notin W \cap c^G$, then for any $g \in G$, $\neg l \notin W \cap c^g$. The second claim here thus follows from the fourth claim in Proposition 7.8.

The remaining two claims are more interesting. For the third, suppose that $c^g$ is some instance of $(c, G)$. Now if $c^g$ is settled by $\langle P, l \rangle$, then we know that $W$ will still be a watching set for it under $\langle P, l \rangle$. Therefore $W \cup V$ will also be a watching set for $c^g$ under $\langle P, l \rangle$. If $c^g$ is unsettled by $\langle P, l \rangle$, the condition of this claim says that $|(W \cup V) \cap c^g \cap U(\langle P, l \rangle)| > 1$, so that $W \cup V$ is a watching set for $c^g$ under $\langle P, l \rangle$. This completes the proof of the third claim.

For the fourth and final claim, there are three cases.

1. If $c^g$ is unsettled by $\langle P, l \rangle$, note first that $c^g \cap S(P) = \emptyset$, so that

$$(W \cup V) \cap c^g \cap [U(\langle P, l \rangle) \cup (S(P) - S(P_-))] = (W \cup V) \cap c^g \cap U(\langle P, l \rangle)$$

and $W \cup V$ is a watching set for $c^g$ under $\langle P, l \rangle$. Since $\neg l \notin U(\langle P, l \rangle)$,

$$(W \cup V) \cap c^g \cap U(\langle P, l \rangle) = (W \cup V - \{\neg l\}) \cap c^g \cap U(\langle P, l \rangle)$$

and $W \cup V - \{\neg l\}$ is a watching set as well.

2. If $c^g$ is unit under $\langle P, l \rangle$, consider:

   (a) If $\neg l \notin c^g$, then we know from the fourth claim of Proposition 7.8 that $W$ is a watching set for $c^g$ under $\langle P, l \rangle$. It follows that $W - \{\neg l\}$ must be as well, since $\neg l \notin c^g$. Thus so is $W \cup V - \{\neg l\}$.

(b) If $\neg l \in c^g$, $c^g$ must be of the form

$$c^g = x_1 \vee \cdots \vee x_k \vee \neg l \vee u$$

for the new unit consequence $u$, where no $x_i \in S(P)$. Note also that $\neg l$ cannot be in either $U(\langle P, l \rangle)$ or $S(P)$. Thus

$$c^g \cap [U(\langle P, l \rangle) \cup (S(P) - S(P_-))] = \{u\}$$

in violation of the premise of the claim.

3. Finally, if $c^g$ is satisfied by $\langle P, l \rangle$, we know that $W$ (and therefore $W \cup V$) is a watching set for $c^g$ under $\langle P, l \rangle$; the trick is to show that we can remove $\neg l$ from $W \cup V$ safely. If $\neg l \notin c^g$, then we can obviously do so.

If $\neg l \in c^g$, we need to show that the third claim of Proposition 7.8 can be applied, so we need to show that

$$|(W \cup V - \{\neg l\}) \cap c^g \cap U(P_{\neg c^g})| > 1 \tag{36}$$

Given the assumption that

$$|(W \cup V) \cap c^g \cap [U(\langle P, l \rangle) \cup (S(P) - S(P_-))]| > 1 \tag{37}$$

note first that since $\neg l \notin U(\langle P, l \rangle) \cup (S(P) - S(P_-))$, $\neg l$ is not in the intersection of (37), which is therefore equivalent to

$$|(W \cup V - \{\neg l\}) \cap c^g \cap [U(\langle P, l \rangle) \cup (S(P) - S(P_-))]| > 1$$

It follows that (36) will follow if we can show that

$$U(P_{\neg c^g}) \supseteq U(\langle P, l \rangle) \cup (S(P) - S(P_-)) \tag{38}$$

But

$$U(P_{\neg c^g}) \supseteq U(\langle P, l \rangle) \tag{39}$$

because $P_{\neg c^g}$ is a (proper) subassignment of $\langle P, l \rangle$. And we also have

$$U(P_{\neg c^g}) \supseteq U(P_-) \supseteq S(P) - S(P_-) \tag{40}$$

The first inclusion holds because since $\neg l \in c^g$ and $c^g$ is satisfied by $\langle P, l \rangle$, $c^g$ must have been satisfied by $P$ as well. Thus $P_{\neg c}$ involves a backtrack from $P$, and since $P_-$ is the last backtrack point before $P$, $P_{\neg c^g}$ is a subassignment of $P_-$ and $U(P_{\neg c^g}) \supseteq U(P_-)$. The second inclusion in (40) holds because the literals that are satisfied in $P$ but not in $P_-$ must necessarily have been unvalued in $P_-$. Combining (39) and (40) gives us (38), and the result is proved. □

**Procedure 7.11** *Given a group $H$, two sets $c, V$ acted on by $H$, and a bound $k \geq 0$, to compute* `overlap`$(H, c, V, k)$, *a collection of elements of $V$ sufficient to guarantee that for any $h \in H$, $|c^h \cap V| > k$, or $\emptyset$ if no such collection exists:*

```
1   m ← 0
2   W ← ∅
3   for each orbit X of H
4       do {B₁, . . . , Bₖ} ← a minimal block system for W under H for which
                c ∩ W ⊆ Bᵢ for some i
5          Δ = |c ∩ X| + min(Bᵢ ∩ V) − |B₁|
6          if Δ > 0
7             then m ← m + Δ
8                  W ← W ∪ (X ∩ V)
9                  if m > k
10                    then return W
11  return ∅
```

**Proposition 7.12** *Procedure 7.11 returns a nonempty set $W$ if and only if Procedure 5.19 returns a value in excess of $k$. In this case, $|c^h \cap W| > k$ for every $h \in H$.*

**Proof.** For the first claim, we examine the two procedures. It is clear that Procedure 7.11 returns as soon as Procedure 5.19 concludes that the minimum overlap is greater than $k$; we need simply argue that $W$ will be nonempty. But each increment to $W$ in line 8 must be nonempty, since if $X \cap V = \emptyset$, $\Delta$ will be zero on line 6.

For the second part, imagine replacing $V$ in the procedure with the set $W$ returned. The computation will be unchanged at every step, so the conclusion follows. □

**Procedure 7.13** *Given groups $H \leq G$, an element $t \in G$, sets $c$, $S$ and $U$, and optionally a watched element $w$, to find $\mathtt{Transport}(G, H, t, c, S, U, w)$, a skeletal set of unit $w$-consequences for $(c, G)$ given $P$:*

1  **if** $w$ is supplied and $w^{t^{-1}} \notin c^H$
2      **then return** $\langle \mathtt{false}, \varnothing, \varnothing \rangle$
3  $V \leftarrow \mathtt{overlap}(H, c, S^{t^{-1}}, 0)$
4  **if** $V \neq \varnothing$
5      **then return** $\langle \mathtt{false}, \varnothing, \varnothing \rangle$
6  $V \leftarrow \mathtt{overlap}(H, c, (S \cup U)^{t^{-1}}, 1)$
7  **if** $V \neq \varnothing$
8      **then return** $\langle \mathtt{false}, \varnothing, V^t \rangle$
9  **if** $c = c_H$
10      **then if** $c^t \cap U = \varnothing$
11              **then return** $\langle \mathtt{true}, 1, \varnothing \rangle$
12              **else return** $\langle \mathtt{false}, \langle c^t \cap U, 1 \rangle, \varnothing \rangle$
13  **if** a pruning lemma can be applied
14      **then return** $\langle \mathtt{false}, \varnothing, \varnothing \rangle$
15  $\alpha \leftarrow$ an element of $c - c_H$. If $w$ is supplied and $w \notin c_H^t$, choose $\alpha$ so that $w^{t^{-1}} \in \alpha^H$.
16  $Y \leftarrow \varnothing$
17  $W \leftarrow \varnothing$
18  **for** each $t'$ in $(H : H_\alpha)$
19      **do** $\langle u, V, X \rangle \leftarrow \mathtt{Transport}(G, H_\alpha, t't, c, S, U, w)$
20          **if** $u = \mathtt{true}$
21              **then return** $\langle \mathtt{true}, Vt', \varnothing \rangle$
22              **else** $W \leftarrow W \cup X$
23                  $Y \leftarrow Y \cup \{\langle l, gt' \rangle | \langle l, g \rangle \in V\}$
24  **return** $\langle \mathtt{false}, Y, W \rangle$

**Proposition 7.14** *Suppose that* $\mathtt{overlap}(H, c, V, k)$ *is computed using Procedure 7.11, or otherwise satisfies the conclusion of Proposition 7.12. Then if there is a* $g \in G$ *such that* $w \in c^g$ *and* $c^g \cap S = c^g \cap U = \varnothing$, $\mathtt{Transport}(G, c, S, U, w)$ *as computed by Procedure 7.13 returns* $\langle \mathtt{true}, g, \varnothing \rangle$ *for such a* $g$. *If there is no such* $g$, *Procedure 7.13 returns* $\langle \mathtt{false}, Z, W \rangle$, *where* $Z$ *is a skeletal set of unit* $w$-*consequences of* $(c, G)$ *given* $P$, *and* $W$ *is such that* $|W^{G_{\{S, U, \{w\}\}}} \cap c^h \cap U| > 1$ *for every* $h \in H$ *such that* $w \in c^h$ *and* $c^h$ *is unsettled by* $P$.

**Proof.** The restriction to permutations $g$ for which $w \in c^g$ is enforced by the first two lines of the procedure; note that if a contradiction is found on line 11, all of the points in $c$ will have been fixed, so $w \in c^g$ for certain. Note that we *can* prune on this basis without affecting the overall correctness of the procedure, since the pruning lemmas have been restricted to the group $K = G_{\{S, U, \{w\}\}}$, which leaves the watched literal $w$ intact.

The only other difference between Procedure 7.14 and Procedure 6.7 involves the computation of the set $W$ of watched literals. When this set is produced on line 8, we know from Proposition 7.12 that the set $W$ is sufficient to guarantee that $|W \cap c^{ht} \cap U| > 1$ for every $c^{ht}$ in the current residual search tree. We must therefore show than any $h$ satisfying the conditions in the proposition is covered by $W^{G_{\{S, U, \{w\}\}}}$. To see this, we consider every point at which a node is pruned in the procedure, and show that all such points are covered by the exclusions in the statement of the proposition:

1. A prune at line 2 will only occur if $w \notin c^{ht}$ for any $h \in H$.

2. A prune at line 5 will only occur if $c^{ht} \cap S \neq \emptyset$ for every $h \in H$, so that $c^{ht}$ is settled by $P$.

3. If a pruning lemma is applied, it must be because an eventual solution $g$ can be shown not to be minimal in the usual double coset $G_{\{c\}} g G_{\{S,U,\{w\}\}}$. But in this case, the watching set itself is operated on with $G_{\{S,U,\{w\}\}}$ in the statement of the proposition itself. □

**Procedure 7.16 (Unit propagation)**  *To compute* Unit-Propagate$(C, P, L)$ *where $C$ is a set of watched augmented clauses, $P$ is an annotated partial assignment, and $L$ is a set of pairs $\langle l, r \rangle$ of literals $l$ and reasons $r$:*

```
1   while L ≠ ∅
2       do ⟨l,r⟩ ← an element of L
3          L ← L − ⟨l,r⟩
4          P ← ⟨P,⟨l,r⟩⟩
5          for each ⟨(c,G),W⟩ ∈ C
6              do if ¬l ∈ W
7                  then ⟨r,H,V⟩ ← Transport(G,c,S(P),U(P),¬l)
8                       if r = true
9                           then lᵢ ← the literal in cᴴ with the highest index in P
10                               return ⟨true,resolve((cᴴ,G),cᵢ)⟩
11                      H′ ← complete(H,G_{S(P),U(P),{l}})
12                      for each h ∈ H′
13                          do z ← the literal in cʰ unassigned by P
14                              if there is no ⟨z,r′⟩ in L
15                                  then L ← L ∪ ⟨z,cʰ⟩
16                      W ← W ∪ (U(P) ∩ V^{G_{S(P),U(P),{l}}})
17                      U ← U(P) ∪ (S(P) − S(P_−))
18                      if H = ∅ ∧ transport(G,c,∅,W ∩ U,1,¬l) = FAILURE
19                          then W ← W − {¬l}
20  return ⟨false,P⟩
```

**Proposition 7.17** *Let $P$ be an annotated partial assignment, and $C$ a set of watched augmented clauses, where for every $\langle (c,G), W \rangle \in C$, $W$ is a watching set for $(c,G)$ under $P$. Let $L$ be the set of unit consequences of clauses in $C$. If* Unit-Propagate$(C, P, L)$ *returns $\langle$true$, c\rangle$ for an augmented clause $c$, then $c$ is a nogood for $P$, and any modified watching sets in $C$ are still watching sets under $P$. Otherwise, the value returned is $\langle$false$, \overline{P}\rangle$ and the watching sets in $C$ will all have been replaced with watching sets under $\overline{P}$.*

**Proof.** This is really just a matter of assembling the pieces. Procedure 7.16 is essentially a loop through the literals in $L$, much like the original procedure 2.7. For each such literal $l$, we find all the unit clauses that contain $l$ by an appropriate call to Transport for each clause where $l$ is watched. If the Transport call reveals the presence of a contradiction, we return the resolvent of the reasons for $l$ and for $\neg l$ as usual. If no contradiction is found, we adjust the partial assignment as in Procedure 2.7, add the new unit consequences to the list of what remains to be analyzed, and update the watching set in accordance with Propositions 7.10 and 7.14.

The only remaining issue is the removal of $\neg l$ from the watching set $W$ in line 19 of Procedure 7.16. We do this precisely when the fourth claim in Proposition 7.10 can be applied. Note that in the call to `transport`, we use $U(P)$ instead of $U(\langle P, l \rangle)$, since $l$ has already been added to $P$ in line 4. We also require that $\neg l$ be in the instance $c^g$, since otherwise the intersection with $c^g$ will obviously be unaffected by the removal of $\neg l$. □

**Lemma 8.3** *If $c_1 \subset c_2$ are two nogoods for $P$, then $c_1 <_P c_2$.*

**Proof.** This is immediate. As soon as the last literal in $c_2$ is not in $c_1$ (which must happen eventually as literals are removed in Definition 8.2), the falsification depth of $c_2$ will exceed that of $c_1$. □

**Procedure 8.4** *Suppose we are given two augmented clauses $(\alpha, G)$ and $(\beta, H)$ that are unit for a partial assignment $P = \langle l_1, \ldots, l_n \rangle$, with $l \in \alpha$ and $\neg l \in \beta$. To find a $<_P$-minimal $l$-resolvent of $(\alpha, G)$ and $(\beta, H)$:*

```
1   U ← {l, ¬l}          ▷ literals you can't avoid
2   α_f ← α
3   β_f ← β
4   p ← [(α ∪ β) − U]^{?P}
5   while p > 0
6       do g ← transport(G, α, {¬l_p, ..., ¬l_n} − U, ∅, 0, l)
7          h ← transport(H, β, {¬l_p, ..., ¬l_n} − U, ∅, 0, ¬l)
8          if g = FAILURE ∨ h = FAILURE
9             then U ← U ∪ {¬l_p}
10            else  α_f ← α^g
11                  β_f ← β^h
12          p ← [(α_f ∪ β_f) − U]^{?P}
13  return resolve(α_f, β_f)
```

**Proposition 8.5** *Suppose that we are given two augmented clauses $(\alpha, G)$ and $(\beta, H)$ such that $\alpha$ and $\beta$ are unit for a partial assignment $P$, with $l \in \alpha$ and $\neg l \in \beta$. Then the value returned by Procedure 8.4 is a $<_P$-minimal $l$-resolvent of $(\alpha, G)$ and $(\beta, H)$.*

**Proof.** We need to show that the procedure terminates, that it returns an $l$-resolvent, and that the result is $<_P$-minimal.

To show that Procedure 8.4 terminates, we show that $p$ is reduced on every iteration of the main loop. At the beginning of each iteration, we know that

$$\neg l_p \in (\alpha_f \cup \beta_f) - U \tag{41}$$

At the end of the iteration, if line 9 is selected, then $\alpha_f$ and $\beta_f$ are unchanged but $\neg l_p$ is added to $U$. This means that (41) will no longer be satisfied, but will still be satisfied for $\neg l_i$ with $i > p$. Thus $p$ is reduced on line 12.

If, on the other hand, lines 10 and 11 are selected, we know from the definition of $g$ and $h$ on lines 6 and 7 that for any literal with $\neg l \in \{\neg l_p, \ldots, \neg l_n\} - U$, we have $\neg l \notin (\alpha_f \cup \beta_f)$. In other words, if $\neg l \in \{\neg l_p, \ldots, \neg l_n\}$, then $\neg l \notin (\alpha_f \cup \beta_f) - U$. Thus $p$ is once again reduced, and the procedure therefore terminates. That it returns a resolvent is clear.

To see that the value returned is $<_P$-minimal, suppose that $g_m$ and $h_m$ are such that $\alpha^{g_m} \vee \beta^{h_m} <_P \alpha_f \vee \beta_f$. We will show that $\alpha_f$ and $\beta_f$ cannot be the permutations returned on line 13.

Set $z = [(\alpha_f \vee \beta_f) - (\alpha^{g_m} \vee \beta^{h_m})]^{?P}$; this is the last point included in the $f$ images produced by the procedure but not in the $m$ images provided by the hypothetical counterexample. Since $\alpha^{g_m} \vee \beta^{h_m} <_P \alpha_f \vee \beta_f$, the two sets agree for literals with index greater than $z$.

Since $\neg l_z \in (\alpha_f \vee \beta_f)$, the initial value for $p$ set in line 4 will be at least $z$; since the procedure terminates when $p < 0$, the final value will be less than $z$.

Consider the point in the procedure at which $p$ changes from a value no less than $z$ to one that is less than $z$. If the change is because $\neg l_p$ is added to $U$, then one of the `transport` calls must have failed, so that it is impossible (say) for the image of $\alpha$ to avoid $\{\neg l_p, \ldots, \neg l_n\} - U$. But we know that $\alpha_f$ avoids $\{\neg l_{p+1}, \ldots, \neg l_n\} - U$. Thus $\alpha^{g_m}$ avoids $\{\neg l_{p+1}, \ldots, \neg l_n\} - U$, but we are assuming that $\neg l_p \notin \alpha^{g_m}$, so `transport`$(G_{l,\neg l}, \alpha, \{\neg l_p, \ldots, \neg l_n\} - U, \emptyset, 0)$ should have succeeded after all.

It follows that the change in $p$ must have been in lines 10 and 11. But this is also impossible, since the fact that we have successfully managed to avoid $\neg l_z$ contradicts the assumption that $z = [(\alpha_f \vee \beta_f) - (\alpha^{g_m} \vee \beta^{h_m})]^{?P}$ so that $\neg l_z \in \alpha_f \vee \beta_f$. ▫

# Appendix B. ZAP Problem Format

Historically, Boolean satisfiability problems are typically in a format where variables correspond to positive integers, literals are nonzero integers (negative integers are negated literals), and clauses are terminated with zeroes. The DIMACS format precedes the actual clauses in the problem with a single line such as `p cnf 220 1122` indicating that there are 220 variables appearing in 1,122 clauses in this problem.

This numerical format makes it impossible to exploit any existing understanding that the user might have of the problem in question; this may not be a problem for a conventional Boolean tool (since the problem structure will have been obscured by the Boolean encoding in any event), but was felt to be inappropriate when building an augmented solver. We felt that it was important for the user to be able to:

1. Specify numerical constraints such as appear in cardinality or parity constraints,

2. Quantify axioms over finite domains, and

3. Provide group augmentations explicitly if the above mechanisms were insufficient.

Before discussing the specific provisions ZAP makes in each of these areas, we remark that each ZAP input file begins with a list of domain specifications, giving the names and sizes of each domain used in the theory. This is followed by predicate specifications, giving the arity of each predicate and the domain type of each argument. After the predicates and domains have been defined, it is possible to refer to predicate instances directly (e.g., `in[1 3]` indicating that the first pigeon is in the third hole) or in a nonground fashion (e.g., `in[x y]`).

**Group definition**   When a group is specified directly, it is assigned a symbolic designator that can then be used in an augmented clause. The group syntax is the conventional one, with a group being described in terms of generators, each of which is a permutation. Each permutation is a list of cycles, and each cycle is a space-separated list of literals.

An augmented clause that uses a previously defined group is of the form

$$\texttt{clause}\ \ \texttt{GROUP}\ \ \texttt{group}_1\ \ \cdots\ \ \texttt{group}_n$$

where the (ground) `clause` is essentially a sequence of literals and each `group`$_i$ is the designator for a group to be used. The group in the augmented clause is then the group collectively generated by the `group`$_i$'s.

As an example, here is the group-based encoding of the pigeonhole instance involving four pigeons and three holes:

```
// domain specs
SORT pigeon 4 ;
SORT hole 3 ;

// predicate specs
PREDICATE in(pigeon hole) ;

// group specs
GROUP G < ((in[1 1] in[2 1]) (in[1 2] in[2 2]) (in[1 3] in[2 3]))
          ((in[1 1] in[3 1] in[4 1]) (in[1 2] in[3 2] in [4 2])
           (in[1 3] in[3 3] in [4 3]))                          // permute pigeons
          ((in[1 1] in[1 2]) (in[2 1] in[2 2]) (in[3 1] in[3 2])
           (in[4 1] in[4 2]))                                   // permute holes
          ((in[1 1] in[1 3]) (in[2 1] in[2 3]) (in[3 1] in[3 3])
           (in[4 1] in[4 3])) > ;

// group-based encoding
-in[1 1] -in[2 1] GROUP G ;
in[1 1] in[1 2] in[1 3] GROUP G ;
```

There are two types of domain variables, pigeons (of which there are four) and holes (of which there are three). There is a single predicate indicating that a given pigeon is in a particular hole. There is a single group, which corresponds to symmetries over both holes and pigeons.

To generate the group, we use four generators. The first two correspond to the symmetry over pigeons, with the first generator swapping the first two pigeons and the second generator rotating pigeons one, three and four. (Recall that the permutations $(1, 2)$ and $(1, 3, 4)$ generate the symmetry group $S_4$ over the pigeons.)

The second pair of generators generate the symmetry over holes similarly, with the first generator swapping the first two holes and the second generator swapping holes one and three. (Once again, $(1, 2)$ and $(1, 3)$ generate $S_3$.)

Turning to the axioms, the first says that the first hole cannot contain both of the first two pigeons, and therefore that no hole can contain two distinct pigeons by virtue of the group action. The second axiom says that the first pigeon has to be in some hole, and therefore that every pigeon does.

**Cardinality and parity constraints** If the group is not specified directly, the general form of a ZAP axiom is

<div align="center">

`quantifiers  clause  result`

</div>

where the `quantifiers` are described below. The `result` includes information about the desired "right hand side" of the axiom, and can be any of the following:

- A simple terminator, indicating that the clause is Boolean,

- A comparison operator (`>`, `<=`, `=`, etc.) followed by an integer, indicating that the clause is a cardinality constraint, or

- A mod-2 operator (`%2=`) followed by an integer $m$, indicating that the sum of the values of the literals is required to be congruent to $m$ mod 2.

**Quantification** The quantifiers are of the form

$$\forall(x_1, \ldots, x_k)$$

or

$$\exists(x_1, \ldots, x_k)$$

where each of the $x_i$ are variables that can then appear in future predicate instances. In addition to the two classical quantifiers above, we also introduce

$$\overline{\forall}(x_1, \ldots, x_k)$$

where the $\overline{\forall}$ quantifier means that the variables can take any values that do not cause any of the quantified predicate's instances to become identical. As an example, the axiom saying that only one pigeon can be in each hole now becomes

$$\overline{\forall}(p_1, p_2, h) \; . \; \neg\mathtt{in}(p_1, h) \vee \neg\mathtt{in}(p_2, h) \tag{42}$$

Contrast this with the conventional

$$\forall(p_1, p_2, h) \; . \; \neg\mathtt{in}(p_1, h) \vee \neg\mathtt{in}(p_2, h) \tag{43}$$

For any specific pigeon $p$ and hole $h$,

$$\neg\mathtt{in}(p, h) \vee \neg\mathtt{in}(p, h) \tag{44}$$

is an instance of (43) but not of (42). Since (44) is equivalent to $\neg\mathtt{in}(p, h)$, it is inappropriate for inclusion in a pigeonhole formulation.

The introduction of the new quantifier should be understood in the light of the discussion of Section 6.1 of ZAP2, where we argued that in many cases, the quantification given by $\overline{\forall}$ is in fact more natural than that provided by $\forall$. The $\overline{\forall}$ quantification is also far easier to represent using augmented clauses, and avoids in many cases the need to introduce or to reason about equality. In any event, ZAP supports both forms of universal quantification.

Here is the quantifier-based encoding of the pigeonhole problem:

```
SORT pigeon 9;
SORT hole 8;
PREDICATE in(pigeon hole);

// quantification-based encoding
NOTEQ (x y z) -in[x z] -in[y z] ;
FORALL(z) EXISTS(h) in[z h] ;
```

This is the nine pigeon instance. The two axioms say directly that no hole $z$ can contain two distinct pigeons $x$ and $y$ (note the use of the `NOTEQ` or $\overline{\forall}$), and every pigeon $z$ has to be in some hole $h$. This encoding is presumably more intuitive than the previous one.